\pdfoutput=1
\newcommand{\mypapersize}{A4}
\newcommand{\mylaterality}{oneside}
\newcommand{\mydraft}{false}
\newcommand{\myparskip}{half}
\newcommand{\myBCOR}{0mm}
\newcommand{\myfontsize}{12pt}
\newcommand{\mylinespread}{1.0}
\newcommand{\mylanguage}{ngerman,british}
\newcommand{\mybiblatexstyle}{alphabetic}
\newcommand{\mybiblatexbackref}{true}  %% "true" or "false"
\newcommand{\mybiblatexfile}{references-biblatex.bib}
\newcommand{\mydispositioncolor}{30,103,182}
\newcommand{\mycolorlinks}{true}  %% "true" or "false"
\newcommand{\mytodonotesoptions}{}
\documentclass[%
fontsize=\myfontsize,%% size of the main text
paper=\mypapersize,  %% paper format
parskip=\myparskip,  %% vertical space between paragraphs (instead of indenting first par-line)
DIV=calc,            %% calculates a good DIV value for type area; 66 characters/line is great
headinclude=true,    %% is header part of margin space or part of page content?
footinclude=false,   %% is footer part of margin space or part of page content?
open=right,          %% "right" or "left": start new chapter on right or left page
appendixprefix=true, %% adds appendix prefix; only for book-classes with \backmatter
bibliography=totoc,  %% adds the bibliography to table of contents (without number)
draft=\mydraft,      %% if true: included graphics are omitted and black boxes
BCOR=\myBCOR,        %% binding correction (depends on how you bind
\mylaterality        %% oneside: document is not printed on left and right sides, only right side
]{scrbook}  %% article class of KOMA: "scrartcl", "scrreprt", or "scrbook".
\usepackage[utf8]{inputenc} %% UTF8 as input characters
\usepackage[\mylanguage]{babel}  %% used languages; default language is *last* language of options

\usepackage[backend=biber, %% using "biber" to compile references (instead of "biblatex")
style=\mybiblatexstyle, %% see biblatex documentation
backref=\mybiblatexbackref, %% create backlings from references to citations
natbib=true, %% offering natbib-compatible commands
hyperref=true, %% using hyperref-package references
]{biblatex}  %% remove, if using BibTeX instead of biblatex

\usepackage[pdftex]{graphicx}

\usepackage{pifont}

\usepackage{ifthen}

\newboolean{myaddcolophon}
\newboolean{myaddlistoftodos}
\newboolean{english_affidavit}

\usepackage{xspace}

\usepackage[usenames,dvipsnames,table]{xcolor}
\definecolor{DispositionColor}{RGB}{\mydispositioncolor} %% used for links and so forth in screen-version

\usepackage[normalem]{ulem}

\usepackage{framed}

\usepackage{eso-pic}

\usepackage{enumitem}

\usepackage[\mytodonotesoptions]{todonotes}  %% option "disable" removes all todonotes output from resulting document

\usepackage{units}

\setboolean{myaddcolophon}{true}  %% "true" or "false"
\setboolean{myaddlistoftodos}{false}  %% "true" or "false"
\setboolean{english_affidavit}{true}  %% "true" or "false"
\newcommand{\myauthor}{David Cemernek}  %% also used for PDF metadata (hyperref)
\newcommand{\myauthorwithexistingtitles}{\myauthor{}, Mag.}  %% including
\newcommand{\mytitle}{Outlier Detection as Instance Selection Method for Feature Selection in Time Series Classification}  %% also used for PDF metadata (hyperref)
\newcommand{\mysubject}{Outlier detection as an instance selection method for feature selection in time series classification}  %% also used for PDF metadata (hyperref)
\newcommand{\mykeywords}{Outlier detection, Instance selection, Feature Selection, Time Series Classification}  %% also used for PDF metadata (hyperref)

\newcommand{\myworktitle}{Master's Thesis}  %% official type of work like ``Master theses''
\newcommand{\mygrade}{Diplom-Ingenieur} %% title you are getting with this work like ``Master of ...''
\newcommand{\mystudy}{Software Engineering and Management} %% your study like ``Arts''
\newcommand{\mydegreeprogramme}{Master's degree programme: \mystudy} %% Master's or PhD degree programme
\newcommand{\myuniversity}{Graz University of Technology} %% your university/school
\newcommand{\myinstitute}{Institute of Interactive Systems and Data Science} %% affiliation
\newcommand{\myinstitutehead}{Univ.-Prof. Dipl.-Inf. Dr. Stefanie Lindstaedt} %% head of institute
\newcommand{\mysupervisor}{Dr. Roman Kern} %% your supervisor
\newcommand{\myhometown}{Graz} %% your home town
\newcommand{\mysubmissionmonth}{February} %% month you are handing in
\newcommand{\mysubmissionyear}{2019} %% year you are handing in
\newcommand{\mysubmissiontown}{\myhometown} %% town of handing in (or \myhometown)

\newcommand{\myfig}[5]{
\begin{figure}[!htp] %enabled for preventing figures changing to different sections
  \centering
  \includegraphics[keepaspectratio,#2]{figures/#1}
  \caption[#4]{#3}
  \label{#5} %% NOTE: always label *after* caption!
\end{figure}
}

\newcounter{myclonecnt}

\usepackage[protrusion=true,factor=900]{microtype}

\usepackage[sc,osf]{mathpazo} %% switches to Palatino with small caps and old style figures

\DeclareRobustCommand{\myacro}[1]{\textsc{\lowercase{#1}}} %%  abbrevations using small caps

\addtokomafont{disposition}{\color{DispositionColor}}
\addtokomafont{caption}{\color{DispositionColor}\footnotesize}
\addtokomafont{captionlabel}{\color{DispositionColor}}

\usepackage[bottom]{footmisc}

\usepackage{enumitem}
\setlist{noitemsep}   %% kills the space between items

\usepackage[babel=true,strict=true,english=american,german=guillemets]{csquotes}

\usepackage[british, ngerman]{babel}
\makeatletter\AtBeginDocument{\let\@elt\relax}\makeatother
\usepackage[backend=biber]{biblatex}
\usepackage{amsmath}
\usepackage{lscape}
\usepackage{longtable}
\usepackage{algorithm}
\usepackage{algorithmic}
\usepackage[utf8]{inputenc}
\usepackage{placeins}
\usepackage{scrhack}

\usepackage[%
unicode=true, % loads with unicode support
backref, %
pagebackref=false, % creates backward references too
bookmarks=false, %
bookmarksopen=false, % when starting with AcrobatReader, the Bookmarkcolumn is opened
pdfpagemode=,% None, UseOutlines, UseThumbs, FullScreen
plainpages=false, % correct, if pdflatex complains: ``destination with same identifier already exists''
urlcolor=DispositionColor, %%
linkcolor=DispositionColor, %%
citecolor=DispositionColor, %%
anchorcolor=DispositionColor, %%
colorlinks=\mycolorlinks, % turn on/off colored links (on: better for
]{hyperref}

\hypersetup{
pdftitle={\mytitle}, %
pdfauthor={\myauthor}, %
pdfsubject={\mysubject}, %
pdfcreator={Accomplished with: pdfLaTeX, biber, and hyperref-package. No animals, MS-EULA or BSA-rules were harmed.},
pdfproducer={\myauthor},
pdfkeywords={\mykeywords}
}

\begin{document}

%\frontmatter                    %% KOMA: roman page numbers and such; only available in scrbook

%%%% Time-stamp: <2013-03-18 14:35:00 vk>
%% ========================================================================
%%%% Disclaimer
%% ========================================================================
%%
%% created by
%%
%%      Karl Voit

\newcommand{\mycolophon}{%%
  This document 
  %% was written with \myacro{GNU}~Emacs, 
  is set in Palatino, compiled with
  \href{http://LaTeX.TUGraz.at}{pdf\LaTeX2e} and
  \href{http://en.wikipedia.org/wiki/Biber_(LaTeX)}{\texttt{Biber}}.

  The \LaTeX{} template from Karl Voit is based on
  \href{http://www.komascript.de/}{KOMA script} and can be found 
  online: \href{https://github.com/novoid/LaTeX-KOMA-template}{https://github.com/novoid/LaTeX-KOMA-template}
}

%%% Local Variables: 
%%% mode: latex
%%% mode: auto-fill
%%% mode: flyspell
%%% eval: (ispell-change-dictionary "en_US")
%%% TeX-master: "../main"
%%% End: 
                %% defines information about editor, LaTeX, font, ...

%% Choose your desired title page:
%%%% Time-stamp: <2017-02-05 16:21:48 vk>
%% ========================================================================
%%%% Disclaimer
%% ========================================================================
%%
%% created by
%%
%%      Stefan Kroboth and Karl Voit
%%
%% this title page fulfills the requirements of the corporate design
%% of Graz University of Technology (correct placement of logo)

\begin{titlepage}

{\sffamily

\begin{center}

\includegraphics[width=30mm]{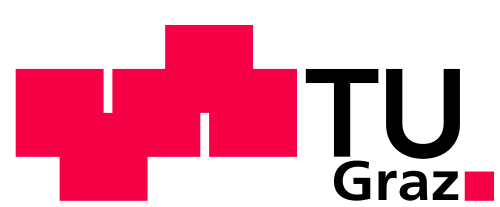}

\vfill\vfill\vfill
\vfill\vfill\vfill

\myauthorwithexistingtitles

\vfill\vfill\vfill

{\large\bfseries\mytitle}

\vfill\vfill\vfill
\vfill\vfill\vfill

{\bfseries\large\myworktitle}

to achieve the university degree of

\mygrade

\mydegreeprogramme

\vfill\vfill\vfill

submitted to

\vfill

{\bfseries\large\myuniversity}

\vfill\vfill\vfill

Supervisor

\mysupervisor
%usually not mentioned on titlepage:% Evaluator: \myevaluator

\vfill

\myinstitute\\
Head: \myinstitutehead\\

\vfill

%% OPTIONAL: second supervisor/name of the faculty, etc.

\vfill\vfill\vfill

\mysubmissiontown, \mysubmissionmonth~\mysubmissionyear

\end{center}
}%% end sffamily
\end{titlepage}

%% if myaddcolophon is set to "true", colophon is added:
\ifthenelse{\boolean{myaddcolophon}}{
  \newpage
  \thispagestyle{empty}  %% no page header or footer

  ~
  \vfill
  \mycolophon
}{}
\newpage

%% end of title page
%%% Local Variables:
%%% mode: latex
%%% mode: auto-fill
%%% mode: flyspell
%%% eval: (ispell-change-dictionary "en_US")
%%% TeX-master: "../main"
%%% End:
            %% include title page

%%%% Time-stamp: <2017-02-14 16:01:12 vk>
%% ========================================================================
%%%% Disclaimer
%% ========================================================================
%%
%% created by
%%
%%      Karl Voit, and Matthias Wölbitsch
%%
%%
%% code for the date and signature fields adapted from
%% http://tex.stackexchange.com/a/20562

\newcommand{\textfield}[2]{
  \vbox{
    \hsize=#1\kern3cm\hrule\kern1ex
    \hbox to \hsize{\strut\hfil\footnotesize#2\hfil}
  }
}

\ifthenelse{\boolean{english_affidavit}}{
  \section*{Affidavit}
  I declare that I have authored this thesis independently, that I have
  not used other than the declared sources/resources, and that I have
  explicitly indicated all material which has been quoted either
  literally or by content from the sources used. The text document
  uploaded to \myacro{TUGRAZ}online is identical to the present master‘s
  thesis.

  \hbox to \hsize{\textfield{4cm}{Date}\hfil\hfil\textfield{7cm}{Signature}}
}{
  \section*{Eidesstattliche Erklärung}
  \foreignlanguage{ngerman}{%
    Ich erkläre an Eides statt, dass ich die vorliegende Arbeit
    selbstständig verfasst, andere als die angegebenen
    Quellen/Hilfsmittel nicht benutzt, und die den benutzten Quellen
    wörtlich und inhaltlich entnommenen Stellen als solche kenntlich
    gemacht habe. Das in \myacro{TUGRAZ}online hochgeladene Textdokument
    ist mit der vorliegenden Dissertation identisch.}

    \hbox to \hsize{\textfield{4cm}{Datum}\hfil\hfil\textfield{7cm}{Unterschrift}}
}

\newpage

%%% Local Variables:
%%% mode: latex
%%% mode: auto-fill
%%% mode: flyspell
%%% TeX-master: "../main"
%%% End:
  %% Statutory Declaration
% \input{thanks}                %% this is a suggestion: you have to create this file on demand
% \input{foreword}              %% this is a suggestion: you have to create this file on demand

%% include the abstract without chapter number but include it on table of contents:
\cleardoublepage
\phantomsection
\addcontentsline{toc}{chapter}{Abstract}
%%%% Time-stamp: <2013-02-25 10:31:01 vk>
\chapter*{Abstract}
\label{cha:abstract}
In order to allow machine learning algorithms to extract knowledge from raw data, these data must first be cleaned, transformed, and put into machine-appropriate form. These often very time-consuming phase is referred to as preprocessing. An important step in the preprocessing phase is feature selection which aims at better performance of prediction models by reducing the amount of features of a data set. Within these datasets instances of different events are often imbalanced, which means that certain normal events are over-represented while other rare events are very limited. Typically these rare events are of special interest since they have more discriminative power than normal events. The aim of this work was to filter instances provided to feature selection methods for these rare instances and thus positively influence the feature selection process. In the course of this work, we were able to show that this filtering has a positive effect on the performance of classification models and that outlier detection methods are suitable for this filtering. For some data sets, the resulting increase of performance was only a few percent, but for other datasets, we were able to achieve increases in performance of up to 16 percent. This work should lead to the improvement of the predictive models and the better interpretability of feature selection in the course of the preprocessing phase. In the spirit of open science and to increase transparency within our research field, we have made all our source code and the results of our experiments available in a publicly available repository.

\chapter*{Zusammenfassung}
Damit Machine Learning Algorithmen Wissen aus Rohdaten extrahieren können, müssen diese Daten zunächst bereinigt, transformiert und in maschinengerechte Form gebracht werden. Diese oft sehr zeitaufwändige Phase wird als Preprocessing bezeichnet. Ein wichtiger Schritt in der Preprocessing-Phase ist Feature Selection, die auf eine bessere Leistung von Vorhersagemodellen abzielt, indem die Anzahl der Features eines Datensatzes reduziert wird. In diesen Datensätzen sind die Instanzen verschiedener Ereignisse oft unausgewogen, was bedeutet, dass bestimmte normale Ereignisse überrepräsentiert sind, während andere seltene Ereignisse sehr begrenzt vorkommen. Typischerweise sind diese seltenen Ereignisse von besonderem Interesse, da sie einen höheren Informationsgehalt besitzen als normale Ereignisse. Ziel dieser Arbeit ist es die vorhandenen Instanzen auf diese seltenen Instanzen zu filtern, um dadurch die Feature Selection positiv zu beeinflußen. Im Verlauf dieser Arbeit konnten wir zeigen, dass sich diese Filterung positiv auf die Leistungsfähigkeit von Klassifikationsmodellen auswirkt und dass Ausreißermittlungsverfahren für diese Filterung geeignet sind. Bei einigen Datensätzen betrug die Leistungssteigerung nur wenige Prozent, bei anderen Datensätzen konnten wir jedoch eine Leistungssteigerung von bis zu 16 Prozent erzielen. Diese Arbeit sollte zur Verbesserung der Vorhersagemodelle und zur besseren Interpretierbarkeit der Feature Selection im Verlauf des Preprocessings führen. Im Sinne einer offenen Wissenschaft und zur Erhöhung der Transparenz in unserem Forschungsbereich haben wir unseren gesamten Quellcode und die Ergebnisse unserer Experimente in einem öffentlich zugänglichen Repository verfügbar gemacht.

%\glsresetall %% all glossary entries should be used in long form (again)
%% vim:foldmethod=expr
%% vim:fde=getline(v\:lnum)=~'^%%%%\ .\\+'?'>1'\:'='
%%% Local Variables:
%%% mode: latex
%%% mode: auto-fill
%%% mode: flyspell
%%% eval: (ispell-change-dictionary "en_US")
%%% TeX-master: "main"
%%% End:
              %% Abstract

\tableofcontents                %% this produces the table of contents - you might have guessed :-)

\listoffigures

%% if myaddlistoftodos is set to "true", the current list of open todos is added:
\ifthenelse{\boolean{myaddlistoftodos}}{
  \newpage\listoftodos          %% handy if you are using todonotes with \todo{}
}{}                             %% with todonotes-package option "disable" you can get rid of any todo in the output

%\mainmatter                     %% KOMA: marks main part using arabic page numbers and such; only available in scrbook

%% \input{example-short-chapter}   %% remove this line to get rid of the example chapter
%% \input{chapters-of-expose}      %% remove this line to get rid of the expose chapters
%% \input{chapters-of-thesis}      %% remove this line to get rid of the thesis chapters
%% \input{example-style-chapter}   %% remove this line to get rid of the style chapter

%% include tex file chapters:
%%%% Time-stamp: <2018-08-22 13:18:01 dc>

\chapter{Introduction}
\label{cha:introduction}
\section{Problem and Motivation}
\label{sec:problem_and_motivation}
In order to gain important insights from data, it is not enough to simply feed appropriate algorithms with raw data. 
Rather extensive and expensive data cleansing, data format preparation, and data transformation are required for these algorithms to work effectively. The phase in which all this tedious work is carried out is referred to as preprocessing.\\
One of the most important steps in preprocessing is the selection of variables to be used for later classification or regression modeling. One way to select these variables is to have them manually filtered by domain experts.\\
In our research unit we are mostly dealing with industrial damage detection, for which the data is represented as sensor data recorded over time referred to as time series data \cite{Chandola2009}. An important characteristic of time series data is its high dimensionality, which is a reason why manual filtering of variables by domain experts does not scale. Hence manual filtering is too slow, too expensive, prone to errors, and scales only up to a certain number of variables \cite{Chattopadhyay2014}.\\
The task for selecting the most relevant variables or features of a dataset is referred to as feature selection. The feature space is therefore divided into relevant and irrelevant or redundant features. Irrelevant or redundant features do not provide additional information and can be left out during modeling of a problem \cite{Sabeena2016}. Even worse, irrelevant or redundant features could lead to incorrect predictions and hence have a negative impact on the prediction performance of models \cite{ArauzoAzofra2011}.\\
To select the most relevant features, feature selection depends on the provided instances. However in our data-driven world, where data and available variables are constantly growing, we are able to provide millions or even hundreds of millions of data instances to feature selection algorithms. The vast majority of these instances are represented by normal and non-discriminatory data, while rare and discriminatory cases only account for a small proportion \cite{Chattopadhyay2014}.\\
Many algorithms in machine learning, including feature selection algorithms, are not able to handle these rare cases since they are not adequately represented \cite{Weiss2004}. With instance selection there already exists a research field which has the main goal to select only these rare and most discriminative instances. However, since instance selection methods are typically used to compile training sets for classification or regression models, they have a strong dependency on these models and are not designed for the task of feature selection.\\ 
A quite similar research field to instance selection is outlier detection. The main goal of outlier detection is to find observations which deviate so much from other observations as to arouse suspicious that it was generated by a different mechanism \cite{Ahmed2016}.\\
In order to circumvent the dependency of instance selection methods to models, we apply outlier detection methods to select the most discriminative instances for feature selection. The idea behind our approach is that outlier detection should rank the instances for their discriminative power and should provide only a specific percentage of these ranked instances to the feature selection algorithms. These feature selection algorithms select the relevant features, which are then provided to classification models. The results of these models are then compared to the results of "conventional" feature selection and classification models.
\section{Research Questions}
\label{sec:research_questions}
The previously presented problem can be summarized in the form of our main research question as follows: 
\textbf{Is it possible to positively increase the performance of learning algorithms by providing only a specific subset of our training instances to feature selection algorithms?} 
This main research question has two aspects:
\begin{itemize}
	\item Are outlier detection algorithms suitable for instance selection to influence feature selection algorithms?
	\item Do these filtered feature subsets have a positive impact on the performance of learning algorithms?
\end{itemize}
\section{Structure of Work}
The rest of this work is organized as follows: In order to create the foundation for our approach, we first of all need a basic understanding of the involved topics. Therefore we give an overview of time series analysis, feature selection, instance selection and outlier detection in Chapter \ref{cha:related_work}. 
Then we introduce the method for our state of the art analysis concerning instance selection based on outlier detection methods. Finally this chapter closes with the presentation of results of our state of the art analysis. We then give a detailed insight into the method we have developed in Chapter \ref{cha:method}. In the following chapter \ref{cha:evaluation} we describe in detail the evaluation and the experimental setup to evaluate our developed approach. This work will then be completed by a thorough summary and a outlook for future work in Chapter \ref{cha:conclusions}.        %% this is a suggestion: you have to create this file on demand
%%%% Time-stamp: <2018-08-22 13:18:01 dc>
\chapter{Related Work}
\label{cha:related_work}
\section{Methodology of Literature Review}
\label{subsec:SOTA_methoddology}
In order to create the expose for this work a simplified literature search was carried out. In this literature research, the main goal was to get an overview of the relevant topics, problems and synonyms. This literature research was carried out in free and non-systematic way using the following websites/tools:

\begin{itemize}
    \item google scholar \footnote{\url{https://scholar.google.at}}
    \item wikipedia \footnote{\url{http://en.wikipedia.org}}
    \item open knowledge maps \footnote{\url{https://openknowledgemaps.org}}
\end{itemize}

Based on the obtained results, the respective areas for each of which a separate extended literature research had to be performed, were broken down. As the name of the work suggests, there are three major topics: time series analysis, feature selection, and outlier detection. Since the selection of training instances is a separate research area, namely instance selection, this in total adds up to four main research topics.
The systematic approach was identical for all four major topics. Based on the short literature search we collected the synonyms for each topic. As example we provide the synonyms for time series analysis: 
\begin{itemize}	
	\item Time-series data or Time series data
	\item Time-series prediction or Time series prediction
	\item Time-series modeling or Time series modeling
	\item Time-series data modeling or Time series data modeling
	\item Time-series data mining or Time series data mining
\end{itemize}
Based on the obtained synonyms a search matrix (= combination of the different synonyms) was developed. The entries of the search matrix correspond to the search terms that were used with different parameters in the various platforms. The presented platforms are based on the research on publications of the authors within the computer science and machine learning field:
\begin{itemize}
	\item ScienceDirect \footnote{\url{https://www.sciencedirect.com/search/advanced}}
	\item arXiv \footnote{\url{https://arxiv.org/search/advanced}}
	\item DE Gruyter \footnote{\url{https://www.degruyter.com/dg/advancedsearchpage}}
	\item Scopus \footnote{\url{https://www.scopus.com/search/form.uri?display=basic}}
	\item IEEE Xplore \footnote{\url{https://ieeexplore.ieee.org/search/advsearch.jsp}}
	\item EmeraldInstight \footnote{\url{https://www.emeraldinsight.com/search/advanced}}
	\item SpringerLink \footnote{\url{https://link.springer.com/advanced-search}}
	\item Google Scholar \footnote{\url{https://scholar.google.at}}
	\item Directory of Open Access Journals \footnote{\url{https://doaj.org/search}}
	\item Web of Science \footnote{\url{http://login.webofknowledge.com} (needs account)}
\end{itemize}

For each topic the search terms from the search matrix were combined with general search parameters and platform specific parameters. The two most used general search parameters were the year (from 2018-2008) and the field where the search term should be searched for (mostly we searched in the title and abstract fields). Specific search parameters often included the subject to search in, for example computer science, or the document type for example,  article, review or survey. In order to better understand the literature search, we have listed an example search with results in Section \ref{app:search_results_example} in the Appendix. 
For each topic, the number of matches found per platform, the number of relevant results and the titles of the relevant results were documented in an overview table. The relevant results were further specified in the "relevant table" (for example, type of article, year of publication, filename, abstract,...). Based on the more detailed specification of the relevant results and the abstract, it was determined whether the paper was relevant for our work or not. In order to get the best possible overview of the topic, the focus was on surveys, evaluations, and reviews. The relevant publications were systematically studied including also their references. Together with recommendations from the used platforms \footnote{Recommendations from platforms such as researchgate \url{https://www.researchgate.net/} or mendeley \url{https://www.mendeley.com}} and our colleagues these references formed two very valuable additional cross-platform sources. Especially the recommendations of the used search platforms led to some promising work. For the overview given outlier detection we already had a very broad collection from previous research. These collection contained already 21 surveys, reviews, evaluations, and articles, which was clearly sufficient for a purpose.
\section{Background}
\subsection{Time Series Analysis}
\label{subsec:time_series_analysis}
\textbf{Time series analysis} is a branch of statistics which deals with the analysis of \textbf{time series data} \cite{Fakhrazari2017}. Before we elaborate on the numerous areas and techniques of time series analysis, we first discuss the basic characteristics of time series data.
Time series data has a natural (temporal) ordering. This ordering between different observations basically distinguishes time series data from non-time serial data \cite{Dagum2010}. For time series data, the dimension of time is explicitly taken into account, respectively given the definition according to \cite{Fu2011} time series data:
\begin{itemize}
	\item are collections of chronological observations
	\item are considering data as a whole instead of collection of numerical fields
	\item are ordered over time
	\item consider time as the primary axis
\end{itemize}

We provided an artificially generated example of time series data in the plot shown in Figure \ref{fig:time_series_example_plot}.
\myfig{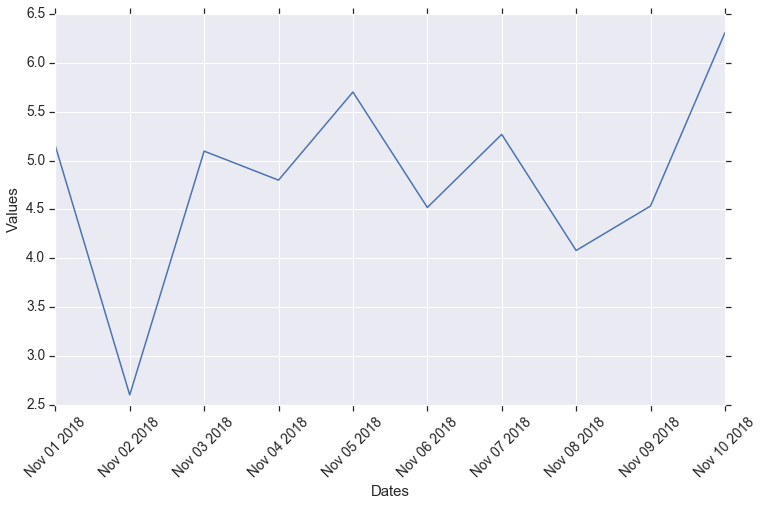}%% filename in figures folder
  {width=0.8\textwidth,height=0.8\textheight}%% maximum width/height, aspect ratio will be kept
  {Figure shows some randomly generated time series data. The x-axis represents the time dimension with days as interval. The y-axis represents the artificially generated values of the different observations.}%% caption
  {Example plot of time series data.}%% optional (short) caption for table of figures
  {fig:time_series_example_plot}%% label

%Real world application examples - Start
Given that a large percentage of the data produced worldwide is time series data and the exponentially growing size of databases, there has recently been an explosion of interest in time series analysis. Among many others the following data from various domains are examples of time series data \cite{Ratanamahatana2010}:
\begin{itemize}
	\item Finance: Presentation of the development of the stock market price of a company over time.
	\item Meteorology: Temperature development over time for a specific area like a country, state, or city.
	\item Trade: Historical store sales data, for example sold products over time.
	\item Medical: Electrocardiograms showing the electrical activity of a heart over time.
\end{itemize}

%Real world application examples - End

Referring to \cite{Fu2011} there are various related research areas concerning time series data, namely finding similar time series, sub-sequence searching in time series, dimensionality reduction, and segmentation.
Generally speaking processing of (time series) data is only expedient if it is processed for the extraction of information and subsequently for the discovery of knowledge. The extraction of information in datasets is referred to as \textbf{data mining} \cite{Fayyad1996}. The equivalent for time series data is consequently denoted \textbf{time series data mining}. \cite{Esling2012} specify the purpose of time series data mining as to extract meaningful knowledge from the \textit{shape} of time series data. Following the definition of \cite{Esling2012} time series data mining involves the following major tasks:
\begin{itemize}
	\item Classification	
	\item Clustering
	\item Motif discovery
	\item Outlier or anomaly detection	
	\item Prediction
	\item Query by content
	\item Segmentation
\end{itemize}

Both \cite{Fu2011} and \cite{Fakhrazari2017} also assign the following topics to the major time series data mining related tasks:
\begin{itemize}
	\item Rule discovery
	\item Summarization
\end{itemize}

In \cite{Fu2011} the authors define pattern discovery the most common mining task, with clustering as the most common method for pattern discovery. Furthermore the authors subsume the tasks anomaly detection, motif discovery and finding discords under the term "pattern discovery".\\
Within the following paragraphs we will provide a brief description for each task related to time series data mining.
%Start ime series tasks description
\textbf{Classification} is the most popular data mining technique. Due to the fact that the classes are determined in advance, classification is also referred to as supervised learning. A classification algorithm assigns these predefined classes represented by so called labels to data instances or within time series domain to time series points or sub-sequences \cite{Ratanamahatana2010}. Since classification of time series data is one of the focus topics of this work, we will go into more detail in section \ref{subsubsec:timeseries_classification}.\\
\textbf{Clustering} algorithms find natural groups, called clusters in data. In contrary to classification there are no predefined labels to mark classes in data, thus clustering is also referred to as unsupervised learning. Clustering aims to organize  instances within a cluster homogeneously (similar instances should belong to the same cluster), but the clusters should be as heterogeneous as possible (different clusters should be as distinct as possible) \cite{Esling2012}.\\
\textbf{Motif discovery} deals with finding recurring patterns in subsequences of time series data. This recurring patterns are referred to as "motifs" \cite{Fakhrazari2017}. \\
\textbf{Outlier detection} focuses on finding abnormal (or anomalous) sequences in time series data. A first step in anomaly detection is often to create a model for detecting normal time series and then find subsequences which deviates from this normal behavior \cite{Fakhrazari2017}. We have a closer look at this topic in Sub-Section \ref{subsec:outlier_detection}.\\
\textbf{Prediction} of subsequent or future values of time series data is based on the principle that observations close together in time are more closely related than observations far away from each other \cite{Dagum2010}. Prediction tasks are modeling these correlations and dependencies between time series data in order to forecast future values \cite{Fakhrazari2017}. Thus this task is also referred to as \textbf{time series forecasting}.\\
\textbf{Query by content} deals with finding of similar time series or time series sub-sequences given a query time series. To define similar time series query by content depends on the definition of a similarity measures between time series data \cite{Esling2012}.\\
\textbf{Rule discovery} is also referred to as association rule mining, and aims at finding relations between variables in (time series) data. \cite{Fakhrazari2017}\\
\textbf{Segmentation} focus on creating approximations of time series data, by means of dimensionality reduction of potential high-dimensional time series data. \cite{Esling2012}. Within the work of \cite{Ratanamahatana2010} segmentation is also referred to as \textbf{summarization}.
%End time series tasks description

%Start time series implementation components
One fundamental problem for almost all of the above tasks of time series data mining is the representation of time series data. A common approach thereby is to transform the time series via some sort of dimensionality reduction followed by various indexing mechanisms. These techniques are complemented by the areas of similarity measures between time series and segmentation of time series data. \cite{Fu2011}. 
These aforementioned steps almost match the definition of \cite{Esling2012}, in which the three major issues for dealing with (high-dimensional) time series data are:
\begin{itemize}
	\item Data representation: Representation or reduction of high-dimensionality data to less dimensions, without changing the basic shape characteristics of the original time series data, for example via sampling or linear regression.
	\item Indexing methods: Organize massive sets of time series data for fast querying, for example via minimum bounding rectangle.
	\item Similarity measurements: Distinguish or match different pairs of time series data, for example via euclidean distance or dynamic time warping \footnote{Dynamic time warping is an algorithm for mapping sequences of different lengths onto each other, for details see reference \cite{Bagnall2014}}.
\end{itemize}
The same authors (\cite{Esling2012}) denote these major issues as "implementation components" which represent the core aspects of time series data systems. \\
Recapitulating \cite{Fu2011} the following two components can also be considered important components concerning time series data:
\begin{itemize}
	\item Segmentation\footnote{According to \cite{Fu2011} segmentation can both be considered as a trend analysis technique and as a preprocessing step for some data mining tasks, which qualifies it also as an implementation component.}: Also referred to as \textbf{summarization}, which performs helpful and necessary discretization of time series data. These techniques span from trivial summarization, such as calculating of summary statistics (for example the mean or variance of windows of time series data) to more sophisticated methods using natural language for summarizing time series data \cite{Ratanamahatana2010} 	 
	\item Visualization: Presents time series data for further analysis to human users, for example via cluster based visualizations.
\end{itemize}
\cite{Fakhrazari2017} denotes representation, similarity measures and the accompanying data mining tasks the three main research orientations in time series data mining.
For a more in-depth look at the underlying components and tasks of time series data mining we can highly recommend the already cited papers: \cite{Ratanamahatana2010}, \cite{Fu2011}, \cite{Esling2012}, and \cite{Fakhrazari2017}.
%End time series implementation components
\subsubsection{Time Series Classification}
\label{subsubsec:timeseries_classification}
\textbf{Time series classification} (TSC) differs from traditional classification in that the elements to be classified are ordered. This ordering may be used for discriminant features \cite{Bagnall2017}. In "traditional" classification this ordering of features is not important, and furthermore interaction between features is considered independent of their relative positions \cite{Bagnall2014}.\\
A possibly more intuitive comparison of traditional classification to TSC is based on the assumption that the former only uses static features, whereas TSC uses dynamic features, for which the change in values over time is relevant.\\
The three main characteristics that make TSC so difficult are: the small number of cases, large number of features and
the highly correlated and/or redundant features. In traditional classification we already have good solutions for these three characteristics. Traditional classification algorithms typical have problems with discriminating features in autocorrelation, phase independence in classes, and embedded discriminative sub-series. Nevertheless, this does not mean that these problems are present in every time series dataset. This qualifies traditional classification algorithms as valuable baseline approaches, and these algorithms may provide important insight into problem characteristics of a specific dataset \cite{Bagnall2017}.\\
Until recently the default baseline algorithm for TSC was the 1-Nearest Neighbor classifier (1-NN) with euclidean distance.
The \textbf{Nearest Neighbor Classifier} is a representative of instance-based learning algorithms. Algorithms of this kind only store training instances during their training phase. To classify a new (unseen) instance this new instance is compared with its closest neighbors within the stored training instances. To compare the closeness to given neighbors instance-based learners are using various different distance functions (probably the most famous one is the euclidean distance function) \cite{Brighton2002}.\\
Since the authors of \cite{Bagnall2014} stated that 1-NN classifier is easy to beat it should not be used as a baseline for TSC any more. Instead the authors recommended the usage of \textbf{1-Nearest Neighbor with dynamic time warping window (DTW1NN)} set through cross-validation as a more meaningful baseline. Furthermore due to the solid results, the authors selected \textbf{Rotation Forest} as their second benchmark algorithm. Rotation Forest is a variant of ensemble learning. Ensemble learners use several (different) base learners, hence the term "ensemble". Within Rotation Forest the base learners are Decision Trees. For each base learner the feature set is randomly split and Principal Component Analysis is performed on each subset. This transformation forms new features for each base learner, since different feature sets will lead to different transformed features and thus different Decision Trees. \cite{Rodriguez2006}
\subsection{Feature Selection}
\label{subsec:feature_selection}
\textbf{Feature selection} is the process of selecting the most relevant features from a dataset. More specific, feature selection should also remove irrelevant and redundant features \cite{Guyon2006}. \\
Based on \cite{Guyon2003}, we use the following terms: \textbf{Variable} refers to the raw input variable and \textbf{feature} refers to the some-how preprocessed or transformed variables. Next to variable the term  "\textbf{attribute}" is used as a synonym for feature. In consequence to these concepts the following terms for feature selection can be found in literature: variable selection, attribute selection and variable or attribute subset selection.\\
Following \cite{Guyon2006} feature selection next to \textbf{feature construction} define the overall concept of \textbf{feature extraction}. Feature construction creates the representation of the data to model, for example defines transformations like standardization, normalization, and discretization.
Another term often used in the environment of feature extraction is "\textbf{feature engineering}". Feature engineering is the process of applying domain knowledge to a problem to obtain the best representation of features that are used by models.\\
Since feature selection is a main part of this work, we only delve into the main concepts for it. The main objectives of feature selection are \cite{Guyon2003}:
\begin{itemize}
	\item improving the prediction performance of predictors
	\item providing faster and more cost-effective predictors
	\item providing a better understanding of the underlying process that generated the data.
\end{itemize}
These objectives result in manifold benefits, such as \cite{Guyon2003}:
\begin{itemize}
	\item facilitating data visualization and data understanding
	\item reducing the measurement and storage requirements
	\item reducing training and utilization times
	\item defying the curse of dimensionality to improve prediction performance
	\item and allow simpler, less complex models.
\end{itemize}

It is worth mentioning that some predictors may turn inefficient or even inapplicable in terms of memory and time if the number of features is too large. Even worse, irrelevant features could confuse some predictors, leading to incorrect predictions \cite{ArauzoAzofra2011}.
\subsubsection{Feature Selection Types}
\label{subsubsec:feature_selection_types}
The three main types of feature selection, namely filter, wrapper and embedded methods are presented an overview in Figure \ref{fig:feature_selection_types}.\\
\myfig{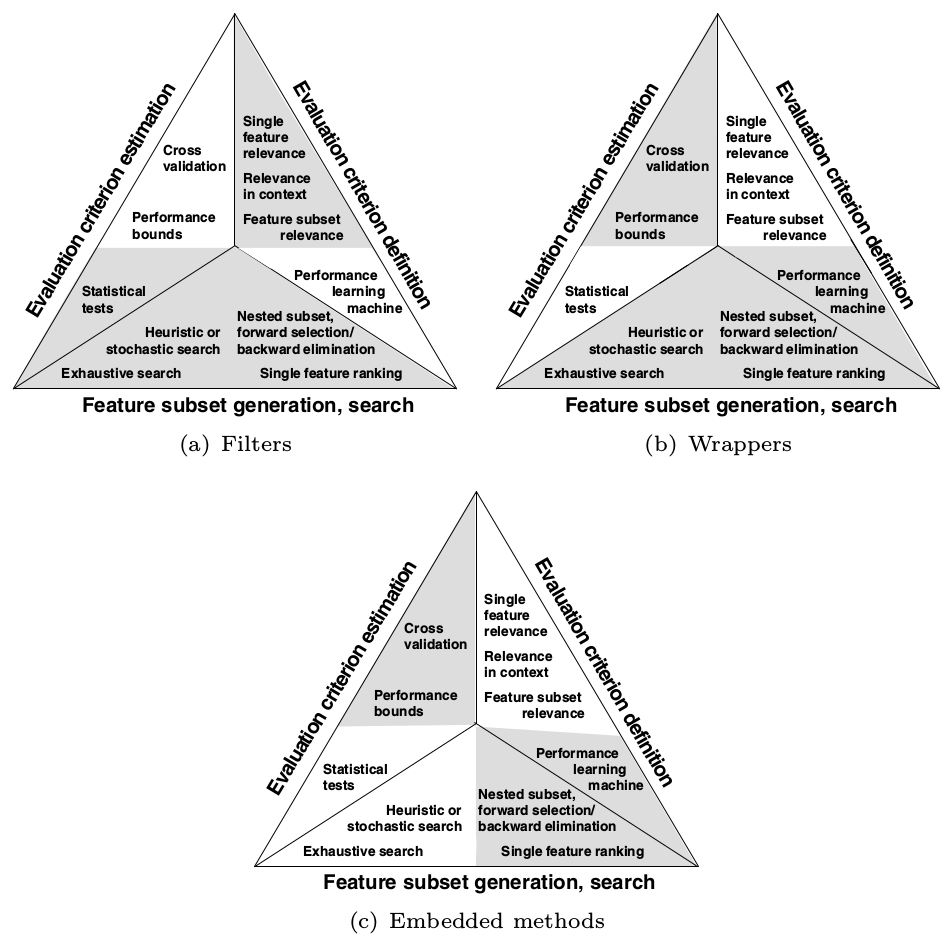}%% filename in figures folder
  {width=1.0\textwidth,height=1.0\textheight}%% maximum width/height, aspect ratio will be kept
  {The tree principal types of feature selection. Gray shading shows the components which the three approaches use. Graphics extracted from \cite{Guyon2006}}%% caption
  {Types of feature selection}%% optional (short) caption for table of figures
  {fig:feature_selection_types}%% label
\textbf{Filter methods} select features without optimizing the performance of a predictor. Mostly filter methods are applied in the form of feature ranking methods which rank features individually (univariate) and take into account their relation to a given target value in regression or labels in classification scenarios \cite{Guyon2003}. 
Examples for feature ranking methods are correlation ranking (for example Pearson or Spearman), Information theoretic ranking criteria (for example Mutual Information) or single feature classifiers (R(i)$^2$ ranking criteria). These methods are complemented by multivariate filter methods like the "Relief" algorithm \cite{Guyon2006}.\\
Since filter methods only have to calculate m number of ranks, where m represents the number of features in a dataset, these methods are considered fast and effective, especially when m is large and the corresponding number of training examples is rather small (for example thousands of features and only hundreds of examples) \cite{Guyon2006}. Additionally filter methods do not depend on specific learning algorithms and therefore avoid over-fitting to given training data \cite{Vergara2014}.\\
One drawback of filter methods is the potential risk that the selected subset is not optimal and that redundant features are selected. Features that are not relevant on their own may be relevant in combination with other features. This could lead to sub-optimal feature sets, but these subsets may be good enough in many cases \cite{Guyon2006}.\\
\textbf{Wrapper methods} are using learning algorithms for the evaluation of feature subsets. The learning algorithm is used as a black-box, for example the evaluation is based on the classification rate of a classification algorithm on a defined test set. This evaluation is performed for each subset which may result in high computational costs and depends on the used learning algorithm especially for high-dimensional datasets \cite{Vergara2014}.\\
Filter and wrapper methods make use of search strategies to search through the feature space. This search through all potential $2^N$ subsets for given data represents a NP-hard problem. This makes the evaluation of all subset inefficient or almost impossible. Often  filter methods are "limited" to feature ranking methods for which the subsets only consist of currently rated feature, although hybrid methods exist, in which filters are used to create feature subsets \cite{Guyon2006}.\\
In contrast to filter and wrapper methods, \textbf{embedded methods} do not separate the learning/training phase of a learning algorithm from the feature selection phase (for example Decision Trees). Embedded methods are less expensive in terms of computational requirements than wrapper methods, but still much slower than filter methods and the selected features strongly depend on the involved learning algorithm \cite{Vergara2014}.\\
Whereas feature selection reduces the features of a given dataset, the main objective of \textbf{instance selection} is the reduction of instances, which is described in the next sub-section.
\subsection{Instance Selection}
\label{subsec:instance_selection}
In classification we need training instances to train a given model to create knowledge which is used to classify new instances. Some of these training instances do not increase or even worse negatively affect our knowledge and therefore are not useful for classification models. The process of selecting only relevant, removing or ignoring useless training instances is known as \textbf{instance selection} \cite{Olvera-Lopez2010}.\\
Although instance selection and feature selection are independent of each other they are often applied jointly to increase the dimensions of data. Instance selection is a subfield in the research area of data reduction. Data sampling for example is considered a data reduction technique but is not an instance selection technique, since sampling randomly reduces data, whereas instance selection involves an intelligent process of categorization of instances, according to a degree of irrelevance or noise \cite{Garcia2015}.\\
Similar to feature selection methods, instance selection methods can be divided regarding the underlying method used for evaluation of instances \cite{Olvera-Lopez2010}:
\begin{itemize}
	\item Filter: Selection criterion uses a selection function independent of a classifier.
	\item Wrapper: Selection criterion is based on evaluation by a classifier.
\end{itemize}
Concerning the types of instance selection methods the literature distinguishes between two different processes which we explain in more detail: \textbf{prototype selection(PS)} and \textbf{training set selection (TSS)}.\\ 
The term "prototype selection" is linked with the advent of instance-based or lazy learning methods \cite{Garcia2015} (see definition of Nearest Neighbor Classifier in subsection \ref{subsubsec:timeseries_classification}). PS are utilizing instance based classifiers to find training sets that offer best classification performance and data reduction rates and thus to perform instance selection.\\
TSS methods conform to the general idea of instance selection since they can be applied to any predictive model (no limitation to instance based classifiers) \cite{Garcia2015}.
Figure \ref{fig:instance_selection_ps} and Figure \ref{fig:instance_selection_ts} illustrates the basic process for both instance selection types.

\myfig{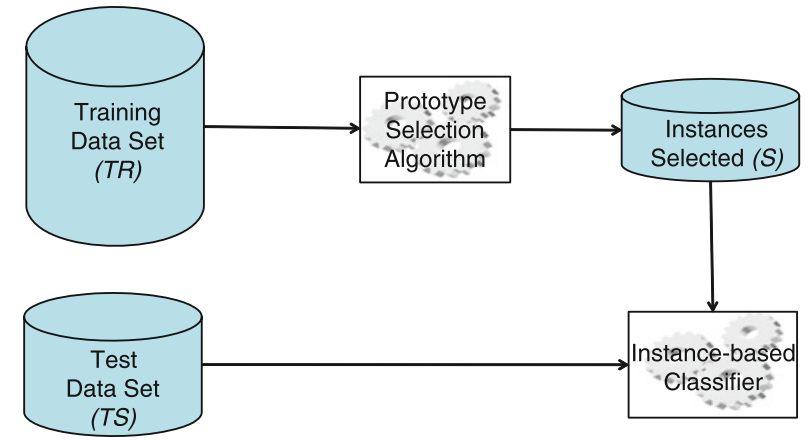}%% filename in figures folder
  {width=0.8\textwidth,height=0.8\textheight}%% maximum width/height, aspect ratio will be kept
  {Process of instance selection based on prototype selection, copied from \cite{Garcia2015}.}%% caption
  {Instance Selection-Prototype Selection.}%% optional (short) caption for table of figures
  {fig:instance_selection_ps}%% label
  
\myfig{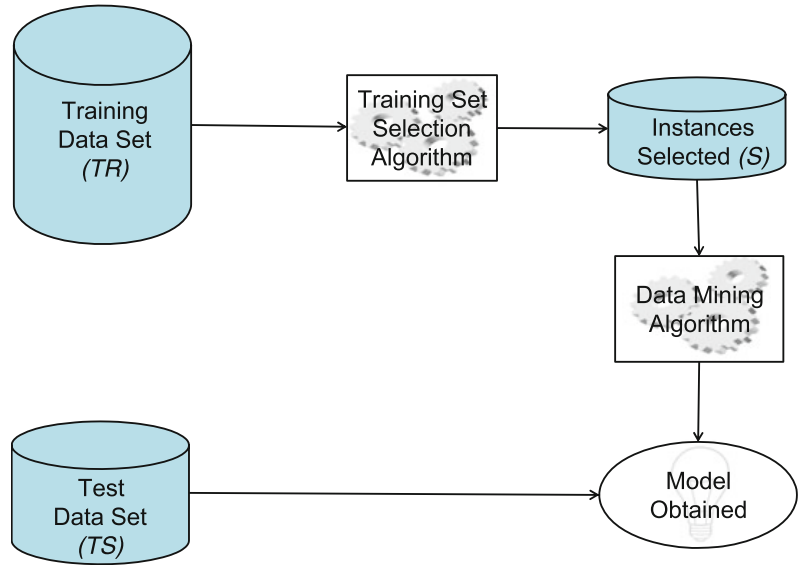}%% filename in figures folder
  {width=0.8\textwidth,height=0.8\textheight}%% maximum width/height, aspect ratio will be kept
  {Process of instance selection based on training subset selection, copied from \cite{Garcia2015}.}%% caption
  {Instance Selection-Training Subset Selection.}%% optional (short) caption for table of figures
  {fig:instance_selection_ts}%% label

Within \cite{Wilson2000} the authors defined criteria to compare different training set reduction algorithms. From our perspective some of these criteria also illustrate the main objectives and functions of instance selection:
\begin{itemize}
	\item Speed increase: Smaller training sizes result in faster training times, or in case of PS methods faster prediction of instances.
	\item Increase generalization accuracy: Size of training set is reduced without reducing of the generalization accuracy, in some case generalization accuracy can even increase with reduction of instances.
	\item Noise tolerance: Removal of certain instances can lead to simple decision boundaries, which could prevent over-fitting, but also could lead to decreasing accuracy.
	 \item Reduction of storage: For PS methods, less training instances require less storage space.
\end{itemize}
A similar topic to instance selection is outlier detection which we focus on in the next sub-section. A significant difference between these two topics is, that instance selection normally operates on already noise free data, whereas outlier detection is also used to remove noisy data.
\subsection{Outlier Detection}
\label{subsec:outlier_detection}
\textbf{Outlier detection} attempts to find patterns in data that do not match the expected normal behavior\cite{Chandola2009}. Following \cite{Hodge2004} "an outlying observation, or \textbf{outlier}, is one that appears to deviate markedly from other members of the sample in which it occurs".\\
\cite{Ahmed2016} defines the main goal of outlier detection as "to find observations which deviate so much from other observations of the same datasets as to arouse suspicious that it was generated by a different mechanism".\\
Often these outliers are the result of exceptional conditions in sensors, measurement equipment, or human errors. 
Within the literature there exists a plethora of synonyms and related topics for outliers and outlier detection. Therefore we want to shed some light on the various terms involved.
Following \cite{Hodge2004} the listed terms are used in the context of outlier detection:
\begin{itemize}
	\item anomaly detection
	\item deviation detection
	\item exception mining
	\item noise detection
	\item novelty detection
\end{itemize}
\cite{Chandola2009} refers to \textbf{novelty detection} as a topic related to outlier detection, which aims at detecting previously unobserved (novel) patterns in data. The main difference is that novel patterns are typically incorporated into the normal model after being detected, in contrast to outliers.\\ 
\cite{Pimentel2014} elaborated a similar delimitation between novelty detection on the one hand and outlier and anomaly detection on the other hand focusing on the different meaning of "normal data". The authors state that anomalies or outliers often refer to irregularities or noisy events in otherwise “normal” data, which has to be removed from data before analysis can be performed.\\
\textbf{Noise detection} or \textbf{noise removal} and \textbf{noise accommodation} are only related topics, which are dealing with unwanted noise in data, which is seen as hindrance in data analysis and needs to be removed or models need to be immunized against their harmful influence.\\
Throughout the rest of this work we use the terms outlier, anomaly and novelty as synonyms, since all terms refer to some kind of disturbance in the underlying process that created the data, which resulted in deviating behavior of an instance. This deviating behavior represents new, relevant and therefore valuable information, that should be incorporated into a model.\\
In contrast to this valuable information, we use the term "noise" for actual errors, which can potentially be harmful for further analysis and thus should be removed from the data. Although the previously listed areas of research differ in some respects, they nevertheless resort to identical techniques for identifying outliers in order to fulfill their problem statement.
\subsubsection{Outlier detection types}
In general we distinguish three different types of outliers, namely point outliers, contextual outliers, collective outliers. We shortly summarize the different types of outliers based on \cite{Chandola2009}.\\
A \textbf{point outlier} is an individual data instance that can be considered as anomalous given the rest of the data. For example, if the typical amount spent in credit card transaction for a customer is hundred Euros, an amount spend of ten thousand Euros is considered an outlier.\\
A \textbf{contextual outlier} is a data instance that is an anomaly only given a specific context, also referred to as conditional outlier. For example, 25 degree Celsius in summer are considered normal, while this is not the case in the mid of December. The applying contexts have to specified as part of the problem formulation. Contextual outliers are often used in spatial or temporal data.\\ 
\textbf{Collective outliers} are a collection of somehow related data instances which are anomalous with respect to the other data instances of a dataset. An individual instance within a collection of outliers may not be considered an outlier, but their occurrence together is considered anomalous. For example, having a very high heart rate for just a few seconds would not be considered as outlier, but having a very high heart rate for hours would be considered a collection of outliers.
\subsubsection{Outlier detection application areas}
Outlier detection is applied in many different areas. \cite{Chandola2009} provide an extensive overview of application areas for outlier detection. We only present a brief summary of the different application areas:
\begin{itemize}
	\item Intrusion Detection: Detection of malicious activity in computer or computer network related systems
	\item Fraud Detection: Detection of criminal activities in banks, insurances, stock market, credit cart companies
	\item Medical and Public Health Anomaly Detection: Detection of anomalies concerning patient conditions, or detecting disease outbreaks
	\item Industrial Damage Detection: Detection of industrial machines and products to prevent escalation and losses due to problems in production
	\item Image Processing: Detection of changes in images over time, or abnormal regions in static images
	\item Anomaly Detection in Text Data:  Detection of novel topics, events or news in collections of documents
	\item Sensor Networks: Detection of anomalies, such as faulty sensors, or abnormal events in wireless sensor networks
\end{itemize}
There are various other domains, for which the interested reader may refer directly to \cite{Chandola2009}.
\subsubsection{Outlier detection categorization}
\cite{Hodge2004} offers a categorization of different methods, based on research areas in which they have been developed. This classification includes the following areas: statistic techniques, neural networks techniques, machine learning techniques, and hybrid systems.
In the very same work the authors presented three fundamental approaches concerning the problem of outlier detection based on the underlying approaches: 
\begin{itemize}
	\item Type 1: Determine outliers with no prior knowledge of the data analogous to unsupervised clustering methods
	\item Type 2: Model both normality and abnormality, analogous to supervised classification
	\item Type 3: Model only normality or in a very few cases model abnormality, which is in fact novelty detection, where soft bounded algorithms can estimate the degree of "outlierness".
\end{itemize}
An extension of these approaches from \cite{Chandola2009} provides a more comprehensive list of approaches based on the underlying techniques involved:
\begin{itemize}
	\item Classification based
	\item Clustering based
	\item Nearest Neighbor based
	\item Statistical
	\item Information Theoretic
	\item Spectral
\end{itemize}
We briefly summarize these approaches within the next paragraphs based on \cite{Chandola2009}.\\
\textbf{Classification based} approaches work analogous to classification: Within a training phase a classifier is trained with labeled trained data, and within the test phase an instance is classified as normal or outlier instance.\\
\textbf{Clustering based} techniques are based on the paradigm that unlabeled instances are assigned to clusters. There are different clustering based approach based on the underlying assumptions concerning outliers. The first category considers an instance as an outlier if it is not assigned to a cluster. The second category considers instances which are far away from their cluster-centroids as outliers. And the third category is based on the assumption that normal instances belong to large and dense clusters, whereas outliers belong to small and sparse clusters.\\
\textbf{Nearest Neighbor Based} techniques are based on the assumption that normal instances occur in dense neighborhoods, while anomalies are located in sparse areas, or are far away from their nearest neighbor.\\
\textbf{Statistical} techniques are analyzing the location of instances given the probability regions, where normal instances are located in high probability regions and outliers occur in low probability regions.\\
\textbf{Information Theoretic} techniques are dealing with the information content by using theoretic measures, for example entropy, or mutual information. The key assumption of these techniques is that outliers change the information content of a dataset.\\
\textbf{Spectral} techniques enforce change in data representation, for example dimensionality reduction, to embed data into lower dimensional subspace in which normal instances are significantly different then outliers.\\

Another categorization of approaches for outlier detection arouses if the different approaches are categorized based on the underlying type of algorithm \cite{Hodge2004}:
\begin{itemize}
	\item distance-based
	\item set-based
	\item density-based
	\item depth-based
	\item model-based
	\item graph-based
\end{itemize}
\section{State of the Art}
\label{sec:SOTA}
\subsection{State of the Art Analysis}
For the background related to instance selection we had 11 recommendations and found 2 references in this recommendations. Out of this 13 publications we finally used 5 publications for our background section and one publication is used as the reference application for a state of the art instance selection algorithm (see LDIS algorithm in Section \ref{subsec:algos_is}).\\
In order to find outlier detection algorithms that were used for instance selection, we performed a state of the art (SOTA) analysis. We used the following search terms and combinations in each of the earlier presented platforms (see Section \ref{subsec:SOTA_methoddology}):
\begin{itemize}
	\item "outlier$\vert$novelty$\vert$anomaly detection" AND "instance selection"
	\item "outlier$\vert$novelty$\vert$anomaly detection" AND "prototype selection"
	\item "outlier$\vert$novelty$\vert$anomaly detection" AND "training set $\vert$ trainingset selection"
\end{itemize}
These searches resulted in 28 papers which we considered relevant for closer inspection. After closer inspection 22 papers where considered not relevant for the following reasons:
\begin{itemize}
	\item Six papers were pure instance selection papers with no relation to outlier detection
	\item One paper simply applied instance selection before outlier detection
	\item One paper applied outlier detection in adaptive protocols without any instance selection
	\item Fourteen papers used totally different approaches, or had no connection of outlier detection and instance selection:
		\begin{itemize}
			\item Three papers were used for meta learning, multiple instance learning or used mutual information selection for forecasting
			\item Four papers applied noise removal instead of instance selection
			\item Two papers applied sampling techniques for instance selection			
			\item Five papers applied preprocessing to streaming data
		\end{itemize}
\end{itemize}
We provide a detailed table of these 22 irrelevant papers in Section \ref{sec:sota_irrelevant_results} in the Appendix.\\
The 28 search results also included three papers that looked at the topic from the perspective of how to use instance selection for outlier detection, see \cite{Li2011}, \cite{Li2009Exp}, \cite{Li2009Opt}.\\
To give an example for a search result which was not suitable for our approach, we will briefly describe the following work: \cite{Vinh2017}. The authors proposed a novel approach for instance selection which consisted of two steps. First unrepresentative (or outlier) instances are removed from the training set using "data editing" (a variant of instance selection)) and then the authors perform "instance reduction" based on compression via minimum description length principle. Although this approach involves all the relevant topics it does not perform outlier detection for instance selection. \\
Finally this results in only two relevant results of our SOTA analysis which we will inspect in more detail in the following sub-section.
\subsection{Outlier Detection for Instance Selection}
\label{subsec:sota_od_for_is}
The most promising paper of our SOTA analysis was \cite{Peng2014}, which is closely related to \cite{Peng2012} since the authors are identical. The authors state that instance selection can be carried out via outlier detection techniques. Their approach is based on local kernel regression, and accounts for the local structure of the data space, by not only considering the distance between a point and its neighbors, but also the distance between neighbors. The basic principle of the underlying work is that the reconstruction error of an inner point is smaller than that of a boundary (or outer) point. Based on this phenomenon the "outlierness" of points is calculated. To get the estimates of "outlierness" it is measured for each point $X_i$ how well its neighbors can estimate every feature value of $X_i$. For a clearer understanding of the used algorithm we depicted it in Listing \ref{alg:lkrr}.

\begin{algorithm}
    \caption{Local Kernel Ridge Regression}
    \label{alg:lkrr}
    \begin{algorithmic}
    	\STATE Provide \textbf{l} (number of outliers to be removed)
		\STATE \textbf{m} = 0 (number of outliers removed)
		\STATE \textbf{O} = Empty list (list of removed outliers)
		\STATE Create neighbor graph \textbf{G} and kernel matrix \textbf{K} from dataset \textbf{D}
		\WHILE{m != l} 
			\STATE Estimate local kernel regression for each instance \textbf{$X_i$}
			\STATE Calculate reconstruction error \textbf{$RE_i$} for each \textbf{$X_i$}
			\STATE Sort all \textbf{$X_i$} based on \textbf{$RE_i$} in descending order
			\STATE Insert {$X_i$} with highest {$RE_i$} into \textbf{O}
			\STATE Remove {$X_i$} with highest {$RE_i$} from \textbf{D}
			\STATE \textbf{m} = \textbf{m} + 1
			\STATE Re-create \textbf{G} and \textbf{K} from \textbf{D}
		\ENDWHILE
    \end{algorithmic}
\end{algorithm}

In addition to the overall algorithm, we briefly explain the most essential steps of estimation and calculation of the reconstruction error. Since the estimation is carried out using local kernel ridge regression, we briefly explain, the basic concept of it, keeping the notation used in \cite{Peng2014}. Kernel ridge regression uses kernel functions to model nonlinear regression. Regression in general is given by training data $(x_i, y_i)_{i=1}^N$, where $x_i$ corresponds to the input and $y_i$ to the target value, and estimates the targets by the given inputs. Kernel ridge regression can be defined as:
\begin{equation}
	g(x) = \sum_{i=1}^{N} \alpha_i \mathcal{K}(x, x_i), 
\end{equation}
where $\mathcal{K}(.,.)$ is a kernel function and $\alpha_i$ are the coefficients, which are estimated by the following objective function:
\begin{equation}\label{equ:ojbective_function}
	\underset{\alpha}{min}||K\alpha-y||^2 + \gamma \alpha^T K \alpha,
\end{equation}
where $\alpha=[\alpha_1,...,\alpha_N]^T$, $y=[y_1, ..., y_N]^T$, $\gamma$ is a small positive regularization parameter and K is the kernel matrix with dimensions  N$\times$N with $k_{i,j}=\mathcal{K}(x_i, x_j)$.
The solution of \ref{equ:ojbective_function} is than given by:
\begin{equation}
	\alpha=(K + \gamma I)^{-1} y,
\end{equation}
where I is an N$\times$N identity matrix. Finally g(x) can be given as
\begin{equation}
	g(x) = k_x^T(K + \gamma I){^-1} y,
\end{equation}
where $k_x=[\mathcal{K}(x, x_1), ..., \mathcal{K}(x, x_N)]^T$ is a vector containing the kernel functions for instance x and $x_1$ to $x_N$.\\
To estimate the value of the r-th feature of an instance $x_i$ by the corresponding r-th feature values of the neighbors of $x_i$ \textbf{local kernel regression} is used. More formally, given a training instance $x_i$ and
${x_j, l_{rj}}_{x_j \epsilon N_i}$, where $N_i$ are the neighbors of $x_i$, $x_j$ is on of this $x_i$ and $l_{rj}$ is the r-th feature of this neighbor. The estimation of $l_{ri}$ with the local kernel ridge regression model for $x_i$ is than given by:
\begin{equation}
	g_{N_i}(l_{ri}) = k_{N_i}^T(K_{N_i} + \gamma I)^{-1} l_{rN_i}
\end{equation}\\
where $k_{N_i}^T$ is a vector containing the kernel function for $x_i$ and its neighbors.
$K_{N_i}$ is the kernel matrix for the kernel functions between the neighbors of $x_i$ $N_i$, for example $K_{N_i} = [\mathcal{K}(x_m, x_n)], x_m, x_n \epsilon N_i$ and I corresponds to an N*N identity matrix.\\
What is missing is the explanation of the calculation of the \textbf{reconstruction error}, which is given by the square of the difference between the estimated value of the r-th feature of the local kernel ridge regression model and the real value of $l_{ri}$, or as equation:
\begin{equation}
	Re(l_{ri}) = (g_{N_i}(l_{ri}) - l_{ri})^2
\end{equation}\\
Since the values for each feature may be different, we need to normalize $Re(l_{ri})$. We therefore calculate the diagonal matrix \textbf{D} $diag(D_{11}, ..., D_{kk}))$ which is given by:
\begin{equation}
D_{ii} = \frac{1}{|N_i|} \sum_{x_j \epsilon N_i} (x_i - x_j)^2
\end{equation}
Additionally a weighted variance $V_w(r)$ for each feature is estimated, as:
\begin{equation}\label{eq:weighted_variance}
	V_w(r) = \sum_{i=1}^{N} (l_{ri}-\mu_r)^2 D_{ii} 
\end{equation}
In Equation \ref{eq:weighted_variance} $\mu_r$ is the weighted mean of the r-th feature and is given by: 
\begin{equation}
	\mu_r = \sum_{i} l_{ri} \frac{D_{ii}}{\Sigma_i D_{ii}}
\end{equation}
Finally the normalized reconstruction error $RE_i$ is given by:
\begin{equation}
	RE_i = \sum_{r=1}^{M} \frac{Re(l_{ri})}{V_w(r)}
\end{equation}          %% this is a suggestion: you have to create this file on demand
\chapter{Method}
\label{cha:method}
\section{Basic Principle of our Approach}
\label{sec:focus}
To introduce our approach we consider it very helpful to first introduce the general or common approach to feature selection in combination with classification (also applicable to regression problems). This general approach is depicted in Figure \ref{fig:principle_general}. The typical sequence starts by dividing available data into disjunct training and test sets. The training set is then provided to a specific feature selection algorithm, which filters the training set for the most relevant features. The training set with only the selected features is then provided to a classification algorithm which builds a model to predict the output for the given input. After the model is build the test set is also filtered for the only selected features and this filtered test set is used to evaluate the performance of the trained model.

\myfig{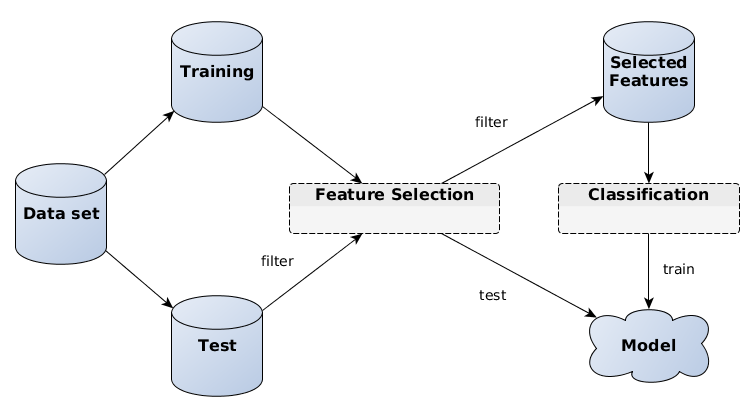}%% filename in figures folder
  {width=1.0\textwidth,height=1.0\textheight}%% maximum width/height, aspect ratio will be kept
  {Figure shows the general approach and sequence for feature selection combined with classification.}%% caption
  {Sequence feature selection and classification}%% optional (short) caption for table of figures
  {fig:principle_general}%% label
  
The difference of our approach is that we filter the instances of the training set for only "relevant" instances before making these selected instances available to a feature selection algorithm. We present our approach in Figure  \ref{fig:principle_approach}. Due to the pre-selection of training instances, this should positively influence the feature selection algorithm (for example, other features are considered relevant as with the generic approach).
  
\myfig{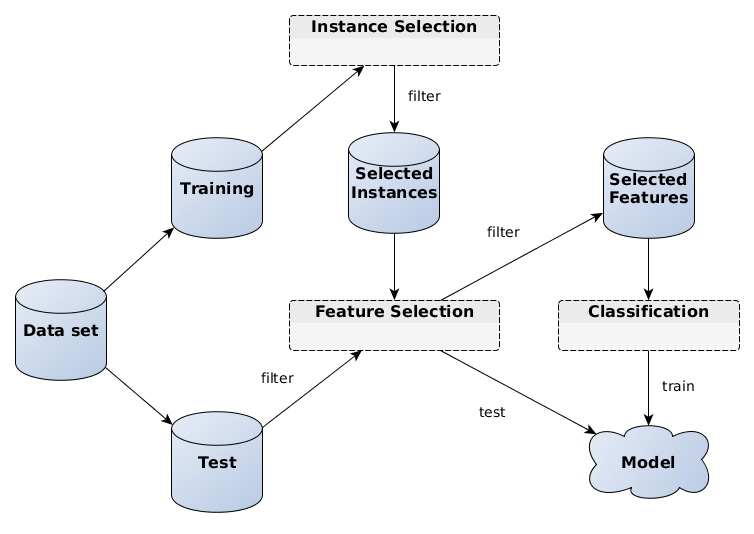}%% filename in figures folder
  {width=1.0\textwidth,height=1.0\textheight}%% maximum width/height, aspect ratio will be kept
  {Figure shows the general approach and sequence for instance selection for feature selection combined with classification.}%% caption
  {Sequence instance selection for feature selection and classification}%% optional (short) caption for table of figures
  {fig:principle_approach}%% label
  
The selection of the most relevant instances in the training set requires an algorithm that assigns a specific score to each training instance. Based on these scores, the training instances are sorted and only the top n instances are selected. We further describe this concept in more detail in the following section.
\section{Concepts}
\label{sec:concepts}
For the instance selection we are interested in discriminant and thus relevant instances, as we expect a more specific selection of features. To achieve this goal we are guided by the simple but elegant principle of nearest neighbor based outlier detection techniques: Normal data instances occur in very dense neighborhoods, whereas outliers occur in sparse neighborhoods \cite{Chandola2009}. If we transfer the principle of density to that of distances, we can simplify this principle by imagining that an outlier is farther away from his nearest neighbors than a normal instance that has many neighbors within a short range. This difference in neighborhood is also visualized in Figure \ref{fig:outlier}.

\myfig{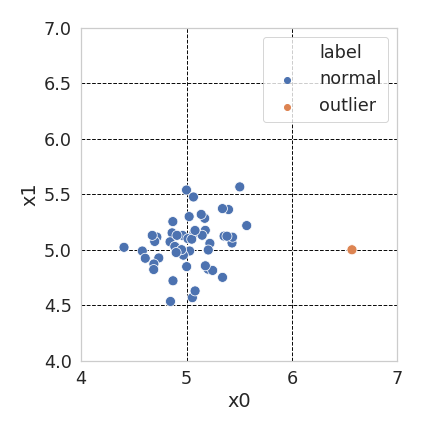}%% filename in figures folder
  {width=0.6\textwidth,height=0.6\textheight}%% maximum width/height, aspect ratio will be kept
  {Figure shows a dense cluster of normal points with low distances to the nearest neighbor for the normal instances and a high distance for the outlier on the right.}%% caption
  {Normal dense cluster compared to an outlier}%% optional (short) caption for table of figures
  {fig:outlier}%% label

This distinction due to the density or distance within a neighborhood represents a direct link between instance selection and outlier detection research areas.
In order to implement our approach we have to adapt the algorithm presented in Section \ref{subsec:sota_od_for_is} as follows: Rather than performing the main loop "m-times" (for removing m outliers) we only need to calculate the total reconstruction error {$RE_i$} for each instance once, sort all instances based on their {$RE_i$} values and then select the top n instances to filter our training instances for the following feature selection. Since we have a criterion independent of the evaluation of a classifier, this approach is a typical filter based instance selection approach. Further, as we focus on selecting training instances (for feature selection though), our approach is a typical training set selection algorithm.\\

For further investigation how instance selection affects the feature selection and consequently the performance of the classification, we first designed our experiments. For this first orientation we used the Waikato Environment for Knowledge Analysis, short \textbf{Weka} \cite{Frank2016}. Weka is a machine learning suite written in Java, which can be used with a graphical user interface (gui) and also programmatically via a well documented API \footnote{\url{http://weka.sourceforge.net/doc.stable-3-8/}}. After the first manual experiments via the Weka gui it soon became clear that due to the many different components and their parameters, manual experiments were by no means an option. So we decided to build our own pipeline that allowed us to do a large number of different experiments in a systematic way. We were able to draw on a basic concept, which we had developed in the course of another research project \cite{Stanisavljevic2019}. Due to changed requirements we had to adapt the pipeline components but not the two main underlying principles. The first main principle we had to follow was modularity. In doing so, we wanted to ensure that the different components (for example, outlier detection and feature selection) could exchange information in a defined manner without creating specific dependencies between these different components. Firstly, it called for a so-called context with which information could be passed on from one component to the next, and second, general methods for uniformly accessing these components. These issues were addressed with the general components \texttt{PipelineContext} and the interface \texttt{PipelineStep} (see details in section \ref{subsubsec:pipeline_components}.\\
The second main principle we were following was interchangeability. By following this principle we wanted to guarantee that the different variants of components (for example different algorithms used for feature selection) are independent from all other steps and are completely interchangeable. We have also implemented this principle with the general \texttt{PipelineStep} principle, but also with so-called wrapper classes, which are also explained in more detail in Subsection \ref{subsubsec:wrapper_classes}.

\myfig{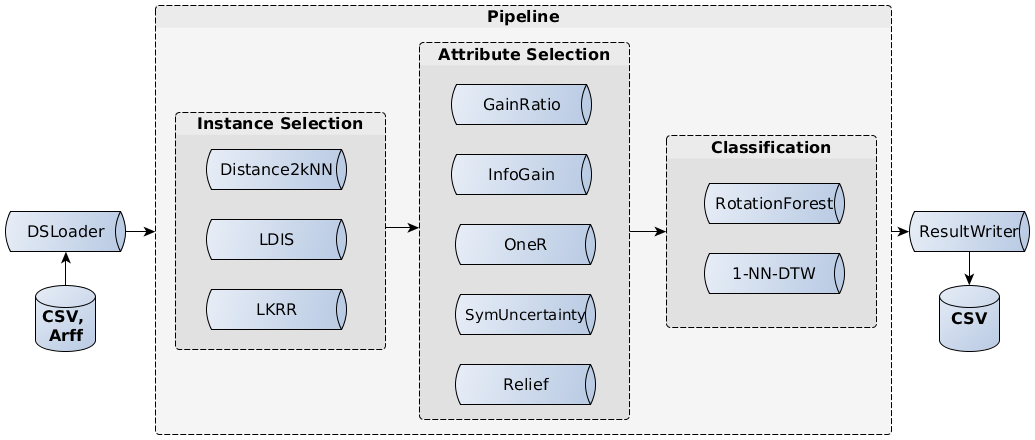}%% filename in figures folder
  {width=1.0\textwidth,height=1.0\textheight}%% maximum width/height, aspect ratio will be kept
  {The different component types with its specific instances of our pipeline.}%% caption
  {Components of developed pipeline}%% optional (short) caption for table of figures
  {fig:pipeline}%% label
\section{Implementation}
\label{sec:implementation}
In this section we describe all components and their parameters of our developed pipeline. We would like to point out that the default terminus for feature in Weka is "attribute". The implication of this is that all feature selection components in Weka are consequently referred to as "attribute selectors". In order to avoid confusion we will also use the term "attribute selection" when it comes to components of Weka or our developed pipeline. In the final Subsection \ref{subsec:install_and_execution} of this chapter we also explain the requirements and the instruction for installing and executing of our pipeline.
\subsection{Components and Parameters}
\label{subsec:comp_and_params}
\subsubsection{Pipeline Components}
\label{subsubsec:pipeline_components}
\texttt{PipelineStep} represents a simple interface which only defines the two functions \texttt{performStep} and \texttt{printPipelineStepDetails()}, where the later only defines a method for logging the basic members for this \texttt{pipelineStep}. The \texttt{performStep} function is the main entry point to perform a defined pipeline step. To load and store data the only parameter is the \texttt{PipelineContext} of the current pipeline.\\
A \texttt{PipelineContext} is used to exchange information between different \texttt{PipelineStep}s, for example an algorithm which is used for instance selection stores these instances in the current \texttt{PipelineContext}. These instances are later loaded from the following attribute selection \texttt{PipelineStep}. The \texttt{PipelineContext} provides several generic structures for the administration and exchange of different instances, parameters, and results.\\
These two components are complemented by the \texttt{Pipeline} object. This is the wrapper class for a pre-defined pipeline consisting of a list of different \texttt{PipelineStep}s and a \texttt{PipelineContext}.
\subsubsection{General Components}
\label{subsubsec:general_components}
\texttt{DSLoader} is the main class for loading the data sets used for our experiments. We use the convention that the data files use the same prefix as the \texttt{datasetName} plus two specified prefix for the training and testing set and a corresponding file ending (for example ".csv" for csv files). After providing the fully qualified folder where the data set is stored under the provided \texttt{datasetName} to the constructor, the training or test data sets can be loaded via the corresponding (\texttt{loadTrainingDataset()}, \texttt{loadTestDataset()}) functions.\\
\texttt{AbstractPipelineCsvResult} is the main abstract class for all csv results and unifies functions that are applicable generally, regardless of the type of result (for example the general structure for storing fields and their corresponding values). Within our pipeline we have the following different types of csv results:
\begin{itemize}
	\item \textbf{\texttt{ClassificationCSVResult}} stores the result of classification experiments (for example, the name of the used classifier, precision and recall)
	\item \textbf{\texttt{AttributeSelectorCSVResult}} stores the result of attribute selection experiments (for example, the name of the used \texttt{AttributeSelector}, number of selected attributes)
	\item \textbf{\texttt{OutlierDetectionCSVResult}} stores the result of outlier detection experiments (for example, the name of the algorithm used, number of instances selected)
\end{itemize}

These aforementioned results can be written to output files by using the \texttt{AbstractPipelineCsvResultWriter}. This class has only one method, \\ \texttt{writeOutputFileFromResultList()} which writes the given \\ \texttt{AbstractPipelineCsvResult}s to the given \texttt{filePath}. 
Our pipeline uses a few auxiliary classes (for example \\ \texttt{InstanceUtils}). For the description of these additional classes we would like to refer to the corresponding documentation in the source files of our git-repository.
\subsubsection{Wrapper Classes}
\label{subsubsec:wrapper_classes}
In principle, Weka already offers corresponding classes and functions for almost all requirements for our approach. To integrate these existing classes into our pipeline and better manage the numerous parameters, we had to pack these existing classes into so-called wrapper classes. This wrapping allows us to continue to comply with our main principles modularity and interchangeability mentioned in Section \ref{sec:concepts}. We briefly describe the developed wrappers in the following sub-section.\\
The \texttt{AbstractNNWrapper} is the main class for instance and outlier detection algorithms. Within this class the main members and functions for all the sub-classes are implemented. These sub-classes share the following relevant parameters mapped as instance variables:
\begin{itemize}
	\item \texttt{kNeigbours}: Number of k-nearest neighbors used for search and distance calculations.
	\item \texttt{odSelRatios}: Ratio which represents the percentage of instances to finally select (for example 0.1 means that 10 percent of instances are selected) 
	\item \texttt{odDescOrder}: Whether to use descending order for sorting the instances (for example, true means use that descending order is used and false means that ascending order is used)
	\item \texttt{nnSearch}: Always \texttt{LinearSearch}, since Weka only supports all distance functions \footnote{We will have a closer look at these distance functions in Section \ref{subsec:param_experiments}} only for the \texttt{LinearSearch} at the moment.
\end{itemize}
Furthermore \texttt{AbstractNNODWrapper} implements the \texttt{PipelineStep} function \texttt{performStep}. Within this function the sub-class specific method \\ \texttt{performFiltering} is called, data is loaded and afterwards stored in the \texttt{PipelineContext}. The concrete sub-classes are listed bellow, and further description for them is available in Section \ref{subsec:algos_is}:
\begin{itemize}
	\item Distance2kNNODWrapper
	\item LDISWrapper
	\item LKRRODWrapper
\end{itemize}
\texttt{AttributeSelectorWrapper} is the wrapper class for the attribute selection algorithms. The concrete attribute selection algorithms are further described in Section \ref{subsec:feature_selectors}. Based on the member variable \texttt{attributeRatio} the given ratio of attributes from the training data set are selected (for example 0.1 means that 10 percent of the attributes are selected). The actual number of attributes to select is calculated via the method \texttt{getNumToSelectFromRatio} and takes the number of attributes without the class attribute as argument. In case \texttt{getNumToSelectFromRatio} returns 0, we override the value to one, since we always want to have at least one best attribute to be selected for further processing. Once the attributes are evaluated the training and test instances are reduced to the given \texttt{attributeRatio} of attributes. These instances are then stored within the \texttt{PipelineContext} which are then used by the classification algorithms.\\
\texttt{AbstractClassifierWrapper} has five concrete classification algorithms as sub-classes which can be used in our pipeline. For the experiments of this work we only used \texttt{Rotation Forest} and \texttt{DTW1NN}. \texttt{AbstractClassifierWrapper} has no relevant members and parameters, Rotation Forest neither, DTW1NN has only the fixed default window size for dynamic time warping which is one by default.
\subsection{Installation and Execution of our Pipeline}
\label{subsec:install_and_execution}
We first want to point out that we provide all our source-code, data and results in a git repository \footnote{\url{https://git.know-center.tugraz.at/summary/?r=~dcemernek/outfisltisl.git}}. Within the root folder of this repository we also provided a \texttt{README} file which contains an overview of our library. Our framework \textbf{Outfisltisl} has the following requirements:
\begin{itemize}
	\item Java Oracle JDK version 1.8.X
	\item Maven version 3.3.X \footnote{Maven is a tool to build source code and manage dependencies}
\end{itemize}
Further dependencies are listed in the corresponding \texttt{pom.xml} in the root folder of our project.
For the reference timeseries classifiers we used the source-code from the paper \cite{Bagnall2014} \footnote{see also: \url{https://bitbucket.org/TonyBagnall/time-series-classification}}. From this source code we extracted the DTW1NN and Rotation Forest classifiers and build the jar file \texttt{timeseries-classification-1.0.0.jar} which is present in the \texttt{lib} folder of our project. The specified commands refer to the execution in a Linux environment.
For the installation of our pipeline the following steps are necessary: 
\begin{itemize}
	\item Requirements from above fulfilled
	\item Clone git-repository: \\\texttt{https://git.know-center.tugraz.at//r/~dcemernek/outfisltisl.git}
	\item Execute the following commands in root folder (= outfisltisl) of project:
	\begin{itemize}
		\item \texttt{mvn initialize} (required only once)
		\item \texttt{mvn package} (each time changes are made within source code)	
	\end{itemize}
\item After execution of \texttt{mvn package} there should be a folder "target"
\item Folder "target" should contain jar file "outfisltisl-jar-with-dependencies.jar"
\end{itemize}
For the execution of the experiments the following steps must be executed:
\begin{itemize}
	\item Requirements: Installation from above fulfilled
	\item Again given commands refer to a Linux environment
	\item Application can be started with command: \\\texttt{java -jar outfisltisl-jar-with-dependencies.jar}
	\item This command also prints the Help information including relevant parameters to use.
	\item Example command that starts the outlier detection experiments for dataset FordA in folder "../data/datasetsExp" using Rotation Forest Classifier and LKRR as instance selection algorithm: \\
  \texttt{java -jar outfisltisl-jar-with-dependencies.jar -i ../data/datasetsExp/ -d FordA -e od -c RotF -o LKRR}
\end{itemize}              %% this is a suggestion: you have to create this file on demand
\chapter{Evaluation}
\label{cha:evaluation}
\section{Recap}
As a quick refresher, we briefly review our main research question: Is it possible to positively increase the performance of learning algorithms by providing only a specific subset of our training instances to feature selection algorithms? In order to evaluate this question and thus our approach we need to define and describe the following components in this section:
\begin{itemize}
	\item Datasets to test our approach
	\item An outlier detection algorithm to perform instance selection to filter training instances
	\item Additional instance selection algorithms to compare our approach to
	\item Feature selection methods to perform feature selection on our filtered training instances
	\item Classification algorithms to measure potential performance influence of our approach
\end{itemize}
Before we give a detailed description of the above components in Section \ref{sec:experimental_design}, we start with a general introduction on how to evaluate time series classification.
\section{Evaluation of Time Series Classification}
\label{sec:evaluation_tsc}
For the current state of the art concerning the evaluation of time series classification we will focus on the following work "The great time series classification bake off: a review and experimental evaluation of recent algorithmic advances" \cite{Bagnall2017}. In this paper the authors implemented 18 recently proposed algorithms and compared them against two benchmark classifiers. In this evaluation they authors found out that only 9 of this 18 algorithms were significantly more accurate than the benchmark classifiers. They authors published all used datasets and detail results on a dedicated website, namely the UEA \& UCR Time Series Classification Repository\footnote{\url{http://timeseriesclassification.com}}. Since the datasets, algorithms and results are constantly extended we decided to use the same standard benchmark classifiers and datasets from this site. Furthermore by only using public available datasets we want to support transparency within our research field.\\
The second piece of work we use to evaluate our approach is the somewhat older but still very relevant paper named "On the Need for Time Series Data Mining Benchmarks: A Survey and Empirical Demonstration" \cite{Keogh2003}. In this paper the authors conducted a survey about time series data mining papers and analyzed their evaluation, underlined by the following relevant statistics per paper:
\begin{itemize}
	\item Size of test datasets - Median only 10,000 test data
	\item Number of rival methods - Median number is 1 so authors only compared to one method
	\item Number of different test datasets - On average each contribution on 1.85 datasets
\end{itemize}
Since we are mostly dealing with Industrial Damage Detection, which involves sensor data, we focus on datasets of this type. The largest dataset for sensor data has only 9236 instances, we can not address the first point, but the authors of \cite{Bagnall2017} already addressed the importance of a better representation of larger datasets in the future.\\
For the number of rival methods we refer to the corresponding Sub-section \ref{subsec:algos_is}, we only state that we evaluated a plethora of different combinations of outlier detection, feature selection and time series classification algorithms.\\
Concerning the number of different test datasets we can clearly state that by using 8 different sets from the UEA \& UCR Time Series Classification Repository in total. The datasets and all other components of our experiments are described in the following section.
\section{Experimental Design}
\label{sec:experimental_design}
\subsection{Datasets}
\label{subsec:data_sets}
For evaluation of our contribution we use datasets from the UEA \& UCR Time Series Classification Repository relaunched 2018 with 128 datasets \cite{UCRTSA2018}.
For this contribution we stick to the following criteria to select datasets: 
\begin{itemize}
	\item Type of dataset: Sensor, since Industrial Damage Detection mostly deals with this kind of data
	\item Number of classes: Two, since most of the time we need to distinguish between normal and abnormal data
	\item Evaluation results for comparison available on UEA \& UCR TSC website
\end{itemize}
These criteria resulted in the datasets described in Table \ref{table:data_sets}. Since the following datasets have no results for comparison we could not include them: DodgerLoopGame, DodgerLoopWeekend, FreezerRegularTrain, FreezerSmallTrain. Additionally we excluded the FordB dataset since it is too similar to the FordA dataset. In contrast we included both the SonyAIBORobotSurface1 and the SonyAIBORobotSurface2 dataset since there are very different.

\begin{table}[tbp]
\begin{tabular}{|l|c|c|c|c|}
\hline
\rowcolor[HTML]{9B9B9B} 
\textbf{Dataset}      & \textbf{Train Size} & \textbf{Test Size} & \textbf{\#Features} & \textbf{\#Classes} \\ \hline
Earthquakes           & 322                 & 139                & 512             & 2                       \\ \hline
FordA                 & 3601                & 1320               & 500             & 2                       \\ \hline
ItalyPowerDemand      & 67                  & 1029               &  24             & 2                       \\ \hline
Lightning2            & 60                  & 61                 & 637             & 2                       \\ \hline
MoteStrain            & 20                  & 1252               & 84              & 2                       \\ \hline
SonyAIBORobotSurface1 & 20                  & 601                & 70              & 2                       \\ \hline
SonyAIBORobotSurface2 & 27                  & 953                & 65              & 2                       \\ \hline
Wafer 				  & 1000                & 6164               & 152             & 2                       \\ \hline
\end{tabular}
\label{table:data_sets}
\caption{Table shows the selected datasets and their attributes.}
\end{table}
The datasets from the UEA \& UCR TSC Repository are typically already split into a training and a testing set and all the datasets are z-normalized. At first it seems cumbersome to use prefixed training and test sets, but this has the reason, that some of the datasets are designed so the train test split removes bias. Combining the different sets again for performing cross validation re-introduces this bias. Furthermore that authors state that "it is not feasible to cross validate everything" \cite{Bagnall2014}.
We also want to draw the readers attention to this very sound work that critically deals with cross-validation in time series analysis \cite{Bergmeir2018}.
\subsection{Algorithms for Instance Selection}
\label{subsec:algos_is}
In the course of the evaluation of our approach we have used the following algorithms for instance selection:
\begin{itemize}
	\item Distance2kNN: Distance to k-Nearest-Neighbors
	\item LDIS: Local density-based instance selection
	\item LKRR: Local kernel ridge regression for instance selection
\end{itemize}
\textbf{Distance2kNN} is based on the idea which is summarized in \cite{Chandola2009}, were various authors (Eskin et al. [2002], Angiulli and Pizzuti [2002] and Zhang and Wang [2006]) calculated an anomaly score for each instance based on the sum of its distances to their k nearest neighbors. Since this algorithm is very simple and quick to implement, it served as a feasibility study and baseline for our approach. Additionally it is a typical representative of a pure distance based outlier detection algorithm.\\
\textbf{LDIS} is a typical state of the art instance selection algorithm, which evaluates instances of each class separately, preserving only the densest instances in a certain neighborhood \cite{Carbonera2015}.\\ 
\textbf{LKRR} is the only relevant result of our state of the art analysis for the usage of outlier detection algorithms explicitly for instance selection and was already described in \ref{subsec:sota_od_for_is}.
\subsection{Feature Selectors}
\label{subsec:feature_selectors}
For the experiments, we generally want to point out that we are using feature selection only and do not change the representation (= feature construction) of the data which we evaluate during instance or feature selection.\\
Following \cite{ArauzoAzofra2011} we focused on feature selection methods which select the most relevant features as a a combination of an individual performance measure and a so called cutting criteria (for example, fixed number of features or ratio of features to select) to select the best number of features. All of the listed feature selectors are filter based feature selection methods. OneR feature selector is the only exception since it is using the OneR classifier to select the best subset of features.
We are using the following feature selectors, which are already provided by Weka (Source: \cite{ArauzoAzofra2011} and Weka Doc):
\begin{itemize}
	\item \textbf{GainRatio} \footnote{\url{http://weka.sourceforge.net/doc.stable-3-8/weka/attributeSelection/GainRatioAttributeEval.html}}
	\begin{itemize}
		\item Gain ratio is the ratio between information gain and the entropy of the feature
		\item GainR(Class, Attribute) = (H(Class) - H(Class $\vert$ Attribute)) / H(Attribute))
	\end{itemize}
	\item \textbf{InfoGain} \footnote{\url{http://weka.sourceforge.net/doc.stable-3-8/weka/attributeSelection/InfoGainAttributeEval.html}}
	\begin{itemize}
		\item Also known as \textbf{Mutual Information}, measures the worth of an feature with the corresponding class
		\item InfoGain(Class,Attribute) = H(Class) - H(Class $\vert$ Attribute)
	\end{itemize}
	\item \textbf{OneR} \footnote{\url{http://weka.sourceforge.net/doc.stable-3-8/weka/attributeSelection/OneRAttributeEval.html}}
	\begin{itemize}
		\item OneR is a single variable classifier used for feature selection
		\item Uses the minimum-error attribute for prediction a given class
	\end{itemize}
	\item \textbf{ReliefF} \footnote{\url{http://weka.sourceforge.net/doc.stable-3-8/weka/attributeSelection
	/ReliefFAttributeEval.html}}
	\begin{itemize}
		\item ReliefF evaluates individual features, but also takes into account the relation among features, by considering the value of a given attribute also by its nearest instance of the same and different class
	\end{itemize}
	\item \textbf{SymmetricalUncertainty} \footnote{\url{http://weka.sourceforge.net/doc.stable-3-8/weka/attributeSelection/SymmetricalUncertAttributeEval.html}}
	\begin{itemize}
		\item Evaluates the worth of an attribute by measuring the symmetrical uncertainty with respect to the class. 			\item SymmU(Class, Attribute) = 2 * (H(Class) - H(Class $\vert$ Attribute)) / H(Class) + H(Attribute).
	\end{itemize}  	
\end{itemize}
\subsection{Classifiers}
\label{subsec:classifiers}
Since our work is not about the development of a new TSC algorithm, we will follow the recommendation of \cite{Bagnall2017} for the usage of DTW1NN and Rotation Forest as base line algorithms. We already provided a brief description of these algorithms in subsection \ref{subsubsec:timeseries_classification}.
In the course of time series analysis and in particular time series classification, a suitable data representation plays an important role (compare sub-section \ref{subsec:time_series_analysis}), therefore with the DTW1NN we decided  for a classifier, which transforms given data representation by means of performing dynamic time warping.
\subsection{Parameters and Experiments}
\label{subsec:param_experiments}
For the instance selection we defined the following value ranges for the parameters and instances of components:
\begin{itemize}
	\item kNeigbours: 1, 5, 10
	\item odSelRatios: 0.0005, 0.05, 0.25, 0.5, 0.9
	\item odDescOrder: true, false
	\item nnSearch: LinearSearch
	\item DistanceFunctions: ChebyshevDistance \footnote{\url{http://weka.sourceforge.net/doc.stable-3-8/weka/core/ChebyshevDistance.html}}, EuclideanDistance\footnote{\url{http://weka.sourceforge.net/doc.stable-3-8/weka/core/EuclideanDistance.html}}, ManhattanDistance\footnote{\url{http://weka.sourceforge.net/doc.stable-3-8/weka/core/ManhattanDistance.html}}, MinkowskiDistance\footnote{\url{http://weka.sourceforge.net/doc.stable-3-8/weka/core/MinkowskiDistance.html}}
\end{itemize}
For the feature selection we defined the following value ranges for the parameters and instances of components:
\begin{itemize}
	\item attSelRatios: 0.002, 0.01, 0.1, 0.33, 0.66
	\item AttributeSelectors: GainRatio, InfoGain, OneR, Relief, SymmatricalUncert
\end{itemize}	
For both the odSelRatios and attSelRatios we followed the datasets with the largest number of instances and features.
The value 0.002 depends on the dataset with the largest number of features (Lightning2 with 637 features). Hereby we want to force that for each dataset at least for the attribute selection only one feature is selected. For the instance selection we wanted to have at least two instances as output of the instance selection.\\
Based on the different characteristics of our parameters and components, such as distance functions, there are 3,000 experiments per dataset and classifier. This number is the product of: \\
num kNeighbors * num odSelRatios * num odDescOrder * num Search * num distanceFunctions * num attSelRatios * and num AttributeSelectors\\
given by: 3 * 5 * 2 * 1 * 4 * 5 * 5 = 3,000.\\
These 3,000 experiments were performed for each combination of dataset, classifier and instance selection method. For clearer differentiation we will refer to these different combinations as \textbf{experimental suites}. For example all experiments on dataset Earthquakes with Rotation Forest combined with the LKRR instance selection method are one experimental suite.\\
Since we have eight different datasets, three different instance selection methods, and two classifiers we have 48 different experimental suites or 144,000 experiments in total.
\subsection{Performance Metrics}
\label{subsec:performance_metrics}
Within this section we briefly discuss the metrics we used to measure the performance of the classifiers.
The results on the UEA \& UCR TSC Repository are only available as \textbf{TP-rate} (or accuracy). Since we need to compare our results to the baseline of the UEA \& UCR TSC Repository, we also use the TP-rate as a metric to measure the performance of the classifiers.\\
The TP-rate computes the fraction of examples that are correctly classified. It is one of the most used evaluation metric for classification tasks. Nevertheless this metric has the fundamental problem that rare classes are neglected in contrast to normal classes. The usage of more comprehensive classification metrics, for example the F-Measure or the geometric mean was already suggested in 2004 by \cite{Weiss2004}.\\
Since we are aware of this problem, we listed for each classifier and each dataset the local baseline, the best feature selection result and the best instance selection result in Section \ref{sec:extended_performance_metrics} in the Appendix.
\subsection{Environment and Experiment Preparation}
\label{subsec:env_and_prep}
In order to inspect the detailed results or even re-run our experiments, we would like to refer to our public git-repository \footnote{\url{https://git.know-center.tugraz.at/summary/?r=~dcemernek/outfisltisl.git}}.\\
In order execute the experiments the requirements in Section \ref{subsec:install_and_execution} must be fulfilled. We also provided an overview of the basic components and their relations in Figure \ref{fig:experimental_approach}.\\
\myfig{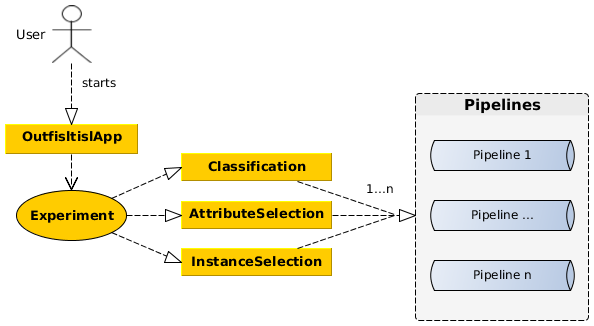}%% filename in figures folder
  {width=1.0\textwidth,height=1.0\textheight}%% maximum width/height, aspect ratio will be kept
  {Figures shows the dependencies and flow of a typical experiment.}%% caption
  {Experimental approach}%% optional (short) caption for table of figures
  {fig:experimental_approach}%% label
Our application has three different types of experiments, namely Classification, Attribute Selection and Instance Selection.\\
The classification experiments only perform pure classification, without any preprocessing of the data. The results of the classification experiments serve as local baseline.\\
Attribute selection experiments involve attribute selection steps before the classification step.\\
The instance selection experiments involve all steps, including the instance selection for the attribute selection followed by classification.
\section{Results}
\label{sec:results}
Within this section we present an overview of the results of our experiments. For a better overview we divided the results per classifier. For each classifier, there is a table that represents the TP-rate (rounded to three digits) for each experimental suite.\\
We also present the baselines for each classifier and dataset based on the UEA \& UCR TSC Repository in column "TSC" for method "NoAS" (= no attribute selection). Unfortunately, our local baselines for the Rotation Forest differ from the results of the UEA \& UCR Time Series Classification Repository, which is the reason we have also specified our local baseline in the column "LocBase" also with the method "NoAS". Furthermore we included the results for attribute selection and classification indicated by method "NoIS" (= no instance selection).\\ 
For better visibility we formatted the following items: best result per row and best method per dataset and classifier are formatted in bold.\\
We also provided bar charts for each dataset which can be found in the Section \ref{sec:result_plots} and an extended version of performance metrics including Precision, Recall and F-Measure in Section \ref{sec:extended_performance_metrics} in the Appendix.
For the results for each single experiment please refer to our git-repository \footnote{\url{https://git.know-center.tugraz.at/summary/?r=~dcemernek/outfisltisl.git}}.
\subsection{Results Rotation Forest}
\label{subsec:results_rotf}
Within this sub-section we provide the results table for the Rotation Forest classifier in Table  \ref{table:results_rotf}.

\begin{landscape}
\begin{longtable}{|l|l|c|c|c|c|c|c|c|}
\hline
\rowcolor[HTML]{9B9B9B} 
\textbf{Dataset} & \textbf{Method} & \textbf{TSC} & \textbf{LocBase} & \textbf{GainR} & \textbf{InfoGain} & \textbf{OneR} & \textbf{Relief} & \textbf{SymmU} \\ \hline
\endhead
Earthquakes & NoAS & 0.748 & 0.755 &  &  &  &  &  \\ \hline
Earthquakes & NoIS &  &  & 0.784 & 0.784 & 0.799 & 0.777 & \textbf{0.806} \\ \hline
Earthquakes & Dist2kNN &  &  & 0.799 & 0.791 & \textbf{0.806} & 0.791 & 0.791 \\ \hline
Earthquakes & \textbf{LDIS} &  &  & 0.791 & 0.799 & 0.799 & \textbf{0.813} & 0.806 \\ \hline
Earthquakes & \textbf{LKRR} &  &  & \textbf{0.813} & 0.799 & 0.806 & 0.799 & 0.806 \\ \hline
FordA & NoAS & 0.845 & 0.749 &  &  &  &  &  \\ \hline
FordA & NoIS &  &  & \textbf{0.76} & \textbf{0.76} & 0.744 & 0.736 & \textbf{0.76} \\ \hline
FordA & Dist2kNN &  &  & \textbf{0.786} & \textbf{0.786} & 0.769 & 0.76 & \textbf{0.786} \\ \hline
FordA & LDIS &  &  & 0.785 & 0.775 & 0.777 & 0.773 & \textbf{0.786} \\ \hline
FordA & \textbf{LKRR} &  &  & 0.778 & \textbf{0.789} & 0.775 & 0.762 & \textbf{0.789} \\ \hline
\begin{tabular}[c]{@{}l@{}}ItalyPower\\ Demand\end{tabular} & NoAS & 0.973 & 0.969 &  &  &  &  &  \\ \hline
\begin{tabular}[c]{@{}l@{}}ItalyPower\\ Demand\end{tabular} & NoIS &  &  & 0.971 & 0.973 & \textbf{0.974} & 0.97 & 0.969 \\ \hline
\begin{tabular}[c]{@{}l@{}}ItalyPower\\ Demand\end{tabular} & Dist2kNN &  &  & 0.973 & 0.973 & 0.972 & \textbf{0.975} & 0.973 \\ \hline
\begin{tabular}[c]{@{}l@{}}ItalyPower\\ Demand\end{tabular} & LDIS &  &  & \textbf{0.974} & 0.973 & \textbf{0.974} & \textbf{0.974} & \textbf{0.974} \\ \hline
\begin{tabular}[c]{@{}l@{}}ItalyPower\\ Demand\end{tabular} & \textbf{LKRR} &  &  & 0.975 & \textbf{0.976} & 0.974 & 0.975 & 0.974 \\ \hline
Lightning2 & NoAS & 0.689 & 0.754 &  &  &  &  &  \\ \hline
Lightning2 & NoIS &  &  & \textbf{0.787} & 0.77 & 0.77 & 0.754 & 0.754 \\ \hline
Lightning2 & \textbf{Dist2kNN} &  &  & 0.82 & 0.82 & 0.82 & \textbf{0.836} & 0.82 \\ \hline
Lightning2 & LDIS &  &  & 0.803 & \textbf{0.82} & \textbf{0.82} & \textbf{0.82} & \textbf{0.82} \\ \hline
Lightning2 & \textbf{LKRR} &  &  & 0.803 & \textbf{0.836} & 0.82 & 0.82 & \textbf{0.836} \\ \hline
MoteStrain & NoAS & 0.880 & 0.868 &  &  &  &  &  \\ \hline
MoteStrain & NoIS &  &  & 0.853 & \textbf{0.887} & 0.875 & 0.861 & 0.853 \\ \hline
MoteStrain & Dist2kNN &  &  & 0.884 & 0.885 & 0.894 & \textbf{0.895} & 0.883 \\ \hline
MoteStrain & LDIS &  &  & 0.883 & \textbf{0.9} & 0.894 & 0.895 & \textbf{0.9} \\ \hline
MoteStrain & \textbf{LKRR} &  &  & 0.902 & 0.897 & 0.902 & \textbf{0.906} & 0.897 \\ \hline
\begin{tabular}[c]{@{}l@{}}SonyAIBORobot\\ Surface1\end{tabular} & NoAS & 0.809 & 0.775 &  &  &  &  &  \\ \hline
\begin{tabular}[c]{@{}l@{}}SonyAIBORobot\\ Surface1\end{tabular} & NoIS &  &  & 0.779 & \textbf{0.799} & 0.667 & 0.759 & 0.779 \\ \hline
\begin{tabular}[c]{@{}l@{}}SonyAIBORobot\\ Surface1\end{tabular} & Dist2kNN &  &  & 0.864 & 0.864 & 0.864 & \textbf{0.88} & 0.864 \\ \hline
\begin{tabular}[c]{@{}l@{}}SonyAIBORobot\\ Surface1\end{tabular} & \textbf{LDIS} &  &  & 0.864 & \textbf{0.885} & 0.864 & 0.88 & \textbf{0.885} \\ \hline
\begin{tabular}[c]{@{}l@{}}SonyAIBORobot\\ Surface1\end{tabular} & LKRR &  &  & 0.867 & 0.864 & 0.864 & \textbf{0.88} & 0.864 \\ \hline
\begin{tabular}[c]{@{}l@{}}SonyAIBORobot\\ Surface2\end{tabular} & NoAS & 0.808 & 0.779 &  &  &  &  &  \\ \hline
\begin{tabular}[c]{@{}l@{}}SonyAIBORobot\\ Surface2\end{tabular} & NoIS &  &  & 0.811 & 0.805 & 0.807 & \textbf{0.837} & 0.835 \\ \hline
\begin{tabular}[c]{@{}l@{}}SonyAIBORobot\\ Surface2\end{tabular} & Dist2kNN &  &  & 0.839 & 0.835 & 0.837 & \textbf{0.854} & 0.837 \\ \hline
\begin{tabular}[c]{@{}l@{}}SonyAIBORobot\\ Surface2\end{tabular} & LDIS &  &  & \textbf{0.841} & \textbf{0.841} & 0.837 & 0.837 & \textbf{0.841} \\ \hline
\begin{tabular}[c]{@{}l@{}}SonyAIBORobot\\ Surface2\end{tabular} & \textbf{LKRR} &  &  & 0.845 & 0.844 & 0.837 & \textbf{0.856} & 0.844 \\ \hline
Wafer & NoAS & 0.994 & 0.992 &  &  &  &  &  \\ \hline
Wafer & NoIS &  &  & 0.996 & \textbf{0.997} & 0.996 & 0.996 & 0.996 \\ \hline
Wafer & Dist2kNN &  &  & 0.997 & \textbf{0.999} & \textbf{0.999} & 0.998 & 0.997 \\ \hline
Wafer & LDIS &  &  & 0.996 & 0.998 & 0.998 & \textbf{0.999} & \textbf{0.999} \\ \hline
Wafer & \textbf{LKRR} &  &  & 0.995 & 0.999 & \textbf{1} & 0.999 & 0.999 \\ \hline
\caption{Best TP-rates for all datasets, instance selection algorithms and attribute selectors for the Rotation Forest classifier.}
\end{longtable}
\label{table:results_rotf}
\end{landscape}

\myfig{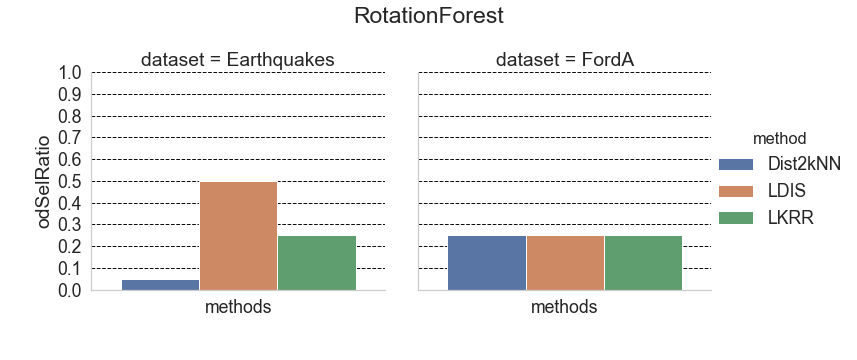}%% filename in figures folder
  {width=0.9\textwidth,height=1.0\textheight}%% maximum width/height, aspect ratio will be kept
  {OdSelRatio values for classifier Rotation Forest and corresponding datasets.}%% caption
  {Result plots odSelRatio - Rotation Forest-Datasets 1}%% optional (short) caption for table of figures
  {fig:result_plot_odSelRatio_rot_f_1}%% label
  
\myfig{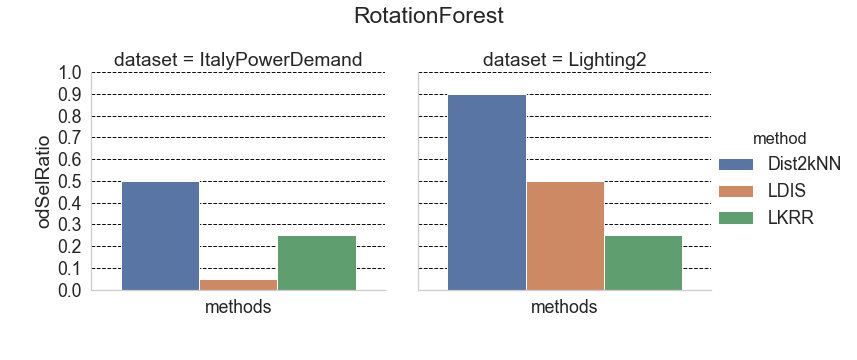}%% filename in figures folder
  {width=0.9\textwidth,height=1.0\textheight}%% maximum width/height, aspect ratio will be kept
  {OdSelRatio values for classifier Rotation Forest and corresponding datasets.}%% caption
  {Result plots odSelRatio - Rotation Forest-Datasets 2}%% optional (short) caption for table of figures
  {fig:result_plot_odSelRatio_rot_f_2}%% label

\myfig{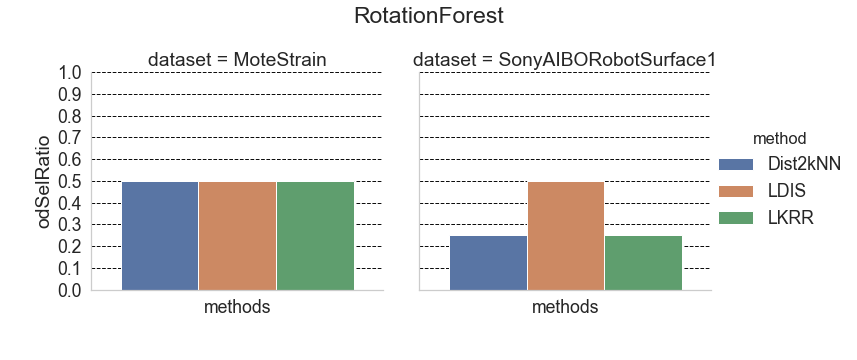}%% filename in figures folder
  {width=0.9\textwidth,height=1.0\textheight}%% maximum width/height, aspect ratio will be kept
  {OdSelRatio values for classifier Rotation Forest and corresponding datasets.}%% caption
  {Result plots odSelRatio - Rotation Forest-Datasets 3}%% optional (short) caption for table of figures
  {fig:result_plot_odSelRatio_rot_f_3}%% label
  
\myfig{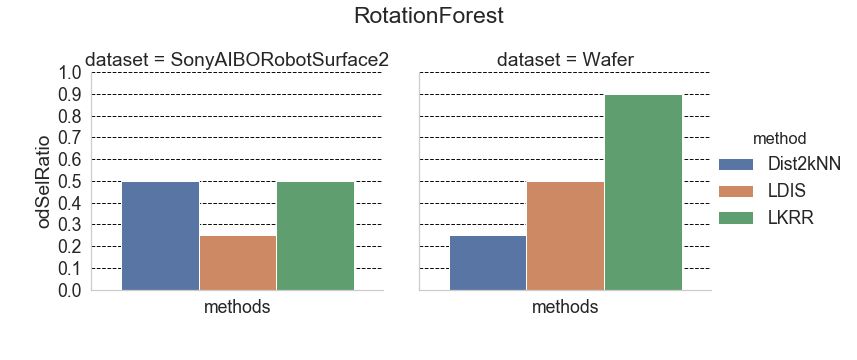}%% filename in figures folder
  {width=0.9\textwidth,height=1.0\textheight}%% maximum width/height, aspect ratio will be kept
  {OdSelRatio values for classifier Rotation Forest and corresponding datasets.}%% caption
  {Result plots odSelRatio - Rotation Forest-Datasets 4}%% optional (short) caption for table of figures
  {fig:result_plot_odSelRatio_rot_f_4}%% label

%we want to keep our figures within this sub-section, so use float barrier
\FloatBarrier
\subsection{Results DTW1NN}
\label{subsec:results_dtw}
Within this sub-section we provide the results table for the 1-Nearest Neighbor with dynamic time warping classifier in Table \ref{table:results_dtw}.

\begin{landscape}
\begin{longtable}{|l|l|c|c|c|c|c|c|c|}
\hline
\rowcolor[HTML]{9B9B9B} 
\textbf{Dataset} & \textbf{Method} & \textbf{TSC} & \textbf{LocBase} & \textbf{GainR} & \textbf{InfoGain} & \textbf{OneR} & \textbf{Relief} & \textbf{SymmU} \\ \hline
\endhead
Earthquakes & NoAS & 0.719 & 0.719 &  &  &  &  &  \\ \hline
Earthquakes & NoIS &  &  & 0.748 & \textbf{0.755} & 0.705 & \textbf{0.755} & \textbf{0.755} \\ \hline
Earthquakes & Dist2kNN &  &  & 0.755 & 0.755 & 0.763 & \textbf{0.77} & 0.748 \\ \hline
Earthquakes & LDIS &  &  & 0.748 & 0.77 & 0.777 & \textbf{0.791} & 0.755 \\ \hline
Earthquakes & \textbf{LKRR} &  &  & \textbf{0.799} & 0.777 & 0.755 & 0.77 & \textbf{0.799} \\ \hline
FordA & NoAS & 0.555 & 0.555 &  &  &  &  &  \\ \hline
FordA & NoIS &  &  & 0.613 & 0.613 & \textbf{0.626} & 0.62 & 0.613 \\ \hline
FordA & Dist2kNN &  &  & \textbf{0.646} & \textbf{0.646} & 0.644 & 0.637 & \textbf{0.646} \\ \hline
FordA & LDIS &  &  & 0.636 & 0.645 & 0.659 & \textbf{0.664} & 0.636 \\ \hline
FordA & \textbf{LKRR} &  &  & 0.646 & 0.646 & \textbf{0.655} & \textbf{0.655} & 0.646 \\ \hline
\begin{tabular}[c]{@{}l@{}}ItalyPower\\ Demand\end{tabular} & NoAS & 0.95 & 0.95 &  &  &  &  &  \\ \hline
\begin{tabular}[c]{@{}l@{}}ItalyPower\\ Demand\end{tabular} & NoIS &  &  & 0.973 & 0.972 & \textbf{0.976} & 0.968 & 0.973 \\ \hline
\begin{tabular}[c]{@{}l@{}}ItalyPower\\ Demand\end{tabular} & Dist2kNN &  &  & 0.974 & 0.973 & \textbf{0.976} & 0.973 & 0.972 \\ \hline
\begin{tabular}[c]{@{}l@{}}ItalyPower\\ Demand\end{tabular} & LDIS &  &  & 0.974 & \textbf{0.976} & 0.975 & 0.974 & 0.971 \\ \hline
\begin{tabular}[c]{@{}l@{}}ItalyPower\\ Demand\end{tabular} & \textbf{LKRR} &  &  & 0.976 & \textbf{0.977} & 0.976 & 0.975 & \textbf{0.977} \\ \hline
Lightning2 & NoAS & 0.869 & 0.869 &  &  &  &  &  \\ \hline
Lightning2 & NoIS &  &  & 0.803 & 0.738 & 0.803 & 0.77 & \textbf{0.82} \\ \hline
Lightning2 & Dist2kNN &  &  & 0.803 & \textbf{0.869} & 0.852 & 0.836 & \textbf{0.869} \\ \hline
Lightning2 & \textbf{LDIS} &  &  & 0.82 & 0.836 & 0.852 & \textbf{0.885} & 0.836 \\ \hline
Lightning2 & LKRR &  &  & 0.852 & \textbf{0.869} & 0.836 & 0.836 & 0.852 \\ \hline
MoteStrain & NoAS & 0.835 & 0.835 &  &  &  &  &  \\ \hline
MoteStrain & NoIS &  &  & 0.868 & 0.852 & \textbf{0.879} & 0.868 & 0.868 \\ \hline
MoteStrain & \textbf{Dist2kNN} &  &  & 0.853 & 0.862 & 0.875 & \textbf{0.89} & 0.862 \\ \hline
MoteStrain & \textbf{LDIS} &  &  & 0.863 & 0.881 & 0.883 & \textbf{0.89} & 0.881 \\ \hline
MoteStrain & \textbf{LKRR} &  &  & 0.883 & 0.872 & 0.886 & \textbf{0.89} & 0.883 \\ \hline
\begin{tabular}[c]{@{}l@{}}SonyAIBORobot\\ Surface1\end{tabular} & NoAS & 0.725 & 0.725 &  &  &  &  &  \\ \hline
\begin{tabular}[c]{@{}l@{}}SonyAIBORobot\\ Surface1\end{tabular} & NoIS &  &  & 0.772 & \textbf{0.799} & 0.759 & 0.76 & 0.772 \\ \hline
\begin{tabular}[c]{@{}l@{}}SonyAIBORobot\\ Surface1\end{tabular} & Dist2kNN &  &  & \textbf{0.854} & 0.845 & 0.845 & 0.837 & 0.845 \\ \hline
\begin{tabular}[c]{@{}l@{}}SonyAIBORobot\\ Surface1\end{tabular} & \textbf{LDIS} &  &  & \textbf{0.885} & 0.845 & 0.824 & 0.852 & 0.87 \\ \hline
\begin{tabular}[c]{@{}l@{}}SonyAIBORobot\\ Surface1\end{tabular} & LKRR &  &  & 0.854 & \textbf{0.87} & 0.845 & 0.865 & \textbf{0.87} \\ \hline
\begin{tabular}[c]{@{}l@{}}SonyAIBORobot\\ Surface2\end{tabular} & NoAS & 0.831 & 0.831 &  &  &  &  &  \\ \hline
\begin{tabular}[c]{@{}l@{}}SonyAIBORobot\\ Surface2\end{tabular} & NoIS &  &  & 0.816 & 0.813 & 0.834 & \textbf{0.85} & 0.836 \\ \hline
\begin{tabular}[c]{@{}l@{}}SonyAIBORobot\\ Surface2\end{tabular} & Dist2kNN &  &  & 0.853 & 0.852 & \textbf{0.866} & 0.841 & 0.853 \\ \hline
\begin{tabular}[c]{@{}l@{}}SonyAIBORobot\\ Surface2\end{tabular} & LDIS &  &  & 0.857 & 0.858 & \textbf{0.867} & \textbf{0.867} & 0.845 \\ \hline
\begin{tabular}[c]{@{}l@{}}SonyAIBORobot\\ Surface2\end{tabular} & \textbf{LKRR} &  &  & 0.858 & 0.857 & \textbf{0.886} & 0.858 & 0.858 \\ \hline
Wafer & NoAS & 0.98 & 0.98 &  &  &  &  &  \\ \hline
Wafer & NoIS &  &  & 0.982 & 0.982 & 0.98 & \textbf{0.991} & 0.979 \\ \hline
Wafer & Dist2kNN &  &  & 0.991 & \textbf{0.994} & 0.992 & \textbf{0.994} & 0.99 \\ \hline
Wafer & \textbf{LDIS} &  &  & \textbf{0.995} & 0.991 & 0.99 & 0.994 & 0.992 \\ \hline
Wafer & \textbf{LKRR} &  &  & 0.994 & \textbf{0.995} & 0.992 & \textbf{0.995} & 0.994 \\ \hline
\caption{Best TP-rates for all datasets, instance selection algorithms and attribute selectors for the DTW1NN classifier.}
\label{my-label}\\
\end{longtable}
\label{table:results_dtw}
\end{landscape}

\myfig{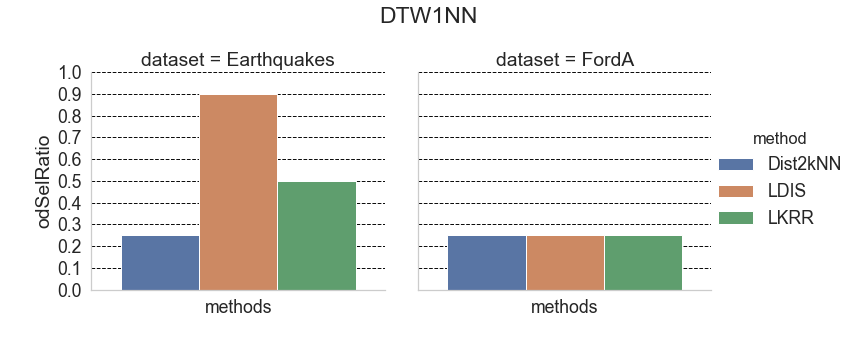}%% filename in figures folder
  {width=0.9\textwidth,height=1.0\textheight}%% maximum width/height, aspect ratio will be kept
  {OdSelRatio values for classifier DTW1NN and corresponding datasets.}%% caption
  {Result plots odSelRatio - DTW1NN-Datasets 1}%% optional (short) caption for table of figures
  {fig:result_plot_odSelRatio_dtw1nn_f_1}%% label
  
\myfig{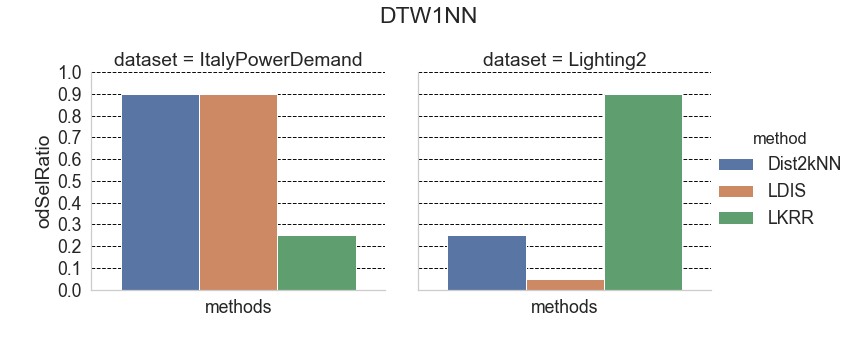}%% filename in figures folder
  {width=0.9\textwidth,height=1.0\textheight}%% maximum width/height, aspect ratio will be kept
  {OdSelRatio values for classifier DTW1NN and corresponding datasets.}%% caption
  {Result plots odSelRatio - DTW1NN-Datasets 2}%% optional (short) caption for table of figures
  {fig:result_plot_odSelRatio_dtw1nn_2}%% label

\myfig{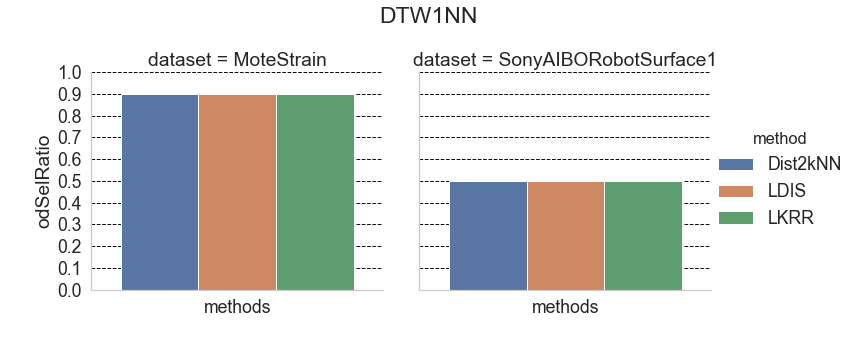}%% filename in figures folder
  {width=0.9\textwidth,height=1.0\textheight}%% maximum width/height, aspect ratio will be kept
  {OdSelRatio values for classifier DTW1NN and corresponding datasets.}%% caption
  {Result plots odSelRatio - DTW1NN-Datasets 3}%% optional (short) caption for table of figures
  {fig:result_plot_odSelRatio_dtw1nn_3}%% label
  
\myfig{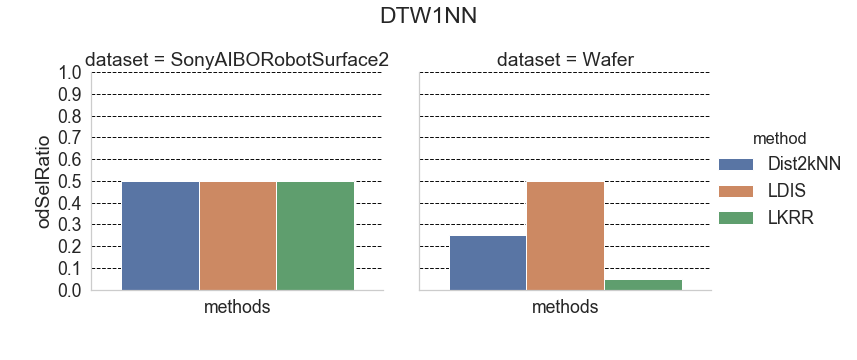}%% filename in figures folder
  {width=0.9\textwidth,height=1.0\textheight}%% maximum width/height, aspect ratio will be kept
  {OdSelRatio values for classifier DTW1NN and corresponding datasets.}%% caption
  {Result plots odSelRatio - DTW1NN-Datasets 4}%% optional (short) caption for table of figures
  {fig:result_plot_odSelRatio_dtw1nn_4}%% label

%we want to keep our figures within this sub-section, so use float barrier
\FloatBarrier
\subsection{Discussion of Results}
It is important that this work should not be a comparison of the performance of different feature selection algorithms, nor a comparison of different classification algorithms. The focus of our experiments lies entirely on the impact of instance selection for feature selection on the classification performance. To investigate this impact we have to answer our main research question, divided into the following two questions:
\begin{itemize}
	\item Are outlier detection algorithms applicable for filtering instances to influence feature selection algorithms?
	\item Do this adapted feature subsets have a positive impact on the performance of learning algorithms?
\end{itemize}
After analyzing the results of our experiments, we can answer both question with a clear "Yes". The results of our experiments suggest that for both classifiers the TP-rates, where we used some form of instance selection always increased, and therefore has a positive impact on the classification performance. Furthermore we found out that the used outlier detection algorithm (LKRR) is applicable for instance selection to positively influence the involved feature selection. We had no dataset where the the local baseline (classification only) was the best performing method, nor did we have a dataset where classification with attribute selection was better than the experiments involving instance selection.
For a deeper analysis, we also want to investigate the following questions:
\begin{itemize}
	\item What is the influence of the different datasets?
	\item What is the influence of different instance selection algorithms?
	\item What is the influence of the parameters \texttt{odSelRatios} and \texttt{odDescOrder}?
	\item What is the influence of parameters and components used for the nearest neighbor search: \texttt{kNeigbours}, \texttt{DistanceFunctions}?
\end{itemize}
The increase of performance of the classification strongly depends on the used datasets, components and parameters. We will inspect the specific questions in the following separate sections.
\subsubsection{Influence of different Datasets}
Concerning the performance of our approach given the different datasets, we found out that the results of our approach strongly depend on the used dataset. For datasets that already had an exceptional baseline result (for example ItalyPowerDemand and Wafer) the increase for the Rotation Forest classifier was only a matter of mills and only a matter of one or two percents for DTW1NN classifier. For these datasets we wanted to investigate whether the results might decrease, but this was not the case. Also the different classifiers had different increasing results given the different datasets. For example the DTW1NN classifier only had an increase of about two percent for the Lightning2 dataset, whereas the Rotation Forest classifier had an increase of eight percent. Interestingly both classifiers had the best performance for the SonyAIBORobotSurface1 dataset where DTW1NN increased for about 16 percent from the baselines and the Rotation Forest classifier increased ten percent. In general it seems that the lower the baseline performance of a dataset is, the higher is the increase of performance when performing instance selection for feature selection.
\subsubsection{Influence of Instance Selection Algorithms}
Referring to the different instance selection methods, we see that LKRR outlier detection algorithm had the largest increase in performance in thirteen out of sixteen experiment suites. We would like to draw the reader's attention to the fact that there are several equally performing methods for four experiments. Distance2KNN method only had the best performance for two out of sixteen experiment suites. Furthermore the instance selection method LDIS achieved the best performance for six out of sixteen experiment suites.
\subsubsection{Influence of Instance Selection and Nearest Neighbor Search Parameters}
\label{subsubsec:influcence_of_params}
This analysis is based on result parameter tables available in Section \ref{sec:result_parameter_tables} in the Appendix. For these tables we assembled for each experimental suite (recap: a combination of a dataset, a classifier and an instance selection method) the best result and all of the used parameters. In the case that there was more than a best result for a suite, all the best results were added to the table.\\
We first start with the parameter of the instance selection methods, namely odSelRatio and odDescOrder. Given the Distance2kNN method the \textbf{odSelRatio} for both the Rotation Forest and DTW1NN classifier seems to have a weak dependency on the distance function for the Wafer and the Earthquakes datasets. For the other datasets it seems that the odSelRatio parameter depends on the dataset (so it seems there is one parameter value specific for a given dataset, where the performance is the best). This is the same for the LDIS method for the DTW1NN classifier and all datasets. For the Rotation Forest classifier this is only different for the ItalyPowerDemand, Lightning2 and Wafer datasets, were there are variations in the odSelRatio among the best performing experiment suits. For the LKRR method each best solution strictly depends only on one specific value for the odSelRatio regardless of the classifier used.\\
For the odSelRatio it is also relevant which values this parameter has for the best performing experiments (we also provided plots in Section \ref{sec:results}).
For the Rotation Forest classifier only the datasets Wafer and Lighting2 needed 90 percent of the instances for the Distance2kNN and the LKRR respectively. The other results indicate that for the best results in each experimental suite only between 25 and 50 percent of the original training instances are required.\\
This situation is different for the DTW1NN classifier. It seems that this classifier needed at least 90 percent of instances for some instance selection methods to achieve the best results. Although some datasets only required a smaller percentage of instances, for example for the Lighting2 and Wafer datasets the LDIS and the LKRR method only needed 5 percent of instances, which is interesting because this is the opposite picture shown for the Rotation Forest classifier.\\
For the parameter \textbf{odDescOrder} concerning the DTW1NN classifier and the methods LDIS and LKRR it seems that the parameter is strictly depended on the dataset. For the Rotation Forest classifier and LKRR method this is not true for the ItalyPowerDemand dataset and for the LDIS method not true for the same dataset and additionally the Lightning and Wafer dataset. Concerning the Distance2kNN method and both classifiers we where not able to find a distinctive pattern.\\
Furthermore, we would like to investigate the influence of the kNeighbors parameter and the DistanceFunctions component.\\
Given the Distance2kNN method the \textbf{kNeighbors} seems to be irrelevant, since we have best results for all three possible values. With regard to the LDIS method, the parameter is only independent for the datasets ItalyPowerDemand, Lightning, and Wafer, but for both classifiers. For the other datasets, there is a clear dependency on the dataset, again for both classifiers. For the LKRR method we have a different analysis for both classifiers. For the DTW1NN classifier kNeighbors seems to depend on the dataset, with the only exception for the Wafer dataset. For the Rotation Forest we have the same dependency on the dataset except for the ItalyPowerDemand and SonyAIBORobotSurface1 dataset.\\
Finally we investigate the influence of the \textbf{distance functions} concerning the results of our experiments. Given the Distance2kNN method all distance functions seem to depend on the dataset. For the LDIS method and the DTW1NN classifier we have independence given the datasets ItalyPowerDemand, MoteStrain and SonyAIBORobotSurface2. The same is true for the Rotation Forest classifier except also the Wafer dataset. For the other dataset it seems that the distance functions are dependent on the datasets. Given the LKRR method all datasets are dependent on the distance function except the ItalyPowerDemand and Wafer dataset for the DTW1NN classifier. For the Rotation Forest classifier this is also true except for the datasets ItalyPowerDemand and SonyAIBORobotSurface1.\\
In summary, for very high performance records (ItalyPowerDemand, Wafer), the likelihood that one parameter depends on the dataset is smaller. For most other datasets, it seems that the parameters are almost always dependent on the dataset.          %% this is a suggestion: you have to create this file on demand
\chapter{Conclusions}
\label{cha:conclusions}
The aim of this work was to overcome the problem of imbalanced datasets, where normal instances are over-represented, while rare events are typically very limited. To tackle this problem we applied outlier detection techniques as instance selection for these rare events, in order to influence following feature selection methods. Furthermore we showed that this approach lead to positive increases of the performance of trailing classifiers. Although outlier detection and instance selection are very similar topics, there have been very few attempts to combine these two techniques until this work. Additionally we found no evidence that instance selection was used only for feature selection methods before. With the help of a dedicated experiment pipeline, we were able to evaluate a very large number of experiments with a great deal of different components and parameters. These experiments were performed on a variety of different datasets using state of the art time series classification algorithms. The evaluation of this large number of experiments showed that the used outlier detection algorithm LocalKernelRidgeRegression outperformed a comparable instance selection algorithm. To closely inspect our approach, or to execute our experiments, we have made all used datasets, detailed results, and our entire source code available in a public git repository.

\section{Outlook}
Our future research focuses on the following topics:
\begin{itemize}
	\item Especially for the DTW1NN classifier, the experiments for the larger datasets took a very long time ($>$ days). In this case, a parallelization of these experiments would be desirable (for example starting experimental suits for certain parameters like attributeSelRatio in parallel)
	\item Due to the very large number of experiments, the manual analysis of the results is very time-consuming. Due to the large number of different parameters, we would like to automate the effect of the different components and parameters (for example, by means of correlation analyzes).
	\item Due to the actual improvement of the feature selection and the associated classification, the next step would be the impact of our approach on other wrappers or embedded methods.
	\item Since instance selection is a data reduction technique, experiments using random sampling would be an interesting comparison to our approach.
	\item These aforementioned sampling methods are already used in so-called ensemble methods. It would be interesting to examine the meaningfulness of our approach in this area.
\end{itemize}         %% this is a suggestion: you have to create this file on demand

\appendix                       %% closes main document, appendix follows until end; only available in book-classes

\addpart*{Appendix}             %% adding Appendix to tableofcontents
\chapter{Appendix}
\label{cha:appendix}
\section{SOTA irrelevant results}
\label{sec:sota_irrelevant_results}
\begin{landscape}
\begin{longtable}{|l|l|l|l|l}
\cline{1-4}
\cellcolor[HTML]{9B9B9B}\textbf{Title}                                                                                                                & \cellcolor[HTML]{9B9B9B}\textbf{Year} & \cellcolor[HTML]{9B9B9B}\textbf{Autors}                                                                                                       & \cellcolor[HTML]{9B9B9B}\textbf{Comment}                                    &  \\ \cline{1-4}
\endhead
\begin{tabular}[c]{@{}l@{}}A methodology for training set \\ instance selection using \\ mutual information in \\ time series prediction\end{tabular} & 2014                                  & \begin{tabular}[c]{@{}l@{}}Miloš B.Stojanovića, \\ Miloš M.Božićb, \\ Milena M.Stanković, \\ Zoran P.Stajić\end{tabular}                      & Pure IS paper                                                               &  \\ \cline{1-4}
\begin{tabular}[c]{@{}l@{}}A survey on addressing high-class \\ imbalance in big data\end{tabular}                                                    & 2018                                  & \begin{tabular}[c]{@{}l@{}}Joffrey L. Leevy, \\ Taghi M. Khoshgoftaar, \\ Richard A. Bauder, \\ Naeem Seliya\end{tabular}                     & Wrong approach                                                              &  \\ \cline{1-4}
\begin{tabular}[c]{@{}l@{}}A survey on data preprocessing \\ for data stream mining: \\ Current status and future \\ directions\end{tabular}          & 2017                                  & \begin{tabular}[c]{@{}l@{}}Sergio Ramírez-Gallegoa, \\ Bartosz Krawczyk, \\ Salvador García, Michał Woźniak, \\ FranciscoHerrera\end{tabular} & Wrong approach                                                              &  \\ \cline{1-4}
\begin{tabular}[c]{@{}l@{}}Bagging of Instance Selection \\ Algorithms\end{tabular}                                                                   & 2014                                  & \begin{tabular}[c]{@{}l@{}}Marcin Blachnik, \\ Mirosław Kordos\end{tabular}                                                                   & Pure IS paper                                                               &  \\ \cline{1-4}
\begin{tabular}[c]{@{}l@{}}Data compression by volume \\ prototypes for streaming data\end{tabular}                                                   & 2010                                  & \begin{tabular}[c]{@{}l@{}}Kenji Tabata, Maiko Sato, \\ Mineichi Kudo\end{tabular}                                                            & Wrong approach                                                              &  \\ \cline{1-4}
\begin{tabular}[c]{@{}l@{}}Enhancing the Quality of \\ Noisy Training Data Using \\ a Genetic Algorithm and \\ Prototype Selection\end{tabular}       & 2008                                  & \begin{tabular}[c]{@{}l@{}}Boseon Byeon, Khaled Rasheed,\\ Prashant Doshi\end{tabular}                                                        & \begin{tabular}[c]{@{}l@{}}Wrong approach \\ (IS before OD)\end{tabular}    &  \\ \cline{1-4}
\begin{tabular}[c]{@{}l@{}}Ensemble Classifier for Mining \\ Data Streams\end{tabular}                                                                & 2014                                  & \begin{tabular}[c]{@{}l@{}}Ireneusz Czarnowski, \\ Piotr Jedrzejowicz\end{tabular}                                                            & Wrong approach                                                              &  \\ \cline{1-4}
\begin{tabular}[c]{@{}l@{}}Ensemble learning for data \\ stream analysis: A survey\end{tabular}                                                       & 2017                                  & \begin{tabular}[c]{@{}l@{}}Bartosz Krawczyk, \\ Leandro L.Minku, \\ João Gama, Jerzy Stefanowski, \\ Michał Woźniak\end{tabular}              & Wrong approach                                                              &  \\ \cline{1-4}
\begin{tabular}[c]{@{}l@{}}Genetic algorithms in feature \\ and instance selection\end{tabular}                                                       & 2013                                  & \begin{tabular}[c]{@{}l@{}}Chih-Fong Tsai,  William Eberle , \\ Chi-Yuan Chu\end{tabular}                                                     & Pure IS paper                                                               &  \\ \cline{1-4}
\begin{tabular}[c]{@{}l@{}}Instance selection based on \\ supervised clustering\end{tabular}                                                          & 2012                                  & \begin{tabular}[c]{@{}l@{}}Jun-Hai Zhai, Hong-Yu Xui, \\ Su-Fang Zhang, Na Li, Ta Li\end{tabular}                                             & Pure IS paper                                                               &  \\ \cline{1-4}
\begin{tabular}[c]{@{}l@{}}Instance Selection for Classifier \\ Performance Estimation in \\ Meta Learning\end{tabular}                               & 2017                                  & Marcin Blachnik                                                                                                                               & Wrong approach                                                              &  \\ \cline{1-4}
\begin{tabular}[c]{@{}l@{}}k-Nearest neighbors \\ optimization-based outlier \\ removal\end{tabular}                                                  & 2015                                  & \begin{tabular}[c]{@{}l@{}}Abraham Yosipof, \\ Hanoch Senderowitz\end{tabular}                                                                & Wrong approach                                                              &  \\ \cline{1-4}
\begin{tabular}[c]{@{}l@{}}Learning to detect representative \\ data for large scale \\ instance selection\end{tabular}                               & 2015                                  & \begin{tabular}[c]{@{}l@{}}Wei-Chao Lin, Chih-Fong Tsai, \\ Shih-Wen Ke, Chia-Wen Hung, \\ William Eberle\end{tabular}                        & Wrong approach                                                              &  \\ \cline{1-4}
\begin{tabular}[c]{@{}l@{}}Manifolds for training set \\ selection through outlier \\ detection\end{tabular}                                          & 2011                                  & Ahmed S. Tolba                                                                                                                                & Wrong approach                                                              &  \\ \cline{1-4}
\begin{tabular}[c]{@{}l@{}}Multiple-instance learning with \\ pairwise instance similarity\end{tabular}                                               & 2014                                  & \begin{tabular}[c]{@{}l@{}}Liming Yuan, Jiafeng Liu, \\ Xianglong Tang\end{tabular}                                                           & Wrong approach                                                              &  \\ \cline{1-4}
\begin{tabular}[c]{@{}l@{}}Mutual Information-Based Input \\ Selection for Electric Load \\ Time Series Forecasting\end{tabular}                      & 2013                                  & \begin{tabular}[c]{@{}l@{}}Miloš Božić, Miloš Stojanović, \\ Zoran Stajić, Nenad Floranović\end{tabular}                                      & Wrong approach                                                              &  \\ \cline{1-4}
\begin{tabular}[c]{@{}l@{}}New method for instance or \\ prototype selection using \\ mutual information in time series \\ prediction\end{tabular}    & 2010                                  & \begin{tabular}[c]{@{}l@{}}A.Guillen, L.J.Herrera, \\ G.Rubio, H.Pomares, \\ A.Lendasse, I.Rojas\end{tabular}                                 & Pure IS paper                                                               &  \\ \cline{1-4}
\begin{tabular}[c]{@{}l@{}}Noisy data elimination using \\ mutual k-nearest neighbor \\ for classification mining\end{tabular}                        & 2012                                  & Huawen Liu, Huawen Liu                                                                                                                        & Wrong approach                                                              &  \\ \cline{1-4}
\begin{tabular}[c]{@{}l@{}}Outlier detection for training-\\ based adaptive protocols\end{tabular}                                                    & 2013                                  & \begin{tabular}[c]{@{}l@{}}Hui Liu, Jialin He, \\ Dinesh Rajan, Joseph Camp\end{tabular}                                                      & \begin{tabular}[c]{@{}l@{}}Wrong field \\ (adaptive protocols)\end{tabular} &  \\ \cline{1-4}
\begin{tabular}[c]{@{}l@{}}Parallel MCNN (pMCNN) with \\ Application to Prototype Selection \\ on Large and Streaming Dat\end{tabular}                & 2017                                  & \begin{tabular}[c]{@{}l@{}}V. Susheela Devi, \\ Lakhpat Meena\end{tabular}                                                                    & Wrong approach                                                              &  \\ \cline{1-4}
\begin{tabular}[c]{@{}l@{}}Recent advances in scaling-down \\ sampling methods in \\ machine learning\end{tabular}                                    & 2017                                  & Amr ElRafey, Janusz Wojtusiak                                                                                                                 & Wrong approach                                                              &  \\ \cline{1-4}
\begin{tabular}[c]{@{}l@{}}Training set selection for the \\ prediction of essential genes\end{tabular}                                               & 2014                                  & \begin{tabular}[c]{@{}l@{}}Cheng J, Xu Z, Wu W, \\ Zhao L, Li X, Liu Y, Tao S\end{tabular}                                                    & Pure IS paper                                                               &  \\ \cline{1-4}
\caption{Results of the SOTA analysis of outlier detection for instance selection which turned out to be not relevant. }
\end{longtable}
\end{landscape}
\section{Extended performance metrics}
\label{sec:extended_performance_metrics}
\subsection{Extended performance metrics Rotation Forest}
\begin{landscape}
\begin{longtable}{|l|l|l|l|l|l|l|l|}
\hline
\rowcolor[HTML]{9B9B9B} 
\textbf{Dataset} & \textbf{ExpSuite} & \textbf{Method} & \textbf{TP\_RATE} & \textbf{FP\_RATE} & \textbf{Precision} & \textbf{Recall} & \textbf{F1} \\ \hline
\endhead
Earthquakes & LocBase & - & 0.755 & 0.613 & 0.718 & 0.755 & 0.711 \\ \hline
Earthquakes & AS-Only & SymmUncert & 0.806 & 0.520 & 0.802 & 0.806 & 0.773 \\ \hline
Earthquakes &  & LKRR & 0.813 & 0.518 & 0.819 & 0.813 & 0.779 \\ \hline
FordA & LocBase & - & 0.749 & 0.252 & 0.749 & 0.749 & 0.749 \\ \hline
FordA & AS-Only & \begin{tabular}[c]{@{}l@{}}InfoGain\\ SymmUncert\\ GainRatio\end{tabular} & 0.760 & 0.241 & 0.760 & 0.760 & 0.760 \\ \hline
FordA & With IS & LKRR & 0.789 & 0.211 & 0.789 & 0.789 & 0.789 \\ \hline
\begin{tabular}[c]{@{}l@{}}ItalyPower\\ Demand\end{tabular} & LocBase & - & 0.969 & 0.031 & 0.969 & 0.969 & 0.969 \\ \hline
\begin{tabular}[c]{@{}l@{}}ItalyPower\\ Demand\end{tabular} & AS-Only & OneR & 0.974 & 0.026 & 0.974 & 0.974 & 0.974 \\ \hline
\begin{tabular}[c]{@{}l@{}}ItalyPower\\ Demand\end{tabular} & With IS & LKRR & 0.976 & 0.024 & 0.976 & 0.976 & 0.976 \\ \hline
Lighting2 & LocBase & - & 0.754 & 0.247 & 0.755 & 0.754 & 0.754 \\ \hline
Lighting2 & AS-Only & GainRatio & 0.787 & 0.213 & 0.788 & 0.787 & 0.787 \\ \hline
Lighting2 & With IS & \begin{tabular}[c]{@{}l@{}}Dist2KNN\\ LKRR\end{tabular} & 0.836 & 0.166 & 0.836 & 0.836 & 0.836 \\ \hline
MoteStrain & LocBase & - & 0.868 & 0.135 & 0.868 & 0.868 & 0.868 \\ \hline
MoteStrain & AS-Only & InfoGain & 0.887 & 0.119 & 0.887 & 0.887 & 0.886 \\ \hline
MoteStrain & With IS & LKRR & 0.906 & 0.093 & 0.907 & 0.906 & 0.906 \\ \hline
\begin{tabular}[c]{@{}l@{}}SonyAIBORobot\\ Surface1\end{tabular} & LocBase & - & 0.775 & 0.172 & 0.846 & 0.775 & 0.772 \\ \hline
\begin{tabular}[c]{@{}l@{}}SonyAIBORobot\\ Surface1\end{tabular} & AS-Only & InfoGain & 0.799 & 0.153 & 0.859 & 0.799 & 0.797 \\ \hline
\begin{tabular}[c]{@{}l@{}}SonyAIBORobot\\ Surface1\end{tabular} & With IS & LDIS & 0.885 & 0.100 & 0.895 & 0.885 & 0.886 \\ \hline
\begin{tabular}[c]{@{}l@{}}SonyAIBORobot\\ Surface2\end{tabular} & LocBase & - & 0.779 & 0.227 & 0.785 & 0.779 & 0.780 \\ \hline
\begin{tabular}[c]{@{}l@{}}SonyAIBORobot\\ Surface2\end{tabular} & AS-Only & Relief & 0.837 & 0.216 & 0.839 & 0.837 & 0.834 \\ \hline
\begin{tabular}[c]{@{}l@{}}SonyAIBORobot\\ Surface2\end{tabular} & With IS & LKRR & 0.856 & 0.188 & 0.857 & 0.856 & 0.854 \\ \hline
Wafer & LocBase & - & 0.992 & 0.047 & 0.992 & 0.992 & 0.992 \\ \hline
Wafer & AS-Only & InfoGain & 0.997 & 0.015 & 0.997 & 0.997 & 0.997 \\ \hline
Wafer & With IS & LKRR & 1.000 & 0.004 & 1.000 & 1.000 & 1.000 \\ \hline
\caption{Extended version of performance metrics for Rotation Forest classifier. Shown are only the local baseline, best attribute selector and best instance selection for each dataset.}
\label{table:extended_results_rotf}\\
\end{longtable}
\end{landscape}
\subsection{Extended performance metrics DTW1NN}
\begin{landscape}
\begin{longtable}{|l|l|l|c|c|c|c|c|}
\hline
\rowcolor[HTML]{9B9B9B} 
\textbf{Dataset} & \textbf{ExpSuite} & \textbf{Method} & \textbf{TP\_RATE} & \textbf{FP\_RATE} & \textbf{Precision} & \textbf{Recall} & \textbf{F1} \\ \hline
\endhead
Earthquakes & LocBase & - & 0.719 & 0.644 & 0.666 & 0.719 & 0.679 \\ \hline
Earthquakes & AS-Only & SymmUncert & 0.755 & 0.518 & 0.731 & 0.755 & 0.737 \\ \hline
Earthquakes & With IS & LKRR & 0.799 & 0.485 & 0.783 & 0.799 & 0.776 \\ \hline
FordA & LocBase & - & 0.555 & 0.449 & 0.554 & 0.555 & 0.553 \\ \hline
FordA & AS-Only & OneR & 0.626 & 0.381 & 0.628 & 0.626 & 0.622 \\ \hline
FordA & With IS & LKRR & 0.655 & 0.349 & 0.655 & 0.655 & 0.654 \\ \hline
\begin{tabular}[c]{@{}l@{}}ItalyPower\\ Demand\end{tabular} & LocBase & - & 0.950 & 0.049 & 0.951 & 0.950 & 0.950 \\ \hline
\begin{tabular}[c]{@{}l@{}}ItalyPower\\ Demand\end{tabular} & AS-Only & OneR & 0.976 & 0.024 & 0.976 & 0.976 & 0.976 \\ \hline
\begin{tabular}[c]{@{}l@{}}ItalyPower\\ Demand\end{tabular} & With IS & LKRR & 0.977 & 0.023 & 0.977 & 0.977 & 0.977 \\ \hline
Lighting2 & LocBase & - & 0.869 & 0.149 & 0.882 & 0.869 & 0.866 \\ \hline
Lighting2 & AS-Only & SymmUncert & 0.820 & 0.185 & 0.820 & 0.820 & 0.819 \\ \hline
Lighting2 & With IS & LDIS & 0.885 & 0.119 & 0.885 & 0.885 & 0.885 \\ \hline
MoteStrain & LocBase & - & 0.835 & 0.167 & 0.835 & 0.835 & 0.835 \\ \hline
MoteStrain & AS-Only & OneR & 0.879 & 0.124 & 0.879 & 0.879 & 0.879 \\ \hline
MoteStrain & With IS & \begin{tabular}[c]{@{}l@{}}Dist2KNN\\ LDIS\\ LKRR\end{tabular} & 0.890 & 0.114 & 0.890 & 0.890 & 0.890 \\ \hline
\begin{tabular}[c]{@{}l@{}}SonyAIBORobot\\ Surface1\end{tabular} & LocBase & - & 0.725 & 0.207 & 0.830 & 0.725 & 0.716 \\ \hline
\begin{tabular}[c]{@{}l@{}}SonyAIBORobot\\ Surface1\end{tabular} & AS-Only & InfoGain & 0.799 & 0.151 & 0.863 & 0.799 & 0.796 \\ \hline
\begin{tabular}[c]{@{}l@{}}SonyAIBORobot\\ Surface1\end{tabular} & With IS & LDIS & 0.885 & 0.090 & 0.904 & 0.885 & 0.886 \\ \hline
\begin{tabular}[c]{@{}l@{}}SonyAIBORobot\\ Surface1\end{tabular} & LocBase & - & 0.831 & 0.210 & 0.830 & 0.831 & 0.829 \\ \hline
\begin{tabular}[c]{@{}l@{}}SonyAIBORobot\\ Surface2\end{tabular} & AS-Only & Relief & 0.850 & 0.176 & 0.849 & 0.850 & 0.849 \\ \hline
\begin{tabular}[c]{@{}l@{}}SonyAIBORobot\\ Surface2\end{tabular} & With IS & LKRR & 0.886 & 0.142 & 0.885 & 0.886 & 0.885 \\ \hline
Wafer & LocBase & - & 0.980 & 0.145 & 0.980 & 0.980 & 0.979 \\ \hline
Wafer & AS-Only & Relief & 0.991 & 0.039 & 0.991 & 0.991 & 0.991 \\ \hline
Wafer & With IS & \begin{tabular}[c]{@{}l@{}}LDIS\\ LKRR\end{tabular} & 0.995 & 0.036 & 0.995 & 0.995 & 0.995 \\ \hline
\caption{Extended version of performance metrics for DTW1NN classifier. Shown are only the local baseline, best attribute selector and best instance selection for each dataset.}
\label{table:extended_results_dtw}\\
\end{longtable}
\end{landscape}
\section{Result plots}
\label{sec:result_plots}
\subsection{Result plots Rotation Forest}
\label{subsec:result_plots_rotf}
\myfig{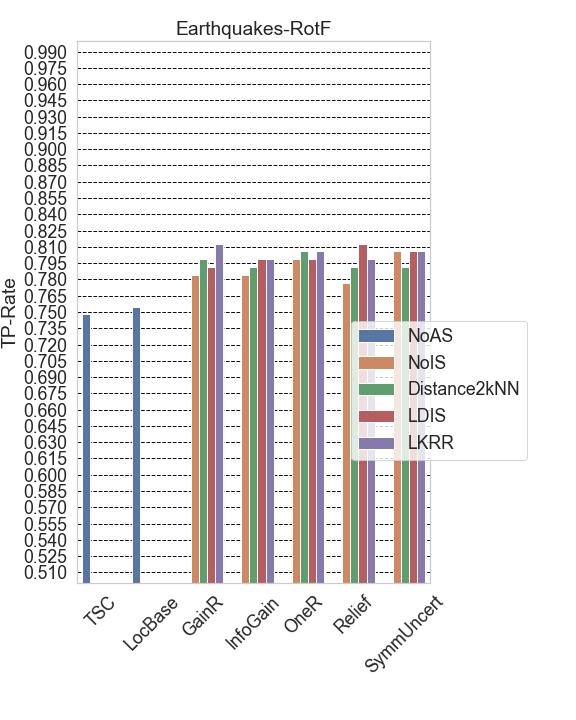}%% filename in figures folder
  {width=0.95\textwidth,height=1.0\textheight}%% maximum width/height, aspect ratio will be kept
  {Results for dataset Earthquakes for all methods (including baseline from TSC and local baseline) and attribute selectors for classifier Rotation Forest.}%% caption
  {Result plot Rotation Forest for dataset Earthquakes}%% optional (short) caption for table of figures
  {fig:result_rotf_earthquakes}%% label
  
\myfig{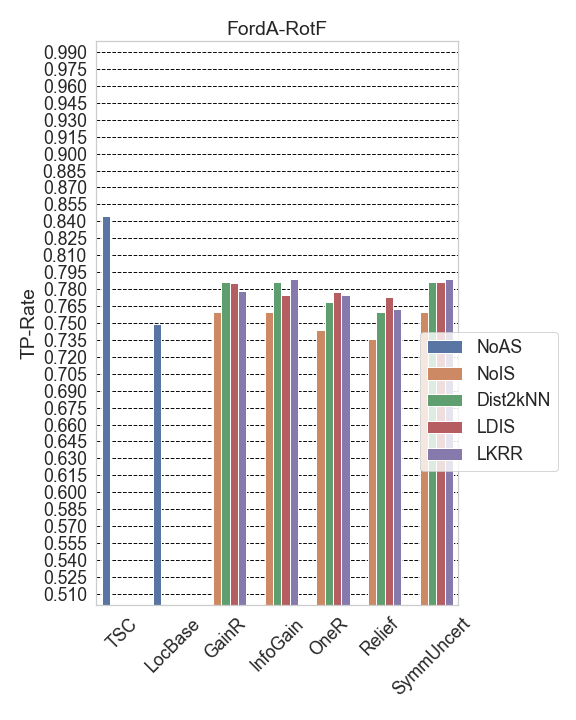}%% filename in figures folder
  {width=0.95\textwidth,height=1.0\textheight}%% maximum width/height, aspect ratio will be kept
  {Results for dataset FordA for all methods (including baseline from TSC and local baseline) and attribute selectors for classifier Rotation Forest.}%% caption
  {Result plot Rotation Forest for dataset FordA}%% optional (short) caption for table of figures
  {fig:result_rotf_fordA}%% label
    
\myfig{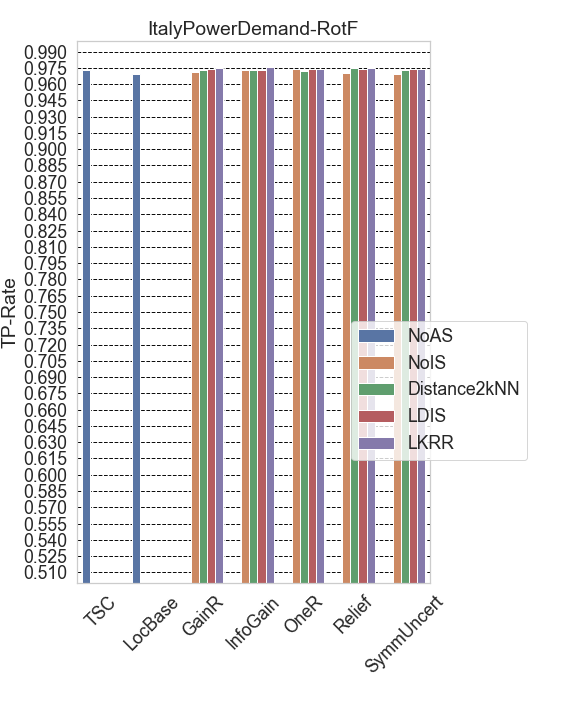}%% filename in figures folder
  {width=0.95\textwidth,height=1.0\textheight}%% maximum width/height, aspect ratio will be kept
  {Results for dataset ItalyPowerDemand for all methods (including baseline from TSC and local baseline) and attribute selectors for classifier Rotation Forest.}%% caption
  {Result plot Rotation Forest for dataset ItalyPowerDemand}%% optional (short) caption for table of figures
  {fig:result_rotf_italy_power_demand}%% label
  
\myfig{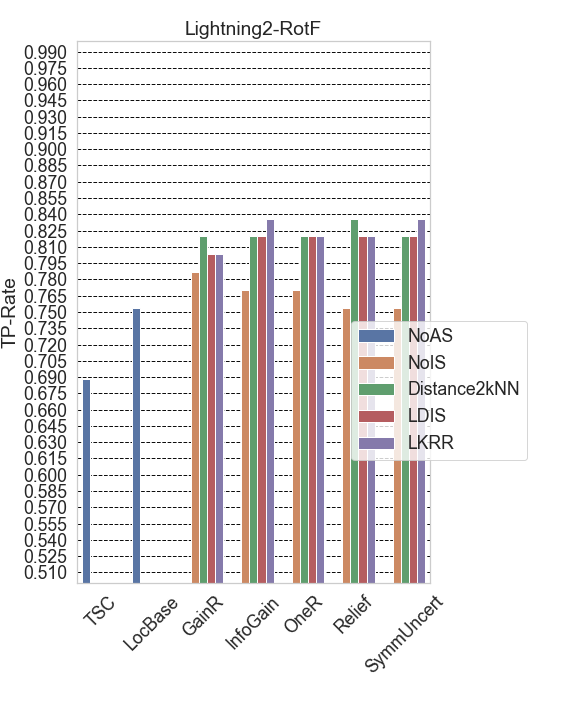}%% filename in figures folder
  {width=0.95\textwidth,height=1.0\textheight}%% maximum width/height, aspect ratio will be kept
  {Results for dataset Lightning2 for all methods (including baseline from TSC and local baseline) and attribute selectors for classifier Rotation Forest.}%% caption
  {Result plot Rotation Forest for dataset Lightning2}%% optional (short) caption for table of figures
  {fig:result_rotf_lightning2}%% label
  
\myfig{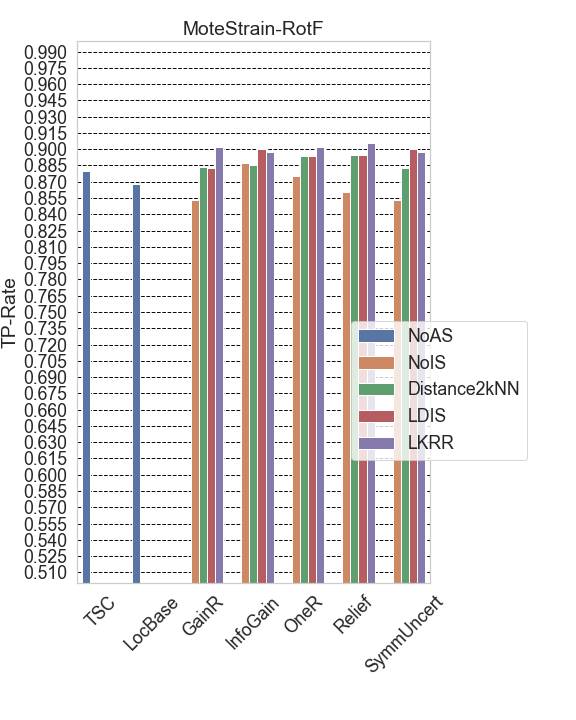}%% filename in figures folder
  {width=0.95\textwidth,height=1.0\textheight}%% maximum width/height, aspect ratio will be kept
  {Results for dataset MoteStrain for all methods (including baseline from TSC and local baseline) and attribute selectors for classifier Rotation Forest.}%% caption
  {Result plot Rotation Forest for dataset MoteStrain}%% optional (short) caption for table of figures
  {fig:result_rotf_motestrain}%% label
  
\myfig{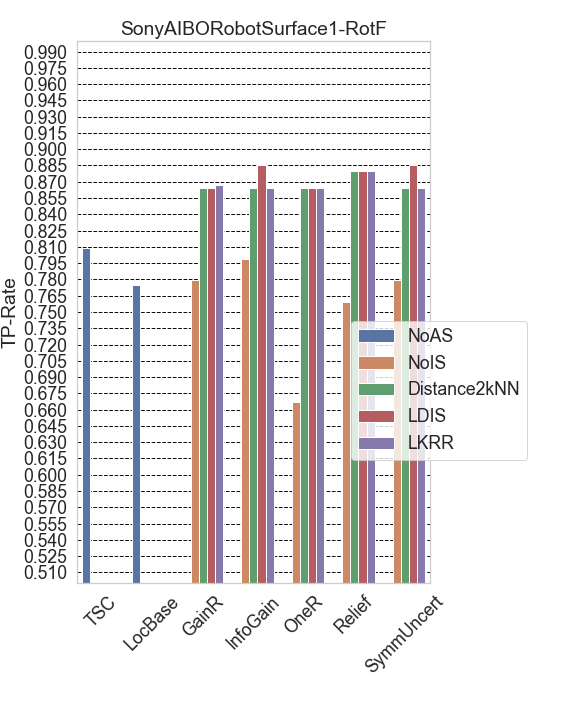}%% filename in figures folder
  {width=0.95\textwidth,height=1.0\textheight}%% maximum width/height, aspect ratio will be kept
  {Results for dataset SonyAIBORobotSurface1 for all methods (including baseline from TSC and local baseline) and attribute selectors for classifier Rotation Forest.}%% caption
  {Result plot Rotation Forest for dataset SonyAIBORobotSurface1}%% optional (short) caption for table of figures
  {fig:result_rotf_sony1}%% label
  
\myfig{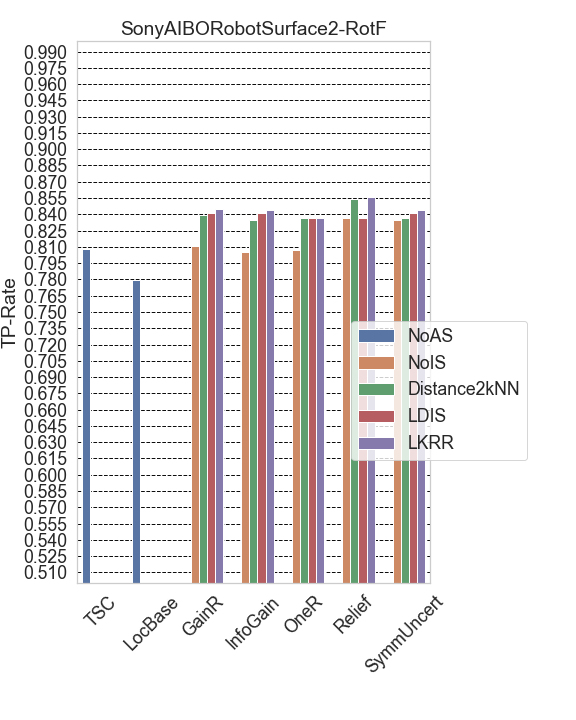}%% filename in figures folder
  {width=0.95\textwidth,height=1.0\textheight}%% maximum width/height, aspect ratio will be kept
  {Results for dataset SonyAIBORobotSurface2 for all methods (including baseline from TSC and local baseline) and attribute selectors for classifier Rotation Forest.}%% caption
  {Result plot Rotation Forest for dataset SonyAIBORobotSurface2}%% optional (short) caption for table of figures
  {fig:result_rotf_sony2}%% label
  
\myfig{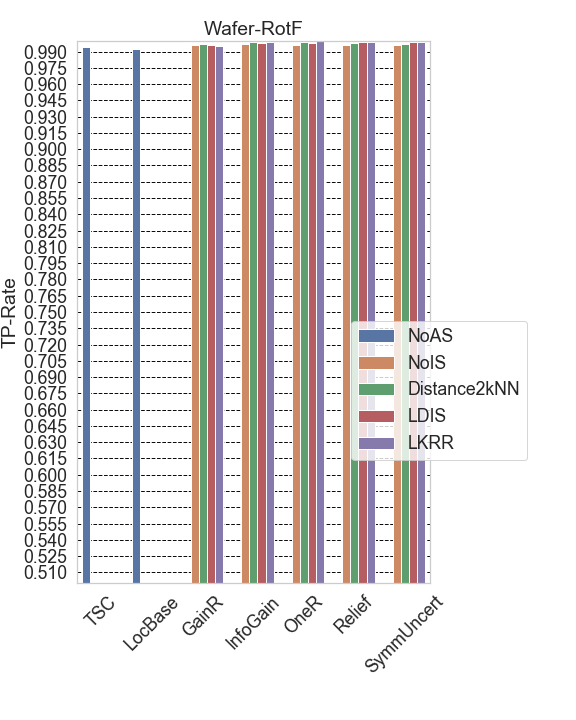}%% filename in figures folder
  {width=0.95\textwidth,height=1.0\textheight}%% maximum width/height, aspect ratio will be kept
  {Results for dataset Wafer for all methods (including baseline from TSC and local baseline) and attribute selectors for classifier Rotation Forest.}%% caption
  {Result plot Rotation Forest for dataset Wafer}%% optional (short) caption for table of figures
  {fig:result_rotf_wafer}%% label
  
%we want to keep our figures within this sub-section, so use float barrier
\FloatBarrier
\subsection{Results plots DTW1NN}
\label{subsec:result_plots_dtw}
\myfig{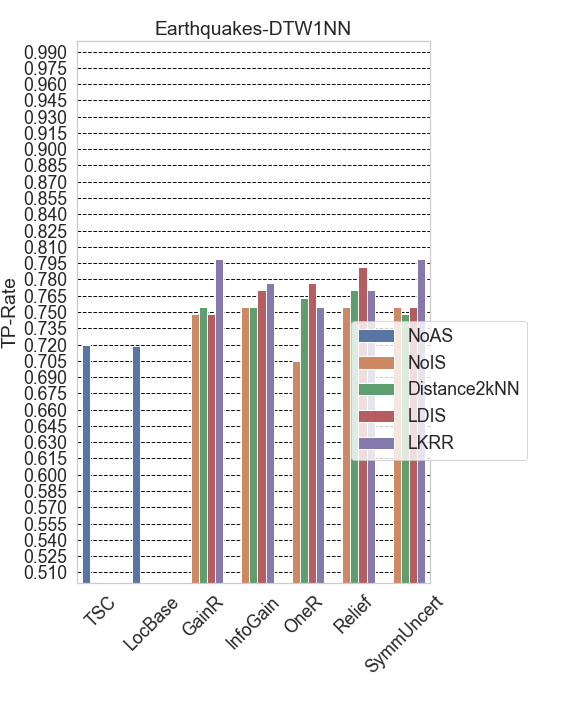}%% filename in figures folder
  {width=0.95\textwidth,height=1.0\textheight}%% maximum width/height, aspect ratio will be kept
  {Results for dataset Earthquakes for all methods (including baseline from TSC and local baseline) and attribute selectors for classifier DTW1NN.}%% caption
  {Result plot DTW1NN for dataset Earthquakes}%% optional (short) caption for table of figures
  {fig:result_dtw_earthquakes}%% label

\myfig{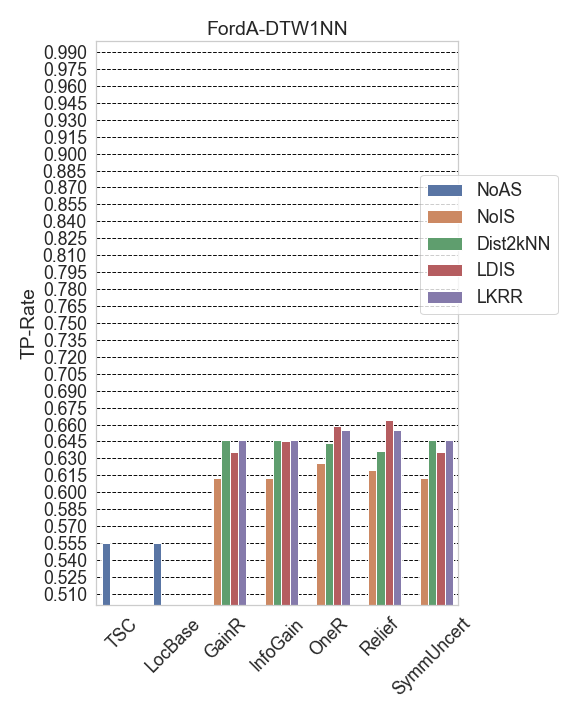}%% filename in figures folder
  {width=0.95\textwidth,height=1.0\textheight}%% maximum width/height, aspect ratio will be kept
  {Results for dataset FordA for all methods (including baseline from TSC and local baseline) and attribute selectors for classifier DTW1NN.}%% caption
  {Result plot DTW1NN for dataset FordA}%% optional (short) caption for table of figures
  {fig:result_dtw_forda}%% label
    
 \myfig{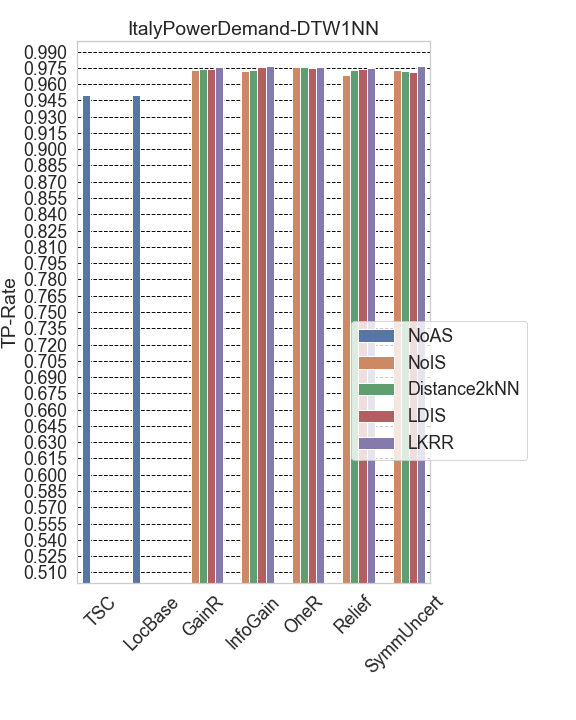}%% filename in figures folder
  {width=0.95\textwidth,height=1.0\textheight}%% maximum width/height, aspect ratio will be kept
  {Results for dataset ItalyPowerDemand for all methods (including baseline from TSC and local baseline) and attribute selectors for classifier DTW1NN.}%% caption
  {Result plot DTW1NN for dataset ItalyPowerDemand}%% optional (short) caption for table of figures
  {fig:result_dtw_italy_power_demand}%% label

\myfig{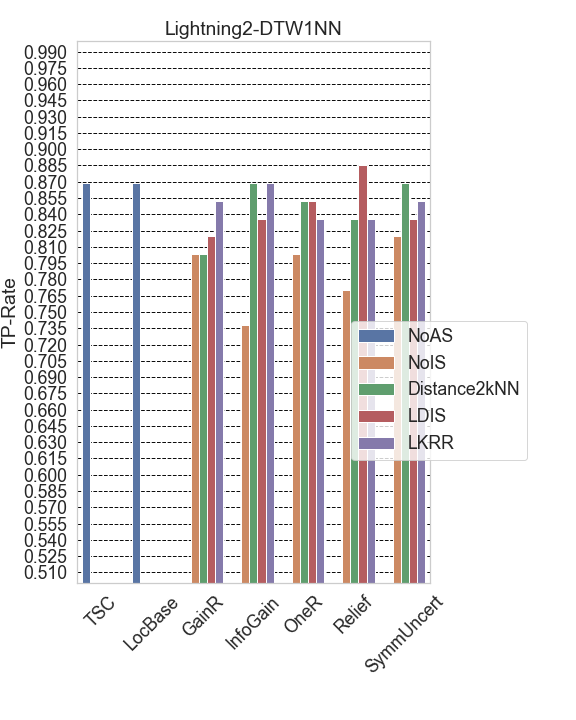}%% filename in figures folder
  {width=0.95\textwidth,height=1.0\textheight}%% maximum width/height, aspect ratio will be kept
  {Results for dataset Lightning2 for all methods (including baseline from TSC and local baseline) and attribute selectors for classifier DTW1NN.}%% caption
  {Result plot DTW1NN for dataset Lightning2}%% optional (short) caption for table of figures
  {fig:result_dtw_lightning2}%% label
  
\myfig{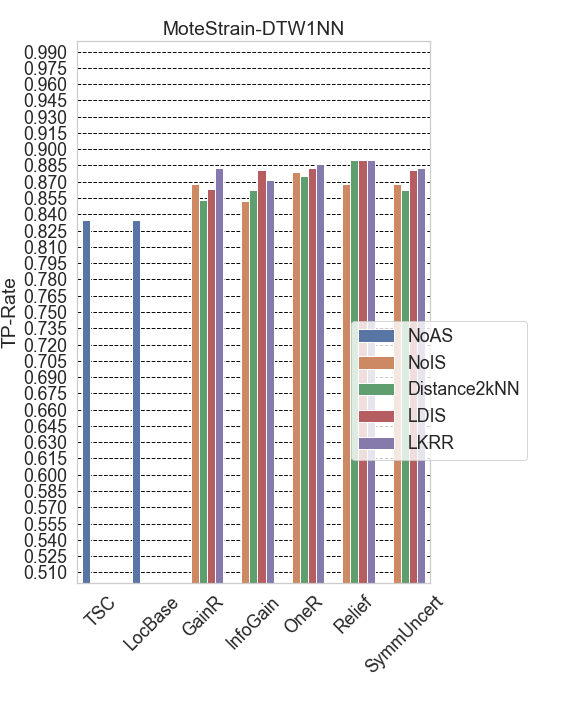}%% filename in figures folder
  {width=0.95\textwidth,height=1.0\textheight}%% maximum width/height, aspect ratio will be kept
  {Results for dataset MoteStrain for all methods (including baseline from TSC and local baseline) and attribute selectors for classifier DTW1NN.}%% caption
  {Result plot DTW1NN for dataset MoteStrain}%% optional (short) caption for table of figures
  {fig:result_dtw_motestrain}%% label
  
\myfig{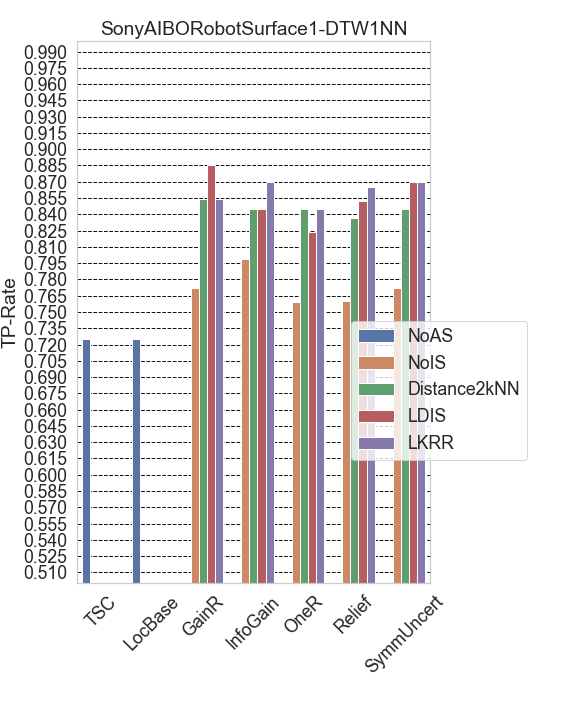}%% filename in figures folder
  {width=0.95\textwidth,height=1.0\textheight}%% maximum width/height, aspect ratio will be kept
  {Results for dataset SonyAIBORobotSurface1 for all methods (including baseline from TSC and local baseline) and attribute selectors for classifier DTW1NN.}%% caption
  {Result plot DTW1NN for dataset SonyAIBORobotSurface1}%% optional (short) caption for table of figures
  {fig:result_dtw_sony1}%% label

\myfig{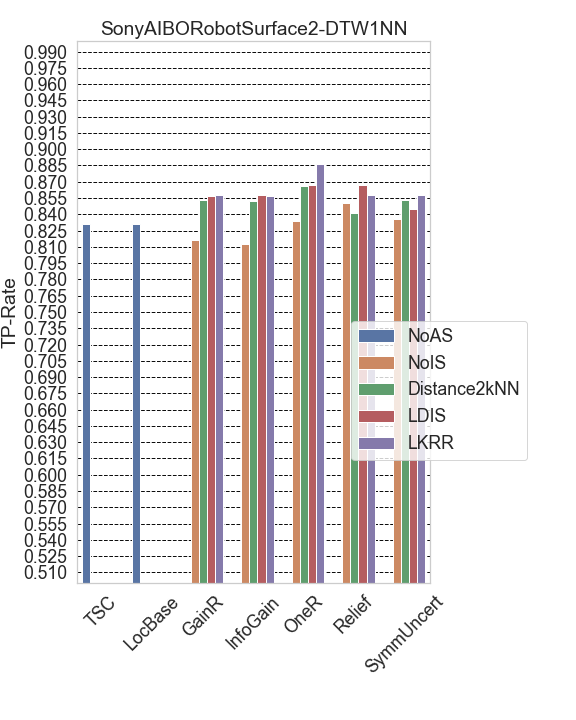}%% filename in figures folder
  {width=0.95\textwidth,height=1.0\textheight}%% maximum width/height, aspect ratio will be kept
  {Results for dataset SonyAIBORobotSurface2 for all methods (including baseline from TSC and local baseline) and attribute selectors for classifier DTW1NN.}%% caption
  {Result plot DTW1NN for dataset SonyAIBORobotSurface2}%% optional (short) caption for table of figures
  {fig:result_dtw_sony2}%% label
  
\myfig{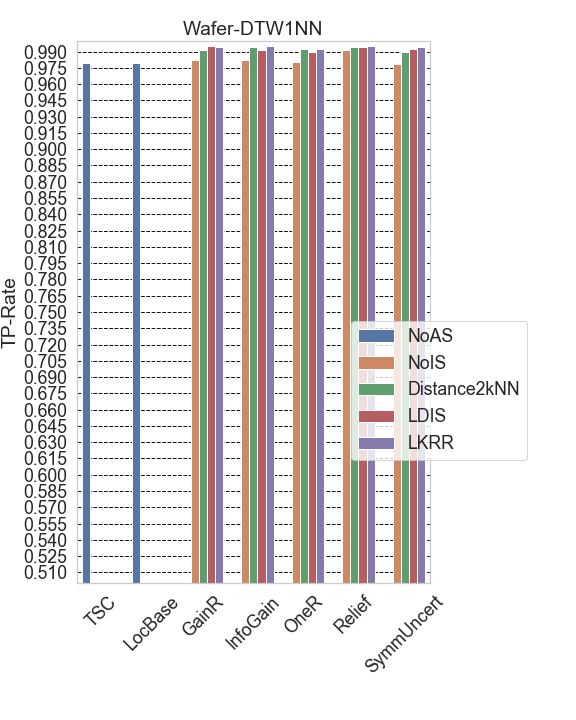}%% filename in figures folder
  {width=0.95\textwidth,height=1.0\textheight}%% maximum width/height, aspect ratio will be kept
  {Results for dataset Wafer for all methods (including baseline from TSC and local baseline) and attribute selectors for classifier DTW1NN.}%% caption
  {Result plot DTW1NN for dataset Wafer}%% optional (short) caption for table of figures
  {fig:result_dtw_wafer}%% label

%we want to keep our figures within this sub-section, so use float barrier
\FloatBarrier
\section{Result parameter tables}
\label{sec:result_parameter_tables}
\subsection{Result parameter tables Rotation Forest}
\begin{landscape}
\begin{longtable}{|l|l|l|l|l|l|}
\hline
\rowcolor[HTML]{9B9B9B} 
\textbf{dataset} & \textbf{kNeighbors} & \textbf{odSelRatio} & \textbf{odDistFunc} & \textbf{odDescOrder} & \textbf{TP\_RATE} \\ \hline
\endhead
Earthquakes & 1 & 0.25 & Euclidean & false & 0.806 \\ \hline
Earthquakes & 1 & 0.25 & Minkowski & false & 0.806 \\ \hline
Earthquakes & 5 & 0.25 & Euclidean & false & 0.806 \\ \hline
Earthquakes & 5 & 0.25 & Minkowski & false & 0.806 \\ \hline
Earthquakes & 10 & 0.25 & Euclidean & false & 0.806 \\ \hline
Earthquakes & 10 & 0.25 & Minkowski & false & 0.806 \\ \hline
Earthquakes & 1 & 0.05 & Manhattan & true & 0.806 \\ \hline
Earthquakes & 5 & 0.05 & Manhattan & true & 0.806 \\ \hline
Earthquakes & 10 & 0.05 & Manhattan & true & 0.806 \\ \hline
FordA & 1 & 0.25 & Euclidean & true & 0.786 \\ \hline
FordA & 1 & 0.25 & Minkowski & true & 0.786 \\ \hline
FordA & 5 & 0.25 & Euclidean & true & 0.786 \\ \hline
FordA & 5 & 0.25 & Minkowski & true & 0.786 \\ \hline
FordA & 10 & 0.25 & Euclidean & true & 0.786 \\ \hline
FordA & 10 & 0.25 & Minkowski & true & 0.786 \\ \hline
\begin{tabular}[c]{@{}l@{}}ItalyPower\\ Demand\end{tabular} & 1 & 0.5 & Manhattan & false & 0.975 \\ \hline
\begin{tabular}[c]{@{}l@{}}ItalyPower\\ Demand\end{tabular} & 5 & 0.5 & Manhattan & false & 0.975 \\ \hline
\begin{tabular}[c]{@{}l@{}}ItalyPower\\ Demand\end{tabular} & 10 & 0.5 & Manhattan & false & 0.975 \\ \hline
\begin{tabular}[c]{@{}l@{}}ItalyPower\\ Demand\end{tabular} & 1 & 0.9 & Manhattan & true & 0.975 \\ \hline
\begin{tabular}[c]{@{}l@{}}ItalyPower\\ Demand\end{tabular} & 5 & 0.9 & Manhattan & true & 0.975 \\ \hline
\begin{tabular}[c]{@{}l@{}}ItalyPower\\ Demand\end{tabular} & 10 & 0.9 & Manhattan & true & 0.975 \\ \hline
Lighting2 & 1 & 0.9 & Euclidean & true & 0.836 \\ \hline
Lighting2 & 1 & 0.9 & Minkowski & true & 0.836 \\ \hline
Lighting2 & 5 & 0.9 & Euclidean & true & 0.836 \\ \hline
Lighting2 & 5 & 0.9 & Minkowski & true & 0.836 \\ \hline
Lighting2 & 10 & 0.9 & Euclidean & true & 0.836 \\ \hline
Lighting2 & 10 & 0.9 & Minkowski & true & 0.836 \\ \hline
MoteStrain & 1 & 0.5 & Chebyshev & false & 0.895 \\ \hline
MoteStrain & 5 & 0.5 & Chebyshev & false & 0.895 \\ \hline
MoteStrain & 10 & 0.5 & Chebyshev & false & 0.895 \\ \hline
MoteStrain & 1 & 0.5 & Chebyshev & false & 0.894 \\ \hline
MoteStrain & 5 & 0.5 & Chebyshev & false & 0.894 \\ \hline
MoteStrain & 10 & 0.5 & Chebyshev & false & 0.894 \\ \hline
\begin{tabular}[c]{@{}l@{}}SonyAIBORobot\\ Surface1\end{tabular} & 1 & 0.25 & Chebyshev & false & 0.88 \\ \hline
\begin{tabular}[c]{@{}l@{}}SonyAIBORobot\\ Surface1\end{tabular} & 5 & 0.25 & Chebyshev & false & 0.88 \\ \hline
\begin{tabular}[c]{@{}l@{}}SonyAIBORobot\\ Surface1\end{tabular} & 10 & 0.25 & Chebyshev & false & 0.88 \\ \hline
\begin{tabular}[c]{@{}l@{}}SonyAIBORobot\\ Surface2\end{tabular} & 1 & 0.5 & Manhattan & false & 0.854 \\ \hline
\begin{tabular}[c]{@{}l@{}}SonyAIBORobot\\ Surface2\end{tabular} & 5 & 0.5 & Manhattan & false & 0.854 \\ \hline
\begin{tabular}[c]{@{}l@{}}SonyAIBORobot\\ Surface2\end{tabular} & 10 & 0.5 & Manhattan & false & 0.854 \\ \hline
Wafer & 1 & 0.25 & Euclidean & false & 0.999 \\ \hline
Wafer & 1 & 0.25 & Minkowski & false & 0.999 \\ \hline
Wafer & 5 & 0.25 & Euclidean & false & 0.999 \\ \hline
Wafer & 5 & 0.25 & Minkowski & false & 0.999 \\ \hline
Wafer & 10 & 0.25 & Euclidean & false & 0.999 \\ \hline
Wafer & 10 & 0.25 & Minkowski & false & 0.999 \\ \hline
Wafer & 1 & 0.5 & Manhattan & true & 0.999 \\ \hline
Wafer & 5 & 0.5 & Manhattan & true & 0.999 \\ \hline
Wafer & 10 & 0.5 & Manhattan & true & 0.999 \\ \hline
\caption{Result parameters for classifier RotationForest and instance selection method Distance2KNN}
\label{tab:result_parameter_tables_rotF_1}\\
\end{longtable}
\end{landscape}

\begin{landscape}
\begin{longtable}{|l|l|l|l|l|l|}
\hline
\rowcolor[HTML]{9B9B9B} 
\textbf{dataset} & \textbf{kNeighbors} & \textbf{odSelRatio} & \textbf{odDistFunc} & \textbf{odDescOrder} & \textbf{TP\_RATE} \\ \hline
\endhead
Earthquakes & 10 & 0.5 & Chebyshev & true & 0.799 \\ \hline
FordA & 10 & 0.25 & Chebyshev & true & 0.786 \\ \hline
ItalyPower & 5 & 0.05 & Manhattan & false & 0.974 \\ \hline
Demand & 5 & 0.05 & Manhattan & false & 0.974 \\ \hline
\begin{tabular}[c]{@{}l@{}}ItalyPowerDemand\end{tabular} & 10 & 0.05 & Euclidean & false & 0.974 \\ \hline
\begin{tabular}[c]{@{}l@{}}ItalyPowerDemand\end{tabular} & 10 & 0.05 & Manhattan & false & 0.974 \\ \hline
\begin{tabular}[c]{@{}l@{}}ItalyPowerDemand\end{tabular} & 10 & 0.05 & Minkowski & false & 0.974 \\ \hline
\begin{tabular}[c]{@{}l@{}}ItalyPowerDemand\end{tabular} & 1 & 0.5 & Manhattan & false & 0.974 \\ \hline
\begin{tabular}[c]{@{}l@{}}ItalyPowerDemand\end{tabular} & 1 & 0.9 & Chebyshev & false & 0.974 \\ \hline
\begin{tabular}[c]{@{}l@{}}ItalyPowerDemand\end{tabular} & 5 & 0.5 & Chebyshev & false & 0.974 \\ \hline
\begin{tabular}[c]{@{}l@{}}ItalyPowerDemand\end{tabular} & 5 & 0.9 & Chebyshev & true & 0.974 \\ \hline
\begin{tabular}[c]{@{}l@{}}ItalyPowerDemand\end{tabular} & 10 & 0.9 & Manhattan & false & 0.974 \\ \hline
Lighting2 & 5 & 0.5 & Manhattan & false & 0.82 \\ \hline
Lighting2 & 5 & 0.9 & Manhattan & false & 0.82 \\ \hline
Lighting2 & 10 & 0.9 & Manhattan & true & 0.82 \\ \hline
MoteStrain & 5 & 0.5 & Euclidean & true & 0.9 \\ \hline
MoteStrain & 5 & 0.5 & Euclidean & true & 0.9 \\ \hline
MoteStrain & 5 & 0.5 & Minkowski & true & 0.9 \\ \hline
MoteStrain & 5 & 0.5 & Minkowski & true & 0.9 \\ \hline
\begin{tabular}[c]{@{}l@{}}SonyAIBORobotSurface1\end{tabular} & 10 & 0.5 & Manhattan & false & 0.885 \\ \hline
\begin{tabular}[c]{@{}l@{}}SonyAIBORobotSurface1\end{tabular} & 10 & 0.5 & Manhattan & false & 0.885 \\ \hline
\begin{tabular}[c]{@{}l@{}}SonyAIBORobotSurface2\end{tabular} & 1 & 0.25 & Euclidean & false & 0.841 \\ \hline
\begin{tabular}[c]{@{}l@{}}SonyAIBORobotSurface2\end{tabular} & 1 & 0.25 & Manhattan & false & 0.841 \\ \hline
\begin{tabular}[c]{@{}l@{}}SonyAIBORobotSurface2\end{tabular} & 1 & 0.25 & Manhattan & false & 0.841 \\ \hline
\begin{tabular}[c]{@{}l@{}}SonyAIBORobotSurface2\end{tabular} & 1 & 0.25 & Minkowski & false & 0.841 \\ \hline
Wafer & 1 & 0.5 & Chebyshev & true & 0.999 \\ \hline
Wafer & 1 & 0.5 & Euclidean & false & 0.999 \\ \hline
Wafer & 1 & 0.5 & Minkowski & false & 0.999 \\ \hline
Wafer & 5 & 0.5 & Chebyshev & true & 0.999 \\ \hline
Wafer & 10 & 0.5 & Chebyshev & true & 0.999 \\ \hline
Wafer & 10 & 0.9 & Euclidean & false & 0.999 \\ \hline
Wafer & 10 & 0.9 & Minkowski & false & 0.999 \\ \hline
Wafer & 10 & 0.5 & Manhattan & true & 0.999 \\ \hline
\caption{Result parameters for classifier RotationForest and instance selection method LDIS}
\label{tab:result_parameter_tables_rotF_2}\\
\end{longtable}
\end{landscape}

\begin{landscape}
\begin{longtable}{|l|l|l|l|l|l|}
\hline
\rowcolor[HTML]{9B9B9B} 
\textbf{dataset} & \textbf{kNeighbors} & \textbf{odSelRatio} & \textbf{odDistFunc} & \textbf{odDescOrder} & \textbf{TP\_RATE} \\ \hline
\endhead
Earthquakes & 5 & 0.25 & Manhattan & false & 0.813 \\ \hline
FordA & 1 & 0.25 & Manhattan & false & 0.789 \\ \hline
ItalyPowerDemand & 5 & 0.25 & Manhattan & true & 0.976 \\ \hline
ItalyPowerDemand & 1 & 0.5 & Euclidean & false & 0.976 \\ \hline
ItalyPowerDemand & 1 & 0.5 & Minkowski & false & 0.976 \\ \hline
Lighting2 & 1 & 0.25 & Euclidean & false & 0.836 \\ \hline
Lighting2 & 1 & 0.25 & Euclidean & false & 0.836 \\ \hline
Lighting2 & 1 & 0.25 & Minkowski & false & 0.836 \\ \hline
Lighting2 & 1 & 0.25 & Minkowski & false & 0.836 \\ \hline
MoteStrain & 5 & 0.5 & Manhattan & false & 0.906 \\ \hline
SonyAIBORobotSurface1 & 1 & 0.25 & Chebyshev & false & 0.88 \\ \hline
SonyAIBORobotSurface1 & 5 & 0.25 & Chebyshev & false & 0.88 \\ \hline
SonyAIBORobotSurface1 & 10 & 0.25 & Chebyshev & false & 0.88 \\ \hline
SonyAIBORobotSurface2 & 10 & 0.5 & Euclidean & false & 0.856 \\ \hline
SonyAIBORobotSurface2 & 10 & 0.5 & Manhattan & false & 0.856 \\ \hline
SonyAIBORobotSurface2 & 10 & 0.5 & Minkowski & false & 0.856 \\ \hline
Wafer & 5 & 0.9 & Manhattan & false & 1 \\ \hline
\caption{Result parameters for classifier RotationForest and instance selection method LKRR}
\label{tab:result_parameter_tables_rotF_3}\\
\end{longtable}
\end{landscape}
\subsection{Result parameter tables DTW1NN}
\begin{landscape}
\begin{longtable}{|l|l|l|l|l|l|}
\hline
\rowcolor[HTML]{9B9B9B} 
\textbf{dataset} & \textbf{kNeighbors} & \textbf{odSelRatio} & \textbf{odDistFunc} & \textbf{odDescOrder} & \textbf{TP\_RATE} \\ \hline
\endhead
Earthquakes & 1 & 0.25 & Euclidean & true & 0.755 \\ \hline
Earthquakes & 1 & 0.25 & Minkowski & true & 0.755 \\ \hline
Earthquakes & 5 & 0.25 & Euclidean & true & 0.755 \\ \hline
Earthquakes & 5 & 0.25 & Minkowski & true & 0.755 \\ \hline
Earthquakes & 10 & 0.25 & Euclidean & true & 0.755 \\ \hline
Earthquakes & 10 & 0.25 & Minkowski & true & 0.755 \\ \hline
FordA & 5 & 0.25 & Minkowski & false & 0.646 \\ \hline
FordA & 5 & 0.25 & Minkowski & false & 0.646 \\ \hline
FordA & 5 & 0.25 & Minkowski & false & 0.646 \\ \hline
FordA & 1 & 0.25 & Euclidean & false & 0.646 \\ \hline
FordA & 1 & 0.25 & Euclidean & false & 0.646 \\ \hline
FordA & 1 & 0.25 & Euclidean & false & 0.646 \\ \hline
FordA & 10 & 0.25 & Euclidean & false & 0.646 \\ \hline
FordA & 10 & 0.25 & Euclidean & false & 0.646 \\ \hline
FordA & 10 & 0.25 & Euclidean & false & 0.646 \\ \hline
FordA & 1 & 0.25 & Minkowski & false & 0.646 \\ \hline
FordA & 1 & 0.25 & Minkowski & false & 0.646 \\ \hline
FordA & 1 & 0.25 & Minkowski & false & 0.646 \\ \hline
FordA & 5 & 0.25 & Euclidean & false & 0.646 \\ \hline
FordA & 5 & 0.25 & Euclidean & false & 0.646 \\ \hline
FordA & 5 & 0.25 & Euclidean & false & 0.646 \\ \hline
FordA & 10 & 0.25 & Minkowski & false & 0.646 \\ \hline
FordA & 10 & 0.25 & Minkowski & false & 0.646 \\ \hline
FordA & 10 & 0.25 & Minkowski & false & 0.646 \\ \hline
\begin{tabular}[c]{@{}l@{}}ItalyPowerDemand\end{tabular} & 1 & 0.9 & Euclidean & false & 0.976 \\ \hline
\begin{tabular}[c]{@{}l@{}}ItalyPowerDemand\end{tabular} & 1 & 0.9 & Manhattan & false & 0.976 \\ \hline
\begin{tabular}[c]{@{}l@{}}ItalyPowerDemand\end{tabular} & 1 & 0.9 & Minkowski & false & 0.976 \\ \hline
\begin{tabular}[c]{@{}l@{}}ItalyPowerDemand\end{tabular} & 5 & 0.9 & Euclidean & false & 0.976 \\ \hline
\begin{tabular}[c]{@{}l@{}}ItalyPowerDemand\end{tabular} & 5 & 0.9 & Manhattan & false & 0.976 \\ \hline
\begin{tabular}[c]{@{}l@{}}ItalyPowerDemand\end{tabular} & 5 & 0.9 & Minkowski & false & 0.976 \\ \hline
\begin{tabular}[c]{@{}l@{}}ItalyPowerDemand\end{tabular} & 10 & 0.9 & Euclidean & false & 0.976 \\ \hline
\begin{tabular}[c]{@{}l@{}}ItalyPowerDemand\end{tabular} & 10 & 0.9 & Manhattan & false & 0.976 \\ \hline
\begin{tabular}[c]{@{}l@{}}ItalyPowerDemand\end{tabular} & 10 & 0.9 & Minkowski & false & 0.976 \\ \hline
Lighting2 & 1 & 0.25 & Euclidean & true & 0.869 \\ \hline
Lighting2 & 1 & 0.25 & Euclidean & true & 0.869 \\ \hline
Lighting2 & 1 & 0.25 & Minkowski & true & 0.869 \\ \hline
Lighting2 & 1 & 0.25 & Minkowski & true & 0.869 \\ \hline
Lighting2 & 5 & 0.25 & Euclidean & true & 0.869 \\ \hline
Lighting2 & 5 & 0.25 & Euclidean & true & 0.869 \\ \hline
Lighting2 & 5 & 0.25 & Minkowski & true & 0.869 \\ \hline
Lighting2 & 5 & 0.25 & Minkowski & true & 0.869 \\ \hline
Lighting2 & 10 & 0.25 & Euclidean & true & 0.869 \\ \hline
Lighting2 & 10 & 0.25 & Euclidean & true & 0.869 \\ \hline
Lighting2 & 10 & 0.25 & Minkowski & true & 0.869 \\ \hline
Lighting2 & 10 & 0.25 & Minkowski & true & 0.869 \\ \hline
MoteStrain & 1 & 0.9 & Euclidean & false & 0.89 \\ \hline
MoteStrain & 1 & 0.9 & Manhattan & false & 0.89 \\ \hline
MoteStrain & 1 & 0.9 & Minkowski & false & 0.89 \\ \hline
MoteStrain & 5 & 0.9 & Euclidean & false & 0.89 \\ \hline
MoteStrain & 5 & 0.9 & Manhattan & false & 0.89 \\ \hline
MoteStrain & 5 & 0.9 & Minkowski & false & 0.89 \\ \hline
MoteStrain & 10 & 0.9 & Euclidean & false & 0.89 \\ \hline
MoteStrain & 10 & 0.9 & Manhattan & false & 0.89 \\ \hline
MoteStrain & 10 & 0.9 & Minkowski & false & 0.89 \\ \hline
\begin{tabular}[c]{@{}l@{}}SonyAIBORobotSurface1\end{tabular} & 1 & 0.5 & Euclidean & true & 0.854 \\ \hline
\begin{tabular}[c]{@{}l@{}}SonyAIBORobotSurface1\end{tabular} & 1 & 0.5 & Minkowski & true & 0.854 \\ \hline
\begin{tabular}[c]{@{}l@{}}SonyAIBORobotSurface1\end{tabular} & 5 & 0.5 & Euclidean & true & 0.854 \\ \hline
\begin{tabular}[c]{@{}l@{}}SonyAIBORobotSurface1\end{tabular} & 5 & 0.5 & Minkowski & true & 0.854 \\ \hline
\begin{tabular}[c]{@{}l@{}}SonyAIBORobotSurface1\end{tabular} & 10 & 0.5 & Euclidean & true & 0.854 \\ \hline
\begin{tabular}[c]{@{}l@{}}SonyAIBORobotSurface1\end{tabular} & 10 & 0.5 & Minkowski & true & 0.854 \\ \hline
\begin{tabular}[c]{@{}l@{}}SonyAIBORobotSurface2\end{tabular} & 1 & 0.5 & Euclidean & true & 0.866 \\ \hline
\begin{tabular}[c]{@{}l@{}}SonyAIBORobotSurface2\end{tabular} & 1 & 0.5 & Manhattan & true & 0.866 \\ \hline
\begin{tabular}[c]{@{}l@{}}SonyAIBORobotSurface2\end{tabular} & 1 & 0.5 & Minkowski & true & 0.866 \\ \hline
\begin{tabular}[c]{@{}l@{}}SonyAIBORobotSurface2\end{tabular} & 5 & 0.5 & Euclidean & true & 0.866 \\ \hline
\begin{tabular}[c]{@{}l@{}}SonyAIBORobotSurface2\end{tabular} & 5 & 0.5 & Manhattan & true & 0.866 \\ \hline
\begin{tabular}[c]{@{}l@{}}SonyAIBORobotSurface2\end{tabular} & 5 & 0.5 & Minkowski & true & 0.866 \\ \hline
\begin{tabular}[c]{@{}l@{}}SonyAIBORobotSurface2\end{tabular} & 10 & 0.5 & Euclidean & true & 0.866 \\ \hline
\begin{tabular}[c]{@{}l@{}}SonyAIBORobotSurface2\end{tabular} & 10 & 0.5 & Manhattan & true & 0.866 \\ \hline
\begin{tabular}[c]{@{}l@{}}SonyAIBORobotSurface2\end{tabular} & 10 & 0.5 & Minkowski & true & 0.866 \\ \hline
Wafer & 1 & 0.25 & Euclidean & true & 0.994 \\ \hline
Wafer & 1 & 0.25 & Minkowski & true & 0.994 \\ \hline
Wafer & 5 & 0.25 & Euclidean & true & 0.994 \\ \hline
Wafer & 5 & 0.25 & Minkowski & true & 0.994 \\ \hline
Wafer & 10 & 0.25 & Euclidean & true & 0.994 \\ \hline
Wafer & 10 & 0.25 & Minkowski & true & 0.994 \\ \hline
Wafer & 1 & 0.5 & Chebyshev & true & 0.994 \\ \hline
Wafer & 5 & 0.5 & Chebyshev & true & 0.994 \\ \hline
Wafer & 10 & 0.5 & Chebyshev & true & 0.994 \\ \hline
\caption{Result parameters for classifier DTW1NN and instance selection method Dist2kNN}
\label{tab:result_parameter_tables_dtw1nn_1}\\
\end{longtable}
\end{landscape}

\begin{landscape}
\begin{longtable}{|l|l|l|l|l|l|}
\hline
\rowcolor[HTML]{9B9B9B} 
\textbf{dataset} & \textbf{kNeighbors} & \textbf{odSelRatio} & \textbf{odDistFunc} & \textbf{odDescOrder} & \textbf{TP\_RATE} \\ \hline
\endhead
Earthquakes & 5 & 0.9 & Manhattan & true & 0.791 \\ \hline
Earthquakes & 10 & 0.9 & Manhattan & true & 0.791 \\ \hline
FordA & 1 & 0.25 & Manhattan & false & 0.664 \\ \hline
\begin{tabular}[c]{@{}l@{}}ItalyPowerDemand\end{tabular} & 1 & 0.9 & Euclidean & false & 0.976 \\ \hline
\begin{tabular}[c]{@{}l@{}}ItalyPowerDemand\end{tabular} & 1 & 0.9 & Manhattan & false & 0.976 \\ \hline
\begin{tabular}[c]{@{}l@{}}ItalyPowerDemand\end{tabular} & 1 & 0.9 & Minkowski & false & 0.976 \\ \hline
\begin{tabular}[c]{@{}l@{}}ItalyPowerDemand\end{tabular} & 5 & 0.9 & Chebyshev & false & 0.976 \\ \hline
\begin{tabular}[c]{@{}l@{}}ItalyPowerDemand\end{tabular} & 5 & 0.9 & Euclidean & false & 0.976 \\ \hline
\begin{tabular}[c]{@{}l@{}}ItalyPowerDemand\end{tabular} & 5 & 0.9 & Manhattan & false & 0.976 \\ \hline
\begin{tabular}[c]{@{}l@{}}ItalyPowerDemand\end{tabular} & 5 & 0.9 & Minkowski & false & 0.976 \\ \hline
\begin{tabular}[c]{@{}l@{}}ItalyPowerDemand\end{tabular} & 10 & 0.9 & Manhattan & false & 0.976 \\ \hline
Lighting2 & 5 & 0.05 & Manhattan & false & 0.885 \\ \hline
MoteStrain & 5 & 0.9 & Chebyshev & true & 0.89 \\ \hline
MoteStrain & 5 & 0.9 & Manhattan & true & 0.89 \\ \hline
MoteStrain & 10 & 0.9 & Chebyshev & true & 0.89 \\ \hline
\begin{tabular}[c]{@{}l@{}}SonyAIBORobotSurface1\end{tabular} & 1 & 0.5 & Chebyshev & false & 0.885 \\ \hline
\begin{tabular}[c]{@{}l@{}}SonyAIBORobotSurface2\end{tabular} & 1 & 0.5 & Euclidean & false & 0.867 \\ \hline
\begin{tabular}[c]{@{}l@{}}SonyAIBORobotSurface2\end{tabular} & 1 & 0.5 & Minkowski & false & 0.867 \\ \hline
Wafer & 5 & 0.5 & Chebyshev & false & 0.995 \\ \hline
\caption{Result parameters for classifier DTW1NN and instance selection method LDIS}
\label{tab:result_parameter_tables_dtw1nn_2}\\
\end{longtable}
\end{landscape}

\begin{landscape}
\begin{longtable}{|l|l|l|l|l|l|}
\hline
\rowcolor[HTML]{9B9B9B} 
\textbf{dataset} & \textbf{kNeighbors} & \textbf{odSelRatio} & \textbf{odDistFunc} & \textbf{odDescOrder} & \textbf{TP\_RATE} \\ \hline
\endhead
Earthquakes & 1 & 0.5 & Manhattan & false & 0.799 \\ \hline
Earthquakes & 1 & 0.5 & Manhattan & false & 0.799 \\ \hline
FordA & 10 & 0.25 & Minkowski & true & 0.655 \\ \hline
FordA & 10 & 0.25 & Euclidean & true & 0.655 \\ \hline
FordA & 10 & 0.5 & Manhattan & true & 0.655 \\ \hline
FordA & 5 & 0.5 & Manhattan & true & 0.655 \\ \hline
\begin{tabular}[c]{@{}l@{}}ItalyPowerDemand\end{tabular} & 10 & 0.25 & Euclidean & true & 0.977 \\ \hline
\begin{tabular}[c]{@{}l@{}}ItalyPowerDemand\end{tabular} & 10 & 0.25 & Euclidean & true & 0.977 \\ \hline
\begin{tabular}[c]{@{}l@{}}ItalyPowerDemand\end{tabular} & 10 & 0.25 & Minkowski & true & 0.977 \\ \hline
\begin{tabular}[c]{@{}l@{}}ItalyPowerDemand\end{tabular} & 10 & 0.25 & Minkowski & true & 0.977 \\ \hline
Lighting2 & 1 & 0.9 & Chebyshev & true & 0.869 \\ \hline
MoteStrain & 1 & 0.9 & Manhattan & false & 0.89 \\ \hline
\begin{tabular}[c]{@{}l@{}}SonyAIBORobotSurface1\end{tabular} & 1 & 0.5 & Chebyshev & true & 0.87 \\ \hline
\begin{tabular}[c]{@{}l@{}}SonyAIBORobotSurface1\end{tabular} & 1 & 0.5 & Chebyshev & true & 0.87 \\ \hline
\begin{tabular}[c]{@{}l@{}}SonyAIBORobotSurface2\end{tabular} & 10 & 0.5 & Chebyshev & true & 0.886 \\ \hline
Wafer & 1 & 0.05 & Euclidean & true & 0.995 \\ \hline
Wafer & 1 & 0.05 & Minkowski & true & 0.995 \\ \hline
Wafer & 5 & 0.05 & Chebyshev & true & 0.995 \\ \hline
\caption{Result parameters for classifier DTW1NN and instance selection method LKRR}
\label{tab:result_parameter_tables_dtw1nn_3}\\
\end{longtable}
\end{landscape}
\section{Literature search}
\label{app:search_results_example}
Example of search results for topic time series analysis with the following search terms being used:
\begin{itemize}
\item “Time series analysis” OR “Time-series analysis”
\item “Time series prediction” OR “Time-series prediction”
\item “Time series modeling” OR “Time-series modeling”
\item “Time series forecasting” OR “Time-series forecasting”
\item “Time series data mining” OR “Time-series data mining”
\item “Time series data modeling” OR “Time-series data modeling”
\end{itemize}

\begin{landscape}
\begin{longtable}{|l|l|l|l|}
\hline
\rowcolor[HTML]{9B9B9B} 
\textbf{Plattform} & \textbf{Params} & \textbf{\#Results} & \textbf{\#Relevant} \\ \hline
\endhead
ScienceDirect & \begin{tabular}[c]{@{}l@{}}Search in: Title, abstract, keywords\\ Article type: Review articles\\ Year $>$ 1990\\ AND (review OR survey OR introduction OR tutorial)\end{tabular} & 138 & 7 \\ \hline
arXiv.org & \begin{tabular}[c]{@{}l@{}}Search in: Title, abstract\\ Subject: Computer Science, Mathematics, \\ Statistics\\ Date range: 1990-2018\end{tabular} & 528 & 17 \\ \hline
DE Gruyter & \begin{tabular}[c]{@{}l@{}}Search in: Title\\ Subject: Computer Sciences, Business and \\ Economics, Physics\\ 1990-2018\end{tabular} & 109 & 1 \\ \hline
Scopus & \begin{tabular}[c]{@{}l@{}}Search field: Document title\\ Date range: 2000-2018\\ Document type: Article OR Review\\ Subject area: Computer Science\end{tabular} & 737 & 11 \\ \hline
Web of Science & \begin{tabular}[c]{@{}l@{}}Search in: Title\\ WEB OF SCIENCE CATEGORIES: \\ (STATISTICS PROBABILITY OR \\ MATHEMATICS OR COMPUTER SCIENCE \\ THEORY METHODS OR \\ COMPUTER SCIENCE SOFTWARE \\ ENGINEERING OR COMPUTER SCIENCE \\ ARTIFICIAL INTELLIGENCE OR \\ COMPUTER SCIENCE \\ INFORMATION SYSTEMS )\\ Timespan: 1900-2018\\ Document Types: Article, Proceedings paper, Review\end{tabular} & 618 & 6 \\ \hline
IEEE Xplore & \begin{tabular}[c]{@{}l@{}}Search in: Document title\\ Content types: Conferences, Journals \& Magazines\\ Year Range: 2000-2018\end{tabular} & 893 & 18 \\ \hline
EmeraldInstight & \begin{tabular}[c]{@{}l@{}}Search in: Publication title\\ Publication date: 01.2000-12-2018\end{tabular} & 130 & 2 \\ \hline
SpringerLink & \begin{tabular}[c]{@{}l@{}}Search field: Title\\ Discipline: Computer Science, Statistics, \\ Mathematics\\ Include Preview-Only content: False\end{tabular} & 390 & 7 \\ \hline
Google Scholar & \begin{tabular}[c]{@{}l@{}}with the exact phrase where my words occur: \\ in the title of the article\\ Return articles dated between: 2000-2018\end{tabular} & 144 & 6 \\ \hline
DOAJ & \begin{tabular}[c]{@{}l@{}}Search field: Title \\ Subject: Science, Technology, Mathematics, \\ Information Technology\end{tabular} & 368 & 4 \\ \hline
Recommendations & - & - & 6 \\ \hline
\caption{Search parameters and results overview for all platforms for the topic "time series analysis"}
\end{longtable}
\end{landscape}
% If the references do not show up, try to run ``biber main'' on the command line and then recompile
\printbibliography              %% remove, if using BibTeX instead of biblatex
% \include{further_ressources}  %% this is a suggestion: you have to create this file on demand

%%%% end{document}
\end{document}